\documentclass[10pt,twocolumn,letterpaper]{article}

\usepackage[pagenumbers]{cvpr} 

\usepackage{graphicx}
\usepackage{amsmath}
\usepackage{amssymb}
\usepackage{booktabs}

\usepackage{subcaption}
\usepackage{algpseudocode}
\usepackage[linesnumbered,ruled]{algorithm2e}
\usepackage{booktabs}
\usepackage[dvipsnames]{xcolor}
\usepackage{nccmath}
\usepackage{array}
\usepackage{enumitem}
\usepackage{siunitx}
\usepackage{multirow}
\newcolumntype{P}[1]{>{\centering\arraybackslash}p{#1}}
\newcolumntype{M}[1]{>{\centering\arraybackslash}m{#1}}
\usepackage{pifont}
\newcommand{\cmark}{\ding{51}}%
\newcommand{\xmark}{\ding{55}}%
\newcommand{\comment}[1]{}

\usepackage[pagebackref,breaklinks,colorlinks]{hyperref}

\usepackage[capitalize]{cleveref}
\crefname{section}{Sec.}{Secs.}
\Crefname{section}{Section}{Sections}
\Crefname{table}{Table}{Tables}
\crefname{table}{Tab.}{Tabs.}

\def\cvprPaperID{5026} 
\def\confName{CVPR}
\def\confYear{2022}

\begin{document}

\title{Ego2Hands: A Dataset for Egocentric Two-hand Segmentation and Detection}


\author{Fanqing Lin\\
Brigham Young University\\
{\tt\small flin2@byu.edu}
\and
Brian Price\\
Adobe Research\\
{\tt\small bprice@adobe.com}
\and 
Tony Martinez\\
Brigham Young University\\
{\tt\small martinez@cs.byu.edu}
}
\maketitle
\begin{abstract}
Hand segmentation and detection in truly unconstrained RGB-based settings is important for many applications. However, existing datasets are far from sufficient in terms of size and variety due to the infeasibility of manual annotation of large amounts of segmentation and detection data. As a result, current methods are limited by many underlying assumptions such as constrained environment, consistent skin color and lighting. In this work, we present Ego2Hands, a large-scale RGB-based egocentric hand segmentation/detection dataset that is semi-automatically annotated and a color-invariant compositing-based data generation technique capable of creating training data with large quantity and variety. For quantitative analysis, we manually annotated an evaluation set that significantly exceeds existing benchmarks in quantity, diversity and annotation accuracy. We provide cross-dataset evaluation as well as thorough analysis on the performance of state-of-the-art models on Ego2Hands to show that our dataset and data generation technique can produce models that generalize to unseen environments without domain adaptation.
\end{abstract}

\begin{figure}
  \begin{subfigure}[b]{\linewidth}
  \centering
  \includegraphics[width=0.9\linewidth]{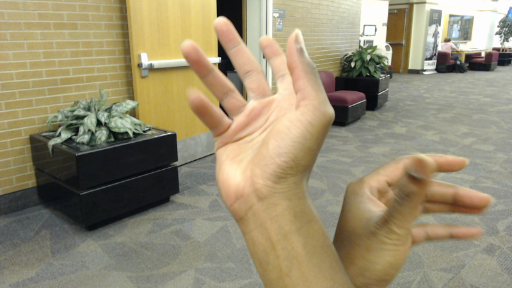}
  \end{subfigure}
  \begin{subfigure}[b]{\linewidth}
  \centering
  \vspace{0.1cm}
  \includegraphics[width=0.9\linewidth]{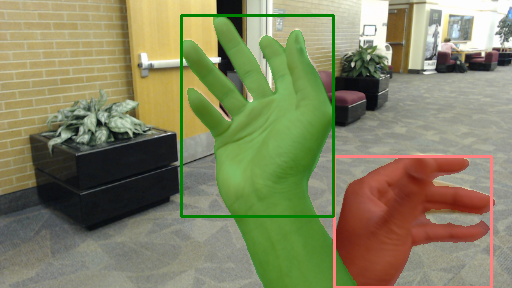}
  \end{subfigure}
  \caption{Our proposed dataset and training scheme enables domain generalization for two-hand segmentation and detection. Given an image with a new environment and hands not present in the training data (top), the trained models can provide accurate segmentation and detection results for both hands regardless of inter-hand occlusion (bottom).}
  \label{fig:intro_img}
  \vspace{-6mm}
\end{figure}
\section{Introduction}
\indent With the rapid growing usage of wearable technologies generating massive volumes of egocentric image data \cite{oculus, htcvive, gopro, narrativeclip}, the ability for machines to understand the human hands becomes crucial for applications such as human-computer interaction (HCI), activity logging, gesture/sign language recognition and VR/AR. Consequently, hand detection and segmentation are fundamental in areas such as 2D/3D hand pose estimation \cite{Taylor, Mueller, FLin} and gesture recognition \cite{Bambach, Khan}. However, hand segmentation on images in the wild is extremely challenging due to numerous factors: vastness of the color space, different skin color/texture, inter-hand occlusion, complex background noise, motion blur, lighting type/color, shadow features, speed and model size requirement, etc. As a result, existing color-based approaches can only perform in constrained environments with proper lighting and skin color consistent with the training data. These limitations are largely due to the lack of annotated segmentation data, a common limiting factor for segmentation tasks because manual annotation is oftentimes required but infeasible for large-scale data generation.\\
\indent In this work, we aim to push the boundary for the task of real-time egocentric two-hand segmentation and detection on images in the wild (Fig. \ref{fig:intro_img}). Since hand segmentation and detection are highly correlated and both imperative for subsequent applications, we find it natural to tackle both tasks simultaneously. \\
\indent We first address the issue of the lack of annotated data. In general, real-world RGB data with segmentation ground truth is very labor-intensive to annotate. For this reason, existing hand segmentation datasets \cite{YLi, CLi, Bambach, Khan, YLi2, Cai} lack the quantity and sufficient variety necessary for learning-based approaches. Although synthetic data \cite{FLin} with perfect ground truth can be generated with little cost, it is more challenging for methods trained on synthetic data to generalize to real-world data as CNNs are sensitive to textural differences between domains. We propose to collect massive amounts of segmentation data for the egocentric right hands that can be automatically annotated in a green screen setting, and a novel data generation technique that composites training instances by combining a pair of randomly selected right hands with one horizontally flipped as the left hand. In addition, we introduce a novel semi-automated annotation method for the heatmap energy of the hand. Unlike hand segmentation, the hand energy excludes the arm and can be used for hand detection. In order to develop a color-invariant approach, we explore the grayscale image space coupled with an edge map as the input space and show successful generalization to unseen environments. This data generation method can push segmentation models beyond the limitation of a fixed-sized training set and evaluation set and enable models to produce accurate segmentation and detection results in unseen environments without domain adaptation, which can also be easily applied to further improve model accuracy for specific environments.\\
\indent We introduce Ego2Hands which includes a training set with $\sim$180,000 unique right-hand instances and an evaluation set with 2,000 manually annotated frames from diverse video sequences. In-depth comparison and cross-dataset evaluation between Ego2Hands and previous datasets show the superiority of our dataset in quantity, diversity, annotation accuracy and generalization ability. For quantitative analysis, we provide comprehensive comparison between the state-of-the-art approaches on our dataset.\\
\indent To demonstrate the benefits of having fast and accurate two-hand segmentation and detection results, we show real-time gesture control in the supplementary video. Our work opens up promising directions for two-hand gesture control systems using only low-cost RGB input devices.
\section{Related Works}
\noindent\textbf{Depth-based datasets.} Early works \cite{Tompson, Sharp} have proposed depth based datasets for single-hand segmentation. \cite{Bojja} introduced a dataset that enabled two-hand segmentation from a third-person viewpoint. Additionally, \cite{Taylor, Mueller} presented datasets with two-hand segmentation data for egocentric pose reconstruction and hand tracking, which indicates the significance of two-hand segmentation for other gesture-related tasks as it provides information not only on the location and the shape of the hands, but also on how interacting hands are occluded by each other or other objects.\\
\indent In this work, we focus on a RGB-based setting as depth cameras have additional setup overhead and indoor requirements with higher power consumption and cost. This leads to limited applicable applications compared to the ubiquitous RGB cameras. We point out that existing depth-based datasets obtain the segmentation ground truth using color thresholding and require subjects to wear thin colored gloves. As a result, depth-based hand segmentation datasets are not suitable for training RGB-based approaches. \\
\noindent\textbf{Color-based datasets.} Pioneering work \cite{CLi} contributed three egocentric videos (EDSH1, EDSH2 and EDSH-kitchen) with varying illumination for binary hand segmentation. For activity recognition, \cite{YLi} proposed the Georgia Tech Egocentric Activity Dataset (GTEA) with 625 frames consisting of two-hand labeling and 38 frames with binary-labeling. \cite{YLi2} later published an extended version (EGTEA) with 12,799 frames with binary-labeling and 1,048 frames with two-hand labeling. To enable hand segmentation in more unconstrained settings, \cite{Bambach} introduced EgoHands as the first large-scale hand segmentation dataset with 4,800 annotated frames consisting of a maximum of 4 interacting hands. For the same purpose, \cite{Khan} additionally introduced EgoYouTubeHands (EYTH) with $\sim$1290 binary-labeled frames from three Youtube videos and HandOverFace (HoF) with 300 annotated frames from third-person Web images. To demonstrate cross-dataset adaptation performance, \cite{Cai} provided binary-labelling for 855 and 488 frames from human grasping datasets UTG \cite{Cai2} and YHG \cite{Bullock} respectively. To address the issue of data scarcity, \cite{FLin} introduced a large-scale synthetic dataset (Ego3DHands) with a total of over 100,000 annotated frames on two hands.
\section{Ego2Hands}
\indent There are several major factors that prevent existing datasets from enabling generalization of two-hand segmentation in unseen domains. First, many datasets \cite{CLi, Khan, Cai} only contain binary-label for hand segmentation, which does not distinguish between the left and right hand. For datasets that contain frames with two-hand labels, the quantity of annotated data is very limited due to the need for manual segmentation annotation. The annotated frames also possess limited diversity in terms of background environment, lighting, skin tone and range of motion. The aforementioned properties are obtainable for synthetic data, however, models trained on synthetic data generally generalize poorly to real-world data due to the domain gap. \\
\indent To enable generalization to real-world domains, we introduce a large-scale dataset Ego2Hands that consists of 188,362 annotated frames in the training set for the right hand. Segmentation masks are obtained by automatically removing the background in a green screen setting. 22 participants with diverse skin colors and hand features are instructed to perform free hand motion that covers a wide range of locations/poses while recording using a head-mount webcam (Logitech C922) at 30 fps. This process allows simple and fast data collection for segmentation data. We obtained consent from all participants and the collected egocentric hand data does not contain personally identifiable information.
\begin{figure}[t!]
  \centering
  \begin{subfigure}[b]{0.44\linewidth}
    \includegraphics[width=\linewidth]{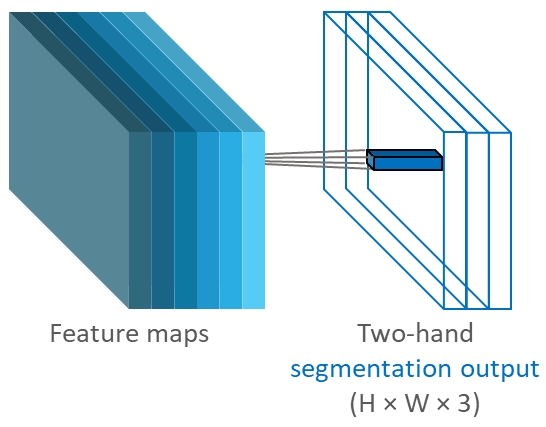}
    \caption{Original output channel}
  \end{subfigure}
  \begin{subfigure}[b]{0.54\linewidth}
    \includegraphics[width=\linewidth]{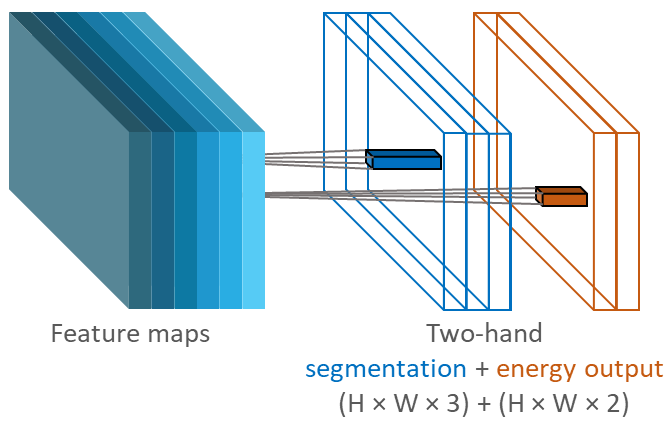}
    \caption{Modified output channels}
  \end{subfigure}
  \caption{Addition of the convolutional output layer to enable estimation of both two-hand segmentation and energy.}
  \label{fig:seg_energy_channel}
  \vspace{-3mm}
\end{figure}
\subsection{Hand Energy Annotation}
There are two major types of representation for hand segmentation. Some datasets \cite{YLi, CLi, YLi2, Cai} include the arm in segmentation and neglect the task of hand detection, while others \cite{Bambach, Khan} contain hand segmentation that excludes the arm, therefore combining the task of segmentation and detection into one. The disadvantage of the second representation is the loss of detection information during inter-hand occlusion. To enable both hand segmentation and detection, we choose to provide annotation for segmentation as well as detection data in the form of heatmap energy. \\
\indent Although segmentation annotation can be automatically obtained by green screen removal, obtaining the hand energy annotation for over 180k frames that excludes the arm is a nontrivial task. Our annotation process can be divided into 1) transfer learning using synthetic and real-world data and 2) semi-automated "human-in-the-loop" annotation. \\
\indent In stage 1, we develop a novel transfer learning technique to leverage segmentation data from two domains and transfer the knowledge of energy estimation from one to the other. To this end, we propose to add an additional output layer to the segmentation network for energy estimation as shown in Fig. \ref{fig:seg_energy_channel}. We theorize that the additional task of energy estimation should not conflict with segmentation as they share similar features. We apply sigmoid activation to the energy output and compute the energy loss $\mathcal{L}_{energy}$ using Mean Squared Error (MSE). The combined loss for network optimization can then be formulated as:
\begin{equation}\label{eq:loss-seg-energy}
\mathcal{L}_{total} = \mathcal{L}_{seg} + \mathcal{L}_{energy}
\end{equation}
where $\mathcal{L}_{seg}$ is computed using the standard Cross Entropy Loss. We show in Section \ref{sec:experiments} that the modified models can achieve high accuracy on both tasks simultaneously.\\
\indent To prepare data from the synthetic domain, we follow \cite{FLin} by generating an internal synthetic dataset $\text{Ego3DHands}_{R}$ with only the right hand, which contains perfect annotation for hand segmentation as well as energy. For data in the real-world domain, we use the $\sim$180k frames from Ego2Hands that contain the annotation only for segmentation. To simplify the input space, all training images have background removed using the hand segmentation masks. Finally, we jointly train the proposed HandSegNet \cite{FLin} on both $\text{Ego3DHands}_{R}$ and Ego2Hands using the following loss:\\
\begin{equation}\label{eq:joint_loss}
\begin{gathered}
\mathcal{L}_{combined} = \mathcal{L}^{synth}_{seg} + \mathcal{L}^{real}_{seg} + \mathcal{L}^{synth}_{energy},
\end{gathered}
\end{equation}
where $\mathcal{L}^{synth}$ and $\mathcal{L}^{real}$ are losses computed using synthetic and real-world data respectively. By learning to estimate hand segmentation from both domains, the model is pushed to learn shared features between domains. As a result, knowledge learned using $\mathcal{L}^{synth}_{energy}$ can be transferred to estimate hand energy on real-world images from Ego2Hands with promising accuracy.\\
\indent In stage 2, inspired by interactive video object segmentation methods \cite{FLin2}, we build an annotation tool that uses the trained HandSegNet to automatically estimate the hand energy given the input images of Ego2Hands. In cases where the estimation becomes inaccurate, the human annotator manually corrects the error to generate the ground truth. HandSegNet is immediately finetuned on the corrected data for more accurate future predictions. This  semi-automated annotation method allows us to generate accurate hand energy for Ego2Hands in a feasible manner. Energy annotation samples are included in the supplementary document.
\begin{figure*}[t!]
  \centering
  \begin{subfigure}[b]{1.0\linewidth}
    \includegraphics[width=\linewidth]{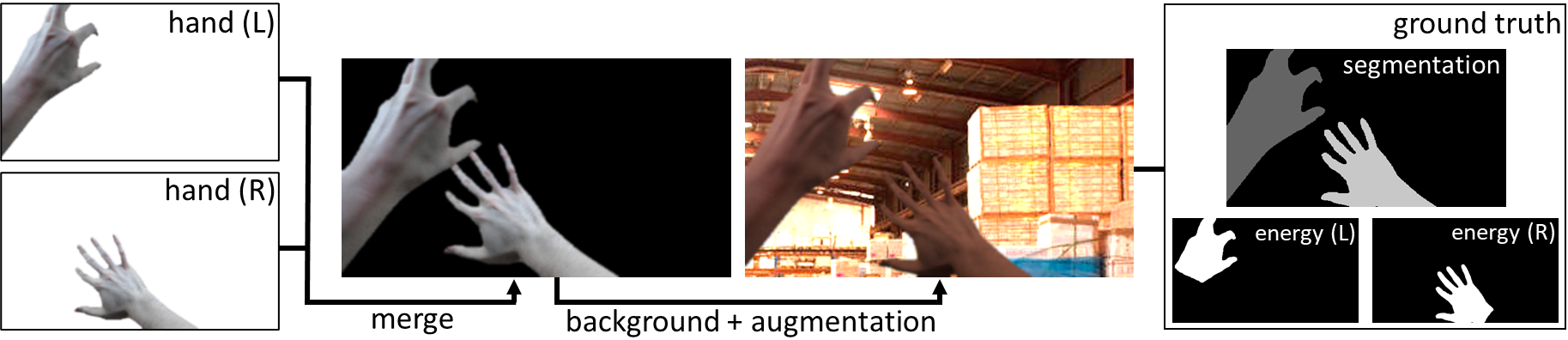}
  \end{subfigure}
  \caption{Pipeline for training data composition. Ground truth data includes segmentation and energy annotation for the left and right hand.}
  \label{fig:data-composition}
  \vspace{-3mm}
\end{figure*}
\subsection{Data Composition for Training}
\indent We use randomly selected pairs of right hand images (with one flipped as the left hand) in the training set of Ego2Hands for data composition. Although actual images of the two hands are more realistic than the composited version, we point out that obtaining two-hand segmentation annotation is infeasible at large-scale if both hands are present in the collected images since green screen removal creates a binary segmentation label. For the background images, we use the 19,216 images provided by \cite{FLin} with the additional 14,997 high-quality images in the DAVIS datasets \cite{Perazzi, PontTuset}. This results in approximately $\num{1.21e15}$ unique hand-scene combinations prior to data augmentation.\\
\indent With the obtained hand energy, we are able to composite more realistic training images by selecting the proper overlaying order. After random selection of the left and right hand, the hand with the larger energy sum is selected to be overlaid on top of the other hand. To obtain a smooth compositing boundary, we apply erosion and gaussian blur on the alpha-channel. For each composited image, we further data augment by applying 1) random horizontal and downward vertical translation within reasonable ranges on each hand, 2) random smoothing with various kernel sizes to simulate blur from motion or auto-focus, 3) color augmentation on the hands and background images, 4) random horizontal flips and cropping on background images, and 5) 10\% drop rate for each hand to accommodate single-hand scenarios. To enable domain adaptation on specific environments, we can simply use the background images collected from that scene for compositing training images. Unlike conventional datasets with fixed sizes, we composite images at training time for much higher quantity and more diversity, which is essential for domain generalization. Fig. \ref{fig:data-composition} illustrates our data composition process.
\subsection{Evaluation Set}
\begin{figure}[t!]
  \centering
  \begin{subfigure}[b]{0.95\linewidth}
    \includegraphics[width=\linewidth]{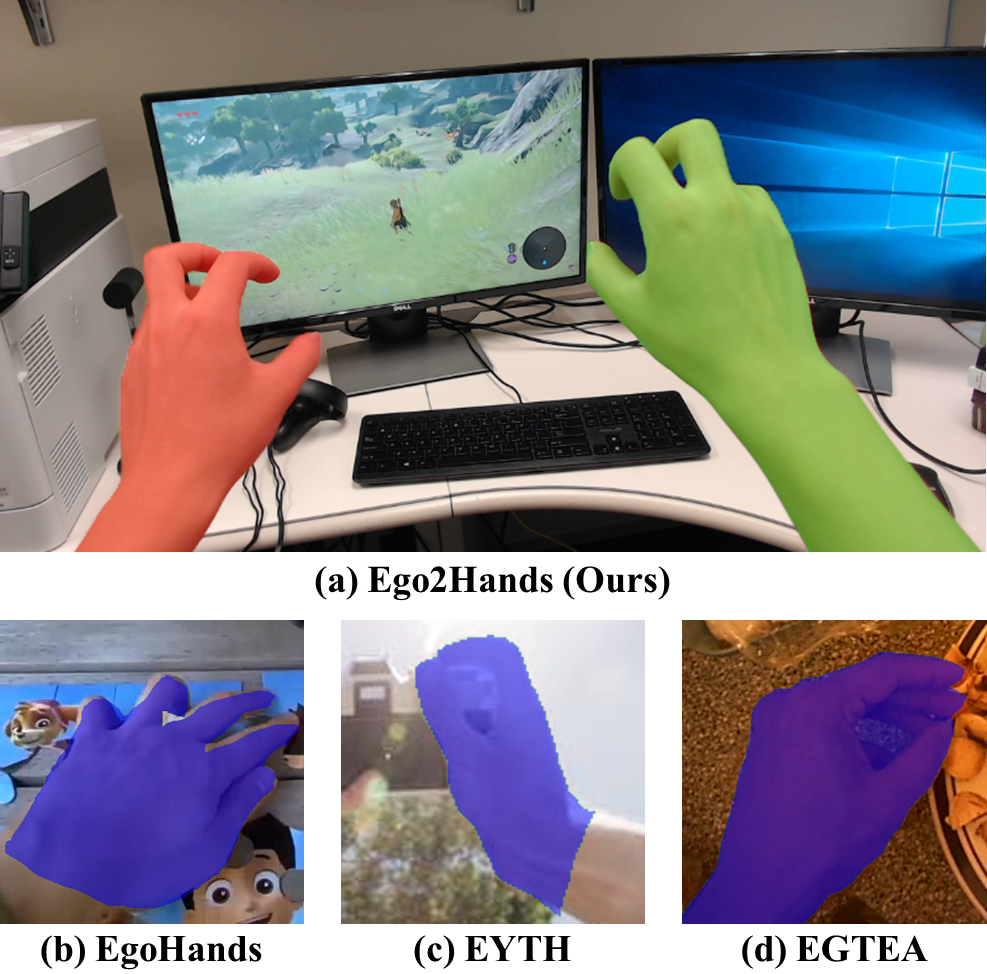}
  \end{subfigure}
  \caption{Illustration of the difference in annotation accuracy between datasets. Existing datasets contain false positive labeling for gaps and holes and potentially inaccurate boundaries (Best viewed in magnification. Annotated masks are overlaid in colors.)}
  \label{fig:dataset_test}
  \vspace{-5mm}
\end{figure}

\indent To support quantitative evaluation, we introduce an evaluation set that includes 8 videos each with 250 annotated frames. We select 4 additional participants with diverse skin tones to perform free two-hand motion in 8 different scenes under various lighting conditions. We manually annotate the segmentation as well as the energy for the left and right hand. Fig. \ref{fig:dataset_test} shows a comparison of annotation quality between Ego2Hands and other datasets. We show additional qualitative examples of all 8 sequences in the supplementary document to demonstrate the annotation quality and diversity in our sequences.
\subsection{Comparison with Existing Datasets}
\indent We show in Table \ref{tab:table_dataset_comparison} a detailed comparison between Ego2Hands and existing benchmark datasets. For real-world datasets with two-hand labeling (EgoHands, EGTEA and GTEA), the largest existing training set is EgoHands that contains 3,600 annotated frames (36/12 split) with 3 scenes in total. $\text{Ego2Hands}_{train}$ consists of 188,362 annotated frames for compositing two-hand training data that can generalize to real-world data. For a comparison between existing test sets, we notice a critical issue that the three aforementioned datasets have test sets that share the same subjects and scenes with their training sets. On top of the small quantity and diversity, this setting further weakens evaluation of the methods' generalization ability. Our $\text{Ego2Hands}_{test}$ not only contains data with diverse subjects and scenes unseen in the training set, but also provides 2,000 frames for comprehensive evaluation, which is more than the test sets of EgoHands (1,200 frames), GTEA (63 frames) and EGTEA (105 frames) combined.\\
\begin{table*}[t]
  \small
  \centering 
  \begin{tabular}{M{2.6cm}|M{0.8cm}|M{2.1cm}|M{1.7cm}|M{1.4cm}|M{1.1cm}|M{1.0cm}|M{1.1cm}|M{1.6cm}}
  \toprule
Datasets 	& 
Type & 
\begin{tabular}{@{}c@{}}\#Annotated \\ Frames\end{tabular} & 
\begin{tabular}{@{}c@{}}\#Hand \\ Instances\end{tabular} & 
\#Subjects & 
\#Scenes & 
Objects & 
\#Classes &
Resolution \\
\midrule
GTEA \cite{YLi}&Real & 663 & 1231 & 4 & 1  & Yes & 2 \& 3 & $720\times405$ \\\hline
EDSH \cite{CLi}& Real & 743 & - & 1 & 3 & Yes & 2 & $1280\times720$ \\\hline
EgoHands \cite{Bambach}& Real & 4800 & 15053 & 4  & 3 & Yes & 5 & $1280\times720$ \\\hline
EYTH \cite{Khan}& Real & 1290 & 2600  & -  & - & Yes & 2 & $384\times216$ \\\hline
HoF \cite{Khan}& Real & 300 & 507  & -  & - & No & 3 & $384\times216$ \\\hline
EGTEA \cite{YLi2}& Real & 13847 & - & 32  & 1  & Yes & 2 \& 3 & $960\times720$ \\\hline
UTG \cite{Cai}& Real & 855 & - & 5  & 2 & Yes & 2 & $480\times360$ \\\hline
YHG \cite{Cai}& Real & 488 & - & 4  & - & Yes & 2 & $480\times360$ \\\hline
Ego3DHands \cite{FLin}& Synth & 110,000 & $\sim$214,500 & 1  & - & No & 3 &  $960\times540$ \\\hline\hline
Ego2Hands (Ours)& Real & \begin{tabular}{@{}c@{}} 188,362 (train) \\ 2,000 (test) \end{tabular} &
\begin{tabular}{@{}c@{}} $\infty$ (train) \\ 4,000 (test) \end{tabular}  & 
\begin{tabular}{@{}c@{}} 22 (train) \\ 4 (test) \end{tabular}  & 
\begin{tabular}{@{}c@{}} - (train) \\ 8 (test) \end{tabular} & No & 3 & $800\times448$ \\\hline
    \end{tabular}
\caption{Statistics of available hand segmentation datasets. Datasets with \#Classes = 2 only support binary segmentation.}
\label{tab:table_dataset_comparison}
\vspace{-2mm}
\end{table*}
\comment{
\begin{figure*}[t!]
  \centering
  \begin{subfigure}[b]{1.0\linewidth}
    \includegraphics[width=\linewidth]{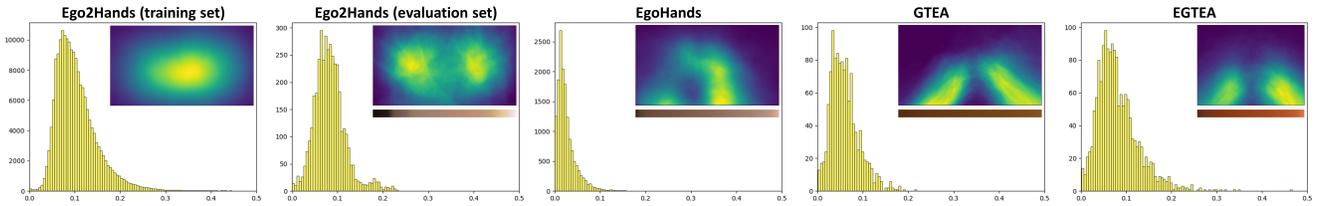}
  \end{subfigure}
  \caption{Visualization of heatmap for spatial hand occurrence, histogram for hand area relative to the image size and the color distribution of the hands for each dataset. Color distribution for the training set of Ego2Hands is not shown since color augmentation is applied. }
  \label{fig:data-dist}
  \vspace{-3mm}
\end{figure*}
}
\begin{figure*}[t]
  \centering
  \begin{subfigure}[t]{0.195\linewidth}
    \includegraphics[width=\linewidth]{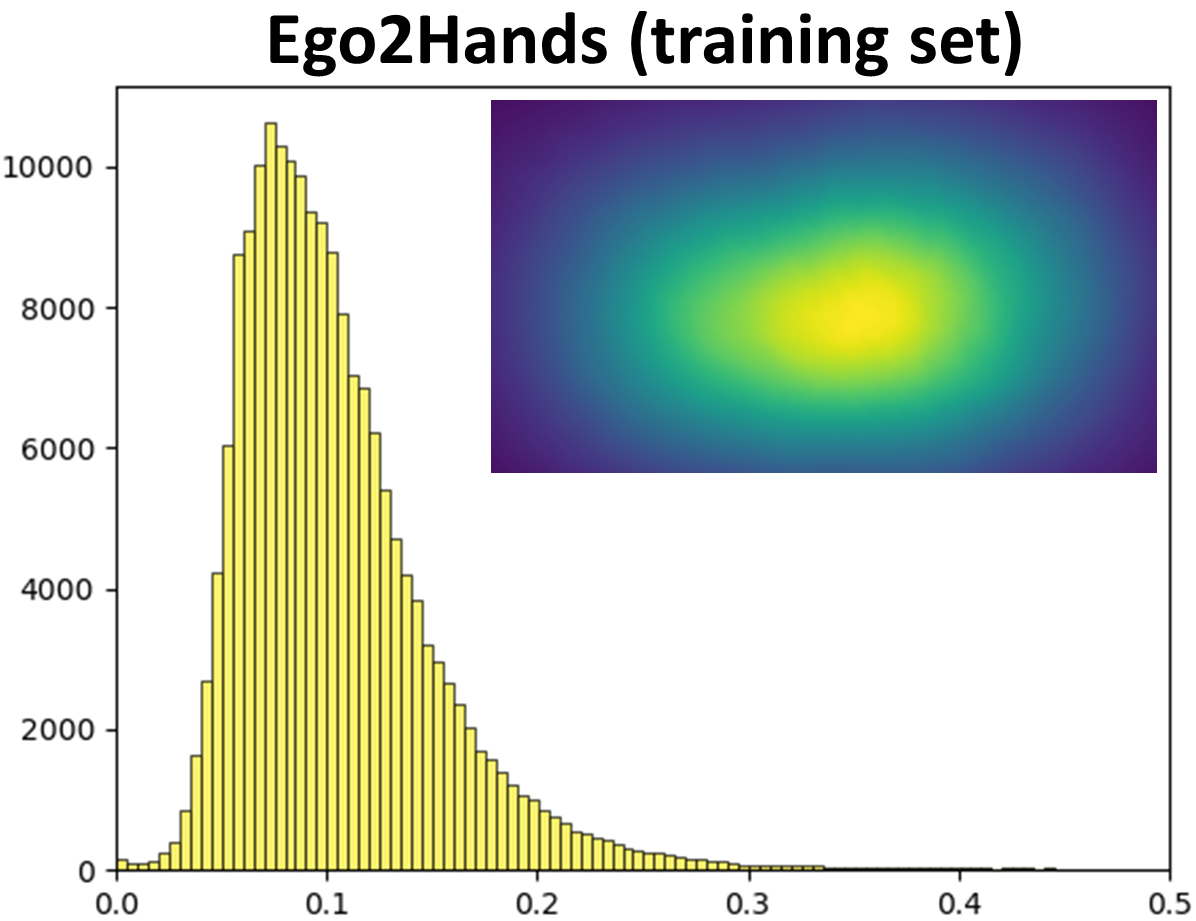}
  \end{subfigure}
  \begin{subfigure}[t]{0.195\linewidth}
    \includegraphics[width=\linewidth]{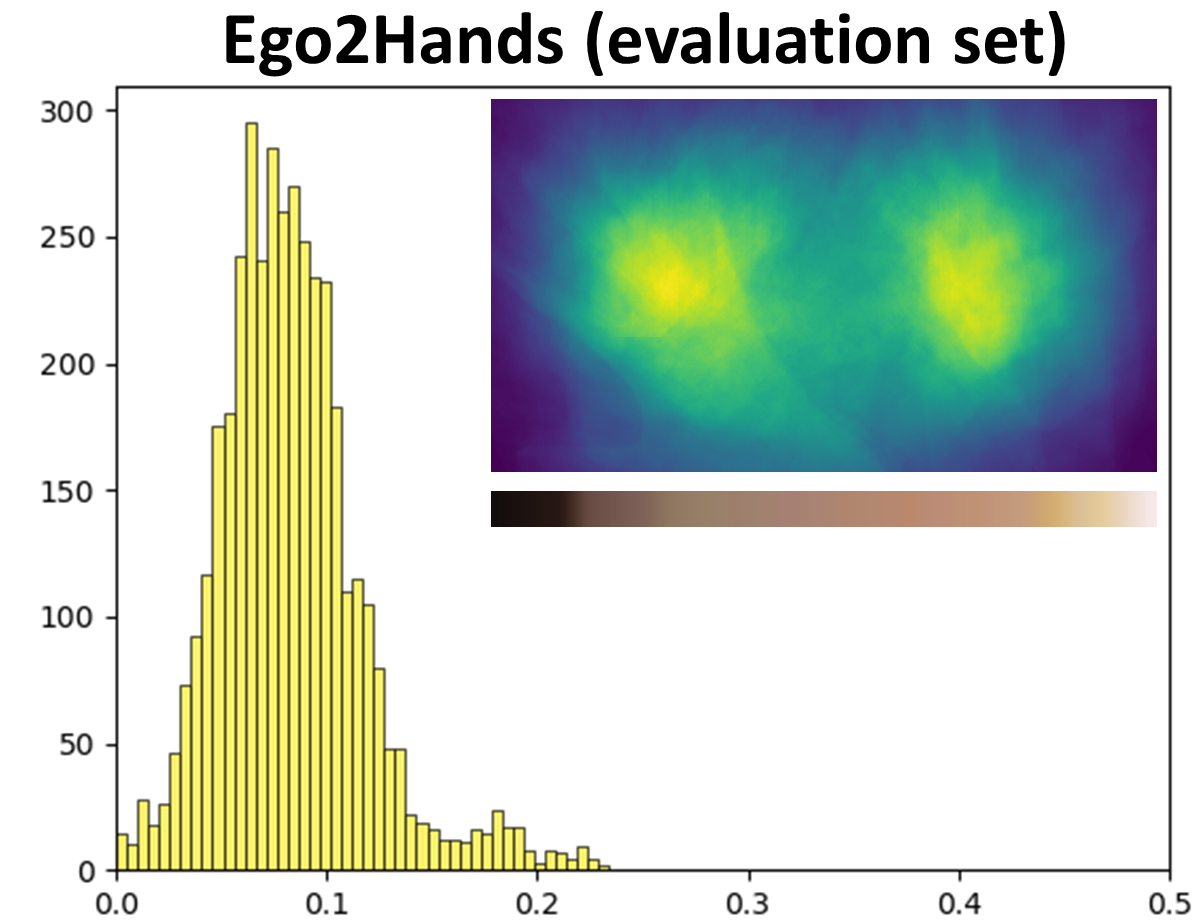}
  \end{subfigure}
  \begin{subfigure}[t]{0.195\linewidth}
    \includegraphics[width=\linewidth]{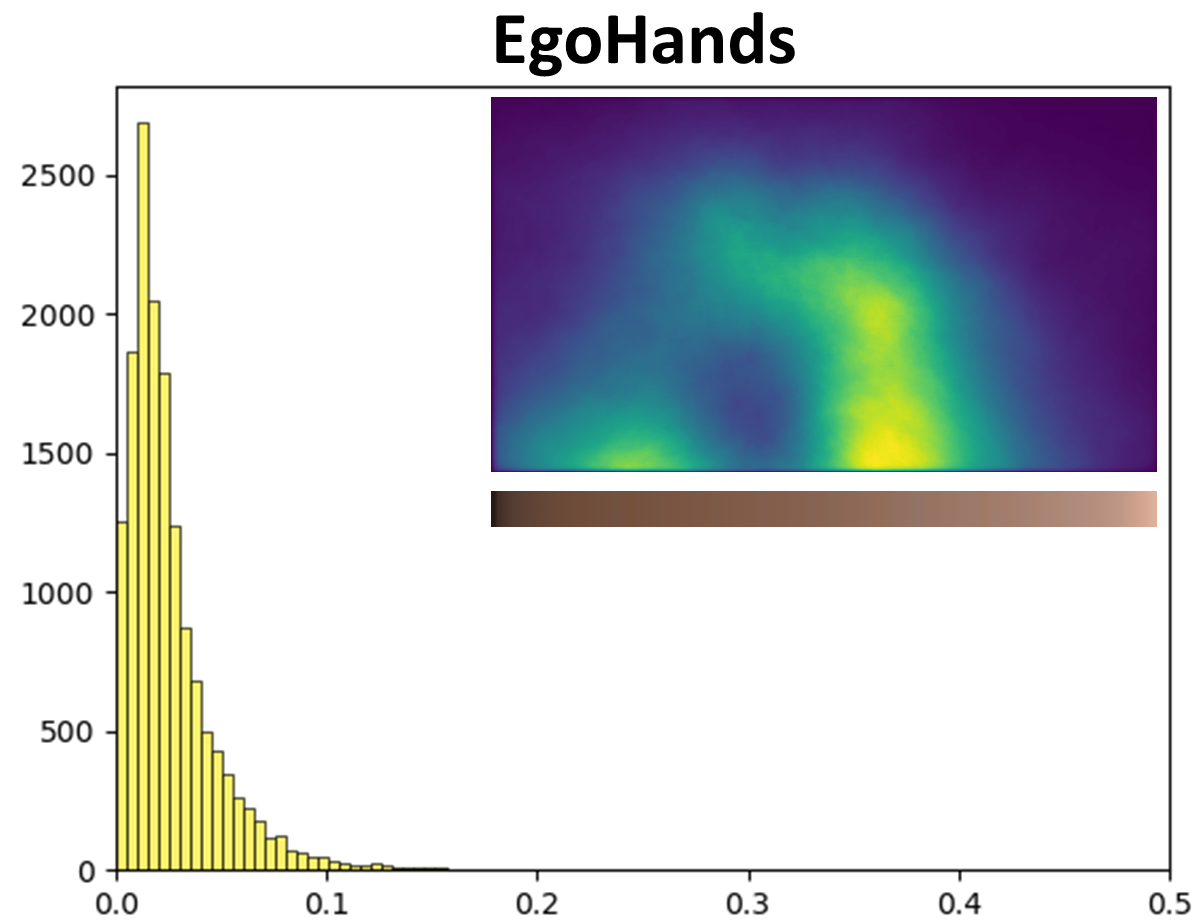}
  \end{subfigure}
  \begin{subfigure}[t]{0.195\linewidth}
    \includegraphics[width=\linewidth]{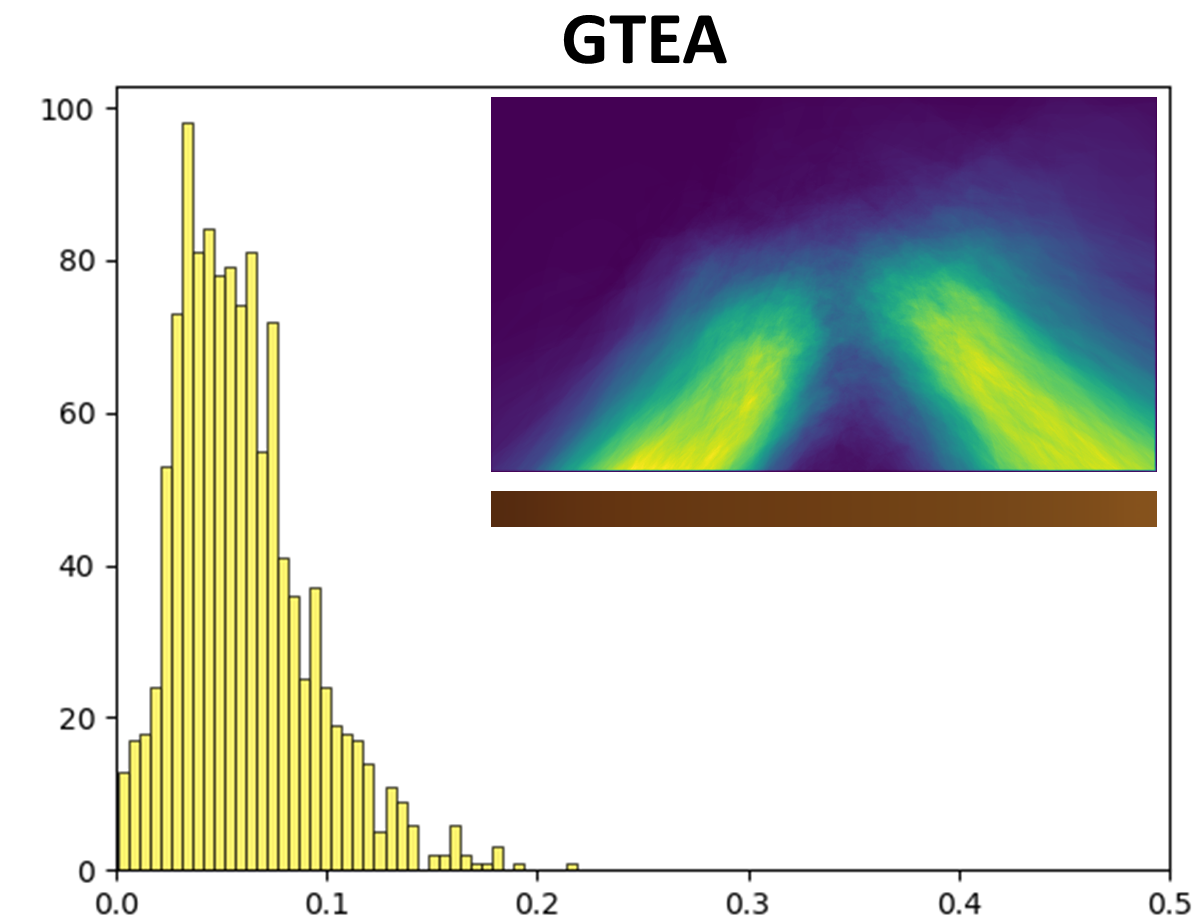}
  \end{subfigure}
  \begin{subfigure}[t]{0.195\linewidth}
    \includegraphics[width=\linewidth]{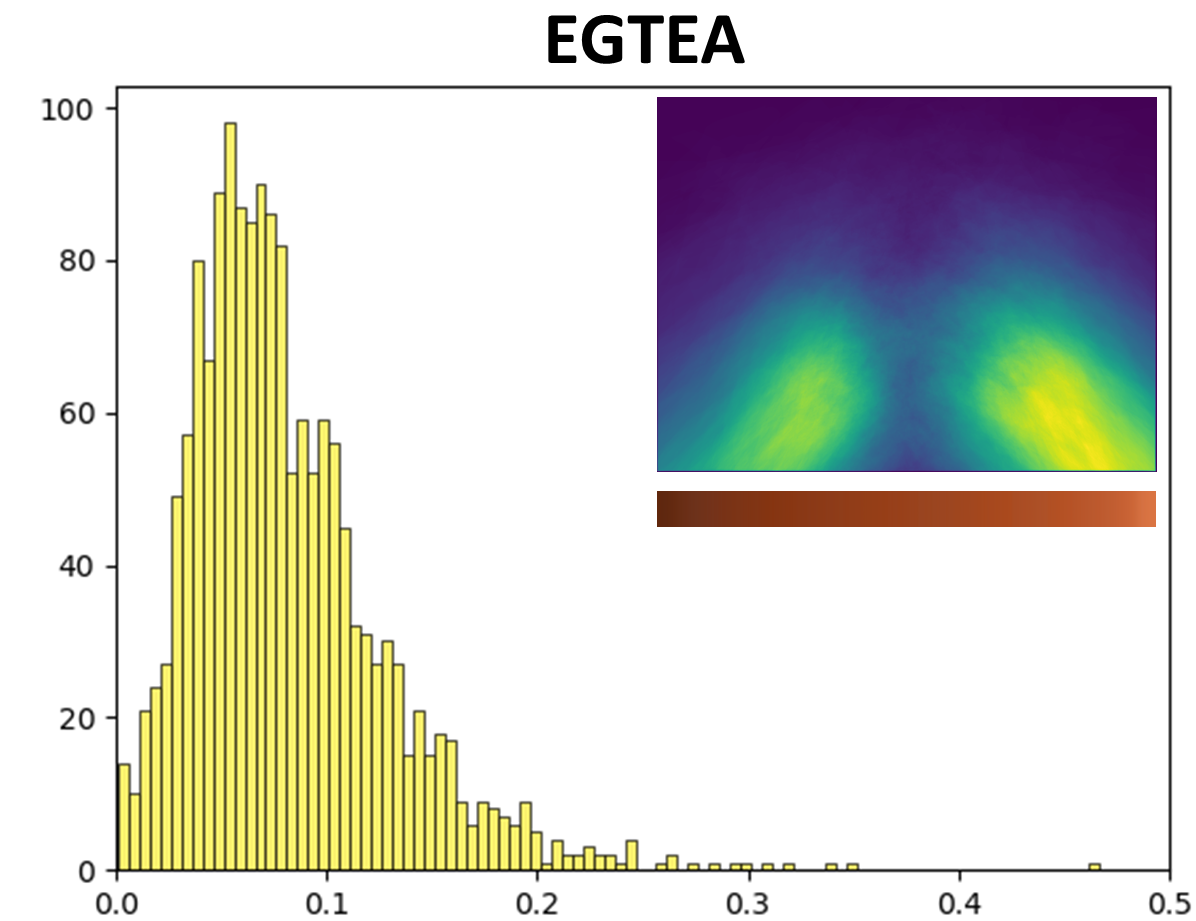}
  \end{subfigure}
  \caption{Visualization of heatmap for spatial hand occurrence, histogram for hand area relative to the image size and the color distribution of the hands for each dataset. Color distribution for the training set of Ego2Hands is not shown since color augmentation is applied.}
  \label{fig:data-dist}
  \vspace{-3mm}
\end{figure*}
\indent We compare other dataset attributes such as diversity in hand locations, hand sizes and skin colors in Fig. \ref{fig:data-dist}. We show that Ego2Hands contains hand instances evenly distributed in a wide range of locations while other datasets have hand spatial occurrence very concentrated in specific regions. Ego2Hands also demonstrates the best hand size distribution while other datasets provide hand size distribution that is heavily biased towards small sizes (the computed hand area of GTEA and EGTEA includes the arm). Hand color distribution is a very important topic rarely discussed in previous work. We see that EgoHands, GTEA and EGTEA all contain very limited variation in skin color, which is largely attributed to the limited number of subjects and scenes. On the other hand, $\text{Ego2Hands}_{test}$ provides skin color that spans a broader spectrum, covering color range that reaches close to $(0, 0, 0)$ and $(255, 255, 255)$. \\
\indent In summary, it is more difficult for existing datasets to support generalization on real-world data given their limitations in quantity and diversity. In comparison, both $\text{Ego2Hands}_{train}$ and $\text{Ego2Hands}_{test}$ provide significant improvements in all dataset attributes while additionally enabling two-hand detection.
\subsection{Color-invariant Input Domain}
\begin{figure}[t]
  \small
  \centering
  \begin{subfigure}[t]{0.24\linewidth}
    \includegraphics[width=\linewidth]{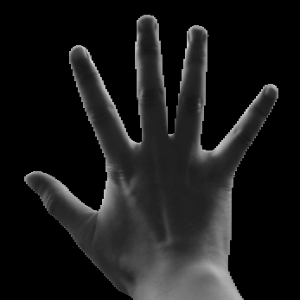}
  \end{subfigure}
  \begin{subfigure}[t]{0.24\linewidth}
    \includegraphics[width=\linewidth]{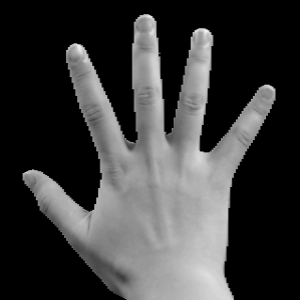}
  \end{subfigure}
  \begin{subfigure}[t]{0.24\linewidth}
    \includegraphics[width=\linewidth]{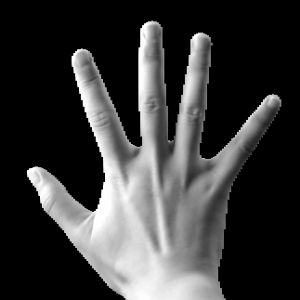}
  \end{subfigure}
  \begin{subfigure}[t]{0.24\linewidth}
    \includegraphics[width=\linewidth]{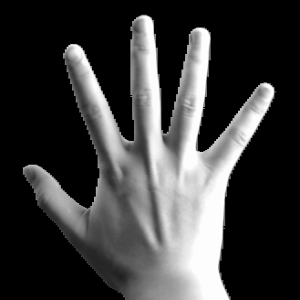}
  \end{subfigure}
  \caption{Hand images (grayscale) with different visual shadow features from different lighting directions.}
  \label{fig:light_feature_img}
  \vspace{-4mm}
\end{figure}

Despite the increased quantity in training instances, it is still challenging for deep networks to learn the complete RGB space. For instance, for hands under a particular-colored lighting (e.g. blue), learning-based models would need sufficient training data with hands in that specific color. This issue is oftentimes overlooked by previous works as their proposed datasets contain skin tone and lighting with limited variation. Consequently, we explore the grayscale image space coupled with an image edge map as input for a color-invariant approach. In the grayscale domain, we find two major factors crucial for generalization in the real-world domains: brightness and shadow features. For diversity in brightness, we replace color augmentation with brightness augmentation and scale the pixel values of both hands to shift the means to a randomly selected value $\beta \in [15, 240]$ while clipping pixel values within [0, 255]. Variation in the brightness of the hands also contributes significantly to diversity in skin tones. For different shadow features, we include light sources from various directions during data collection (Fig. \ref{fig:light_feature_img}). For quantitative comparison, we also include results obtained using the normal RGB color domain to support this design choice in a later section. 
\begin{table*}[t]
  \centering
  \begin{tabular}{P{2.7cm}|c|c|c|c|c}
  \toprule
Dataset 	 &
\begin{tabular}{@{}c@{}}Ego2Hands \\ S = 8, F = 2,000\end{tabular}&
\begin{tabular}{@{}c@{}}Ego3DHands \\ S = NA, F = 5,000\end{tabular} &
\begin{tabular}{@{}c@{}}EgoHands \\ S = 3, F = 4,800\end{tabular}&
\begin{tabular}{@{}c@{}}EGTEA \\ S = 1, F = 1,048\end{tabular} &
\begin{tabular}{@{}c@{}}GTEA \\ S = 1, F = 625\end{tabular}
\\
\midrule
Ego2Hands (Ours) & \textbf{0.832} & {\color{blue}0.749} & 0.330 & {\color{blue}0.638} & {\color{ForestGreen}0.602}\\\hline
Ego3DHands \cite{FLin} & {\color{ForestGreen}0.502} & \textbf{0.953} & {\color{ForestGreen}0.399} & 0.284 & 0.086\\\hline
EgoHands \cite{Bambach} & 0.262 & 0.160 & \textbf{0.824} & {\color{ForestGreen}0.541} & 0.376\\\hline
EGTEA \cite{YLi2} & {\color{blue}0.557} & {\color{ForestGreen}0.591} & {\color{blue}0.411} & \textbf{0.910} & {\color{blue}0.698}\\\hline
GTEA \cite{YLi} & 0.113 & 0.069 & 0.248 & 0.376 & \textbf{0.930}\\\hline
\end{tabular}
\caption{Cross-dataset evaluation computed using mIoU on hand segmentation datasets with two-hand annotation. Each row and column represent the training and evaluation set respectively. The number of different scenes and frames included in each dataset are shown for each column. The top-3 scores on each evaluation dataset are marked as {\color{black}\textbf{first}}, {\color{blue}second} and {\color{ForestGreen}third}. }
\label{tab:table_cross_eval}
\vspace{-5mm}
\end{table*}
\subsection{Cross-dataset Generalization}\label{sec:cross_eval}
\indent To demonstrate that the composited data using $\text{Ego2Hands}_{train}$ enables generalization to the real-world domains, we first perform cross-dataset evaluation on 5 existing datasets with two-hand labeling. For the base architecture, we select RefineNet \cite{GLin} for its high generalization ability across various datasets \cite{Khan}. The proposed compositing-based data augmentation on grayscale images coupled with edge map is used for training using Ego2Hands. For training using other datasets, we apply color/smoothness augmentation on the original RGB images. For the training of each dataset, the author-defined or a 90/10 split is used. For evaluation, we use the test sets of Ego2Hands and Ego3DHands which contain 2k and 5k instances respectively. Since the other datasets contain fewer instances, we evaluate on the entire dataset for more accurate results. Note that EgoHands contains 4-hand annotation and only the 2 egocentric hands are used for consistency. Averaged mean Intersection over Union (mIoU) computed on the segmentation of two hands from 3 trained model instances is reported.\\
\indent It is generally expected for a model to achieve the best score on the dataset it was trained on, for this reason, we focus on the generalization scores obtained by evaluating on other datasets. Table \ref{tab:table_cross_eval} shows that training on Ego2Hands achieves the best generalization results on Ego3DHands and EGTEA. Ego2Hands also achieves high scores on GTEA despite performing slightly worse compared to EGTEA. However, both EGTEA and GTEA are collected in 1 kitchen scene that contains very similar features. We find that training on Ego2Hands gives lower scores on EgoHands, which contains lower-quality images that can lead to a domain gap in the edge map space. \\
\indent Unlike other datasets that contain small-scale training/test sets that share the same subjects and scenes with similar features, Ego2Hands has test sequences and subjects not present in the training data. As a result, the best evaluation score of a mIoU = 0.832 obtained on $\text{Ego2Hands}_{test}$ using models trained on $\text{Ego2Hands}_{train}$ also indicates the strong generalization ability of our dataset. We argue that evaluating on Ego2Hands which contains scenes and subjects with diverse features produces more comprehensive results compared to other datasets. For example, we show in the supplementary document that models trained on other datasets all fail to generalize to sequence 2, 5, 6 and 8 of $\text{Ego2Hands}_{test}$, which contain skin color and illumination different from the data distribution of existing datasets. In addition, since only Ego2Hands and Ego3DHands contain instances with extensive inter-hand occlusion, models trained on other datasets perform poorly when hands occlude each other or cross over.\\
\indent Interestingly, Ego3DHands with synthetic data achieves decent results on Ego2Hands and EgoHands, indicating that pretraining or mixed training on synthetic data can be beneficial for real-world estimation. Surprisingly, EGTEA achieves promising results across various datasets with only 1,048 training instances. However, the large performance gap between EGTEA and $\text{Ego2Hands}_{train}$ evaluating on $\text{Ego2Hands}_{test}$ suggests that small-scale datasets are insufficient for high-quality generalization.
\section{Quantitative benchmarking}\label{sec:experiments}
\indent We evaluate existing state-of-the-art methods on the proposed evaluation set of Ego2Hands (8 sequences each with 250 annotated frames) and compare the two-hand segmentation and detection accuracy as well as the corresponding model sizes and inference speed for benchmarking. We use the mIoU as the metric for the segmentation task. For hand detection, we use the conventional metric of Average Precision that classifies a detection bounding box as correct if its IoU between the ground truth bounding box exceeds 50\% ($AP_{0.5}$). The predicted bounding boxes are obtained using the output energy thresholded at 0.5. The closing operation with a kernel size of 7 is performed on the energy for noise removal. The ground truth bounding boxes are obtained using the manually annotated hand energy heatmaps.\\
\indent We compare the models' performance in the RGB domain, with/without the edge map in the color invariant domain and with/without the additional output energy channel as an ablation study to justify our design choices. As it is impossible for static pretrained models to produce highly accurate results in all scenes, we also perform experiments to study the impact of domain adaptation. To support scene-specific adaptation, we include a collected background sequence ($\sim$30 seconds) for each evaluation sequence where the scene-specific background images can be used to composite training instances. The background collection process simulates an environment scanning procedure using prospective egocentric color-based hand tracking devices. \\
\indent The following state-of-the-art architectures are selected for evaluation: UNet and $\text{UNet}_{1/8}$ (1/8 of the original network width) \cite{Ronneberger}, RecUNet \cite{WWang}, SegNet \cite{Badrinarayanan}, ICNet \cite{Zhao}, DeepLab V3+ \cite{Chen} and RefineNet \cite{GLin}. 
\begin{table*}[t]
  \small
  \centering
  \begin{tabular}{P{2.5cm}|c|c|c|c}
  \toprule
Model 	&
\#Params &
\begin{tabular}{@{}c@{}}Inference \\ time (ms)\end{tabular}&
		\begin{tabular}{@{}c@{}}Pretrained\\\hline
		\begin{tabular}{P{1.1cm}|P{1.2cm}|P{1.4cm}|P{2.4cm}}
                \begin{tabular}{@{}c@{}}RGB \\ energy \xmark\end{tabular}&
		\begin{tabular}{@{}c@{}}edge \xmark \\ energy \xmark\end{tabular}&\begin{tabular}{@{}c@{}}edge \cmark \\energy \xmark\end{tabular}&\begin{tabular}{@{}c@{}}w/ edge \& energy\\\begin{tabular}{P{1.0cm}P{1.0cm}}mIoU & $AP_{0.5}$\end{tabular}\end{tabular}
		\end{tabular}
		\end{tabular}&
                     \begin{tabular}{@{}c@{}}Adapted\\\hline
		\begin{tabular}{P{2.2cm}}
		\begin{tabular}{@{}c@{}}w/ edge \& energy\\\begin{tabular}{P{0.8cm}P{0.8cm}}mIoU & $AP_{0.5}$\end{tabular}\end{tabular}
		\end{tabular}
		\end{tabular}\\
\midrule
$\text{UNet}_{1/8}$ \cite{Ronneberger}&0.2M&9.5&\begin{tabular}{P{1.075cm}|P{1.2cm}|P{1.4cm}|P{1.1cm}|P{0.85cm}}0.405&0.722&0.749&0.754&0.633\end{tabular}&\begin{tabular}{P{1cm}|P{0.9cm}}0.844&0.739\end{tabular}\\
RecUNet \cite{WWang}&1.1M&78.1&\begin{tabular}{P{1.075cm}|P{1.2cm}|P{1.4cm}|P{1.1cm}|P{0.85cm}}0.615&0.812&0.834&\textbf{0.844}&0.805\end{tabular}&\begin{tabular}{P{1cm}|P{0.9cm}}0.874&0.839\end{tabular}\\
UNet \cite{Ronneberger}&13.4M&10.3&\begin{tabular}{P{1.075cm}|P{1.2cm}|P{1.4cm}|P{1.1cm}|P{0.85cm}}0.464&0.652&0.631&0.651&0.655\end{tabular}&\begin{tabular}{P{1cm}|P{0.9cm}}0.775&0.737\end{tabular}\\
ICNet \cite{Zhao} &28.3M&43.1&\begin{tabular}{P{1.075cm}|P{1.2cm}|P{1.4cm}|P{1.1cm}|P{0.85cm}}0.792&0.828&0.824&0.823&\textbf{0.886}\end{tabular}&\begin{tabular}{P{1cm}|P{0.9cm}}\textbf{0.885}&\textbf{0.945}\end{tabular}\\
SegNet \cite{Badrinarayanan} &29.4M&12.3&\begin{tabular}{P{1.075cm}|P{1.2cm}|P{1.4cm}|P{1.1cm}|P{0.85cm}}0.536&0.687&0.668&0.645&0.670\end{tabular}&\begin{tabular}{P{1cm}|P{0.9cm}}0.789&0.787\end{tabular}\\
DeepLabV3+* \cite{Chen} &59.3M&42.2&\begin{tabular}{P{1.075cm}|P{1.2cm}|P{1.4cm}|P{1.1cm}|P{0.85cm}}0.617&0.729&0.777&0.777&0.821\end{tabular}&\begin{tabular}{P{1cm}|P{0.9cm}}0.866&0.906\end{tabular}\\
RefineNet* \cite{GLin} &113.9M&50.5&\begin{tabular}{P{1.075cm}|P{1.2cm}|P{1.4cm}|P{1.1cm}|P{0.85cm}}0.650&0.825&0.847&0.836&0.874\end{tabular}&\begin{tabular}{P{1cm}|P{0.9cm}}0.884&0.903\end{tabular}\\
\end{tabular}
\caption{Evaluation of state-of-the-art models on Ego2Hands. Only mIoU is reported for experiments without the energy output channel (denoted as "energy \xmark"). Models with * are trained using pretrained Resnet encoders.}
\label{tab:table_comparison}
\vspace{-4mm}
\end{table*}
\subsection{Training Details}
\indent Experiments are divided into the pretraining and the adaptation phase. In the pretraining phase, models are trained using random background images for 100k iterations with a batch size of 4 and an initial learning rate of $\num{1.e-4}$ decreased with a ratio of 0.5 every 20k iterations. In the adaptation phase, we finetune the pretrained models (with input edge map and output energy channel) for 10k iterations using scene-specific background images and an initial learning rate of $\num{1.e-5}$ decaying with the same ratio every 5k iterations. We use the Adam optimizer and find general convergence from all models using this setup. Averaged results from 3 trained model instances are reported for each experiment. GeForce RTX 2080 Ti is used as the GPU in our experiments.\\
\indent As illumination (not skin tone) is the dominant factor for the brightness of the hands in input images and is known for specific environments, we perform brightness augmentation within ranges specific to the scenes in the adaptation phase. The mean brightness value $\beta$ of the composited hands is scaled to be in the range [0, 55], [55,200], [55, 255] for scenes with dark (seq5), normal (seq1, 3, 4, 6, 7) and bright (seq2, 8) illumination respectively. Bright scenes have a wider brightness range due to the possible presence of shadow. $\beta$ for background images is jittered by $\pm 50$.
\subsection{Quantitative Analysis}\label{sec:quantitative_analysis}
\indent Table \ref{tab:table_comparison} provides detailed quantitative results for the selected models on different settings. We point out that a comprehensive comparison involves various factors including model size, inference speed, segmentation/detection accuracy, the ability to generalize and adapt. \\
\indent First, we perform experiments using the RGB input domain and compare with our proposed color-invariant domain. To generate more realistic instances in the color domain, we generate a version of $\text{Ego2Hands}_{train}$ with the green screen color spill problem resolved using the Keylight feature of Adobe After Effects. Note that this issue is largely circumvented in our color-invariant domain. During training, we specifically color-augment the composited data according to the color distribution of $\text{Ego2Hands}_{test}$ shown in Fig. \ref{fig:data-dist}. Results show that training in the RGB space does not outperform the color-invariant domain on $\text{Ego2Hands}_{test}$. In addition, we justify our design choice of the additional input edge map and energy output channel. Models with an input edge map and energy output show overall improvement in segmentation while additionally estimating the output energy, which provides hand detection information essential for many applications.\\
\indent In our experiments, UNet and SegNet underperform with lower generalization and adaptation accuracy. With a small number of parameters, $\text{UNet}_{1/8}$ and RecUNet achieve good segmentation accuracy but suffer from lower detection accuracy. Additionally, the recursion on internal network states improves accuracy while notably increasing the inference time, making RecUNet the slowest models in our analysis. It is worth mentioning that inference speed is crucial in real-world applications as there are oftentimes subsequent pose estimation/gesture recognition modules. Heavy models (DeepLabV3+, RefineNet) generally achieve high pretrained and adapted accuracy in both segmentation and detection. We find that heavy models are dependent on pretrained encoders for optimal performance. \\
\begin{figure*}
  \centering
  \begin{subfigure}[t]{0.19\linewidth}
    \includegraphics[width=\linewidth]{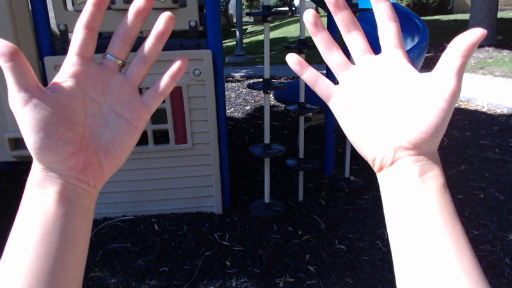}
  \end{subfigure}
  \begin{subfigure}[t]{0.19\linewidth}
    \includegraphics[width=\linewidth]{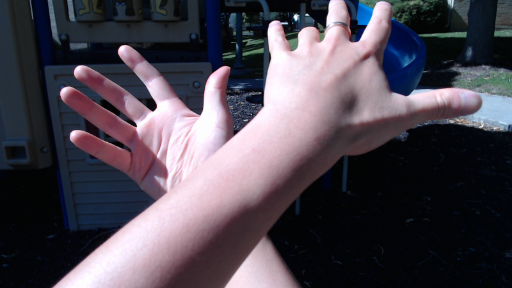}
  \end{subfigure}
  \begin{subfigure}[t]{0.19\linewidth}
    \includegraphics[width=\linewidth]{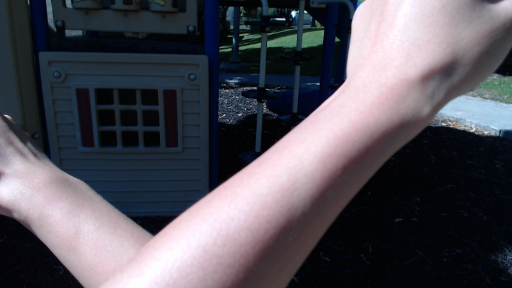}
  \end{subfigure}
  \begin{subfigure}[t]{0.19\linewidth}
    \includegraphics[width=\linewidth]{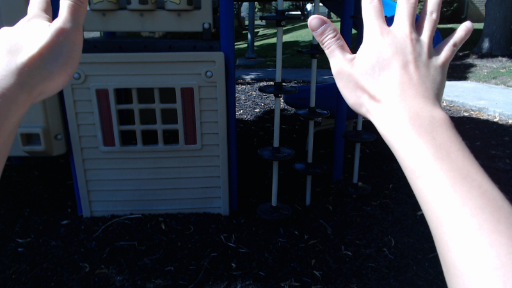}
  \end{subfigure}
  \begin{subfigure}[t]{0.19\linewidth}
    \includegraphics[width=\linewidth]{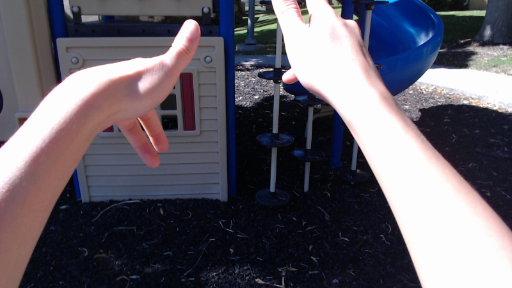}
  \end{subfigure}
  \begin{subfigure}[t]{0.19\linewidth}
    \includegraphics[width=\linewidth]{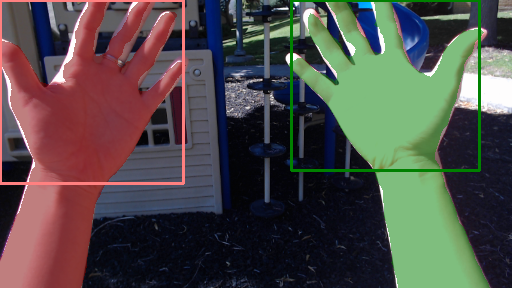}
  \end{subfigure}
  \begin{subfigure}[t]{0.19\linewidth}
    \includegraphics[width=\linewidth]{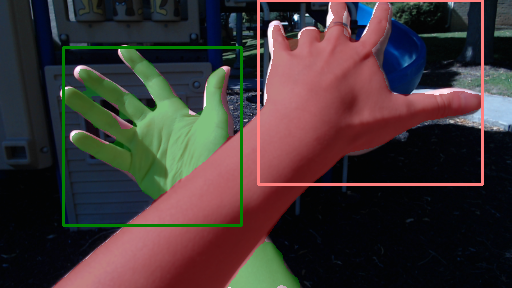}
  \end{subfigure}
  \begin{subfigure}[t]{0.19\linewidth}
    \includegraphics[width=\linewidth]{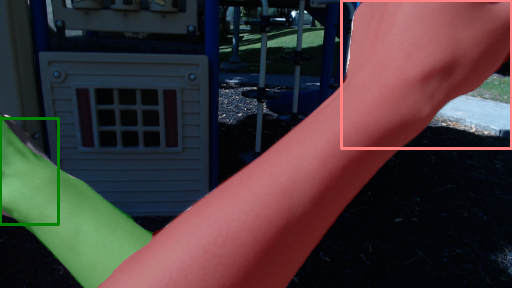}
  \end{subfigure}
  \begin{subfigure}[t]{0.19\linewidth}
    \includegraphics[width=\linewidth]{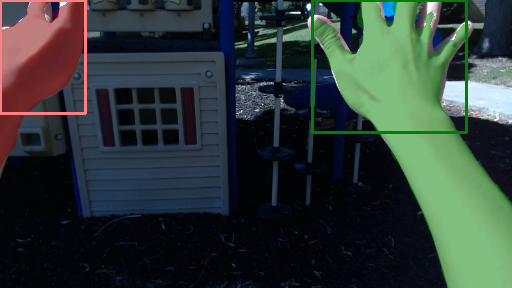}
  \end{subfigure}
  \vspace{1mm}
  \begin{subfigure}[t]{0.19\linewidth}
    \includegraphics[width=\linewidth]{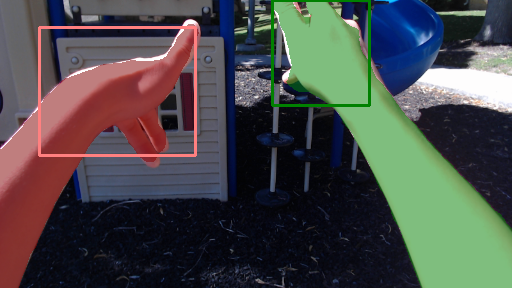}
  \end{subfigure}
  \begin{subfigure}[t]{0.19\linewidth}
    \includegraphics[width=\linewidth]{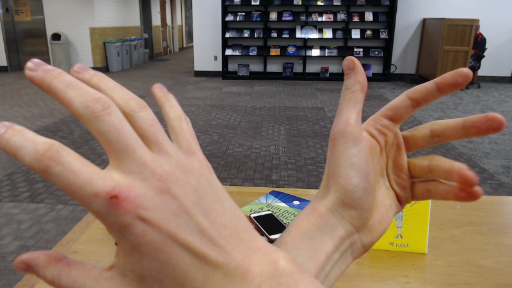}
  \end{subfigure}
  \begin{subfigure}[t]{0.19\linewidth}
    \includegraphics[width=\linewidth]{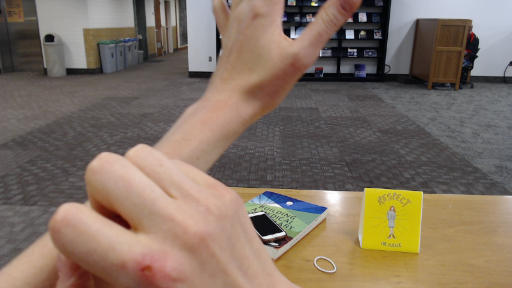}
  \end{subfigure}
  \begin{subfigure}[t]{0.19\linewidth}
    \includegraphics[width=\linewidth]{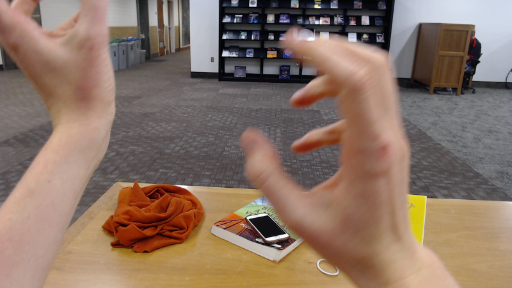}
  \end{subfigure}
  \begin{subfigure}[t]{0.19\linewidth}
    \includegraphics[width=\linewidth]{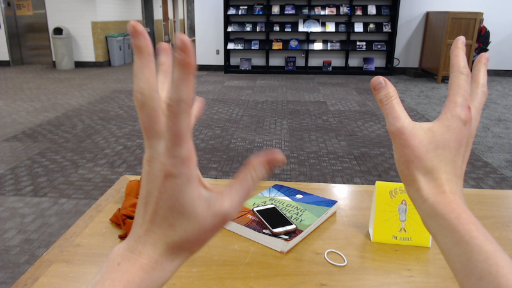}
  \end{subfigure}
  \begin{subfigure}[t]{0.19\linewidth}
    \includegraphics[width=\linewidth]{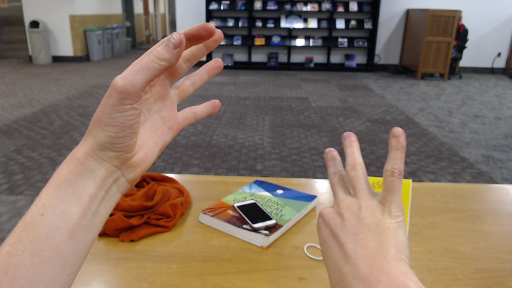}
  \end{subfigure}
  \begin{subfigure}[t]{0.19\linewidth}
    \includegraphics[width=\linewidth]{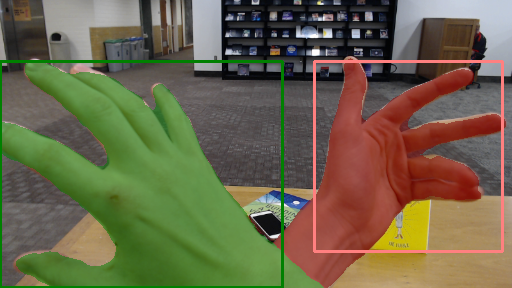}
  \end{subfigure}
  \begin{subfigure}[t]{0.19\linewidth}
    \includegraphics[width=\linewidth]{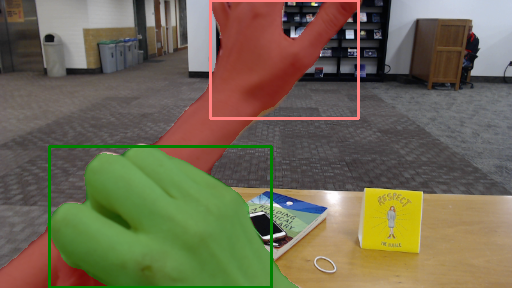}
  \end{subfigure}
  \begin{subfigure}[t]{0.19\linewidth}
    \includegraphics[width=\linewidth]{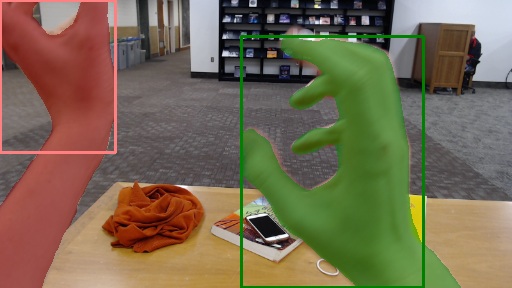}
  \end{subfigure}
  \begin{subfigure}[t]{0.19\linewidth}
    \includegraphics[width=\linewidth]{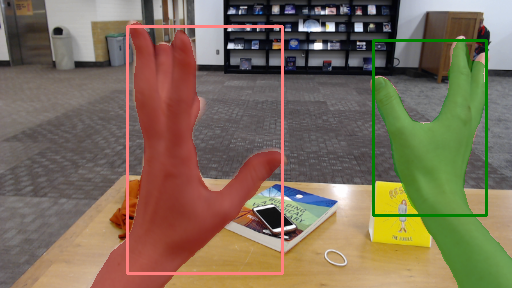}
  \end{subfigure}
  \vspace{1mm}
  \begin{subfigure}[t]{0.19\linewidth}
    \includegraphics[width=\linewidth]{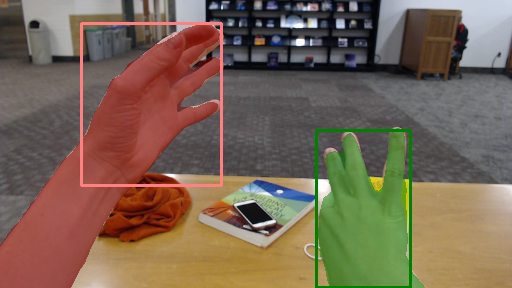}
  \end{subfigure}
  \begin{subfigure}[t]{0.19\linewidth}
    \includegraphics[width=\linewidth]{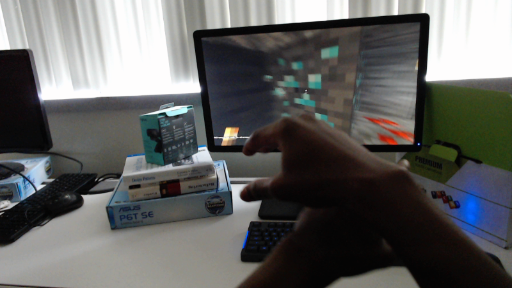}
  \end{subfigure}
  \begin{subfigure}[t]{0.19\linewidth}
    \includegraphics[width=\linewidth]{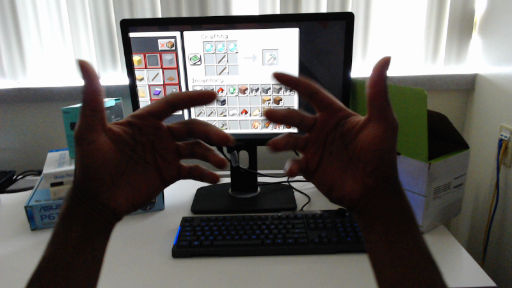}
  \end{subfigure}
  \begin{subfigure}[t]{0.19\linewidth}
    \includegraphics[width=\linewidth]{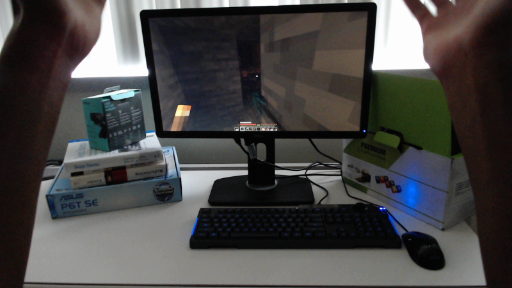}
  \end{subfigure}
  \begin{subfigure}[t]{0.19\linewidth}
    \includegraphics[width=\linewidth]{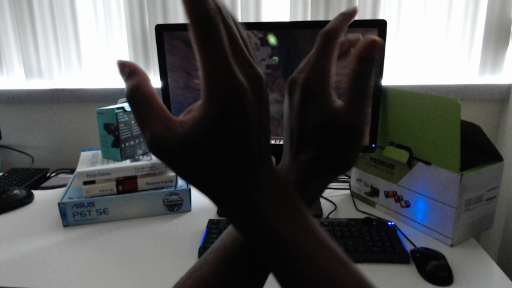}
  \end{subfigure}
  \begin{subfigure}[t]{0.19\linewidth}
    \includegraphics[width=\linewidth]{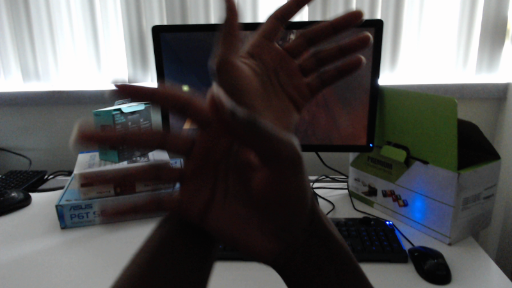}
  \end{subfigure}
  \begin{subfigure}[t]{0.19\linewidth}
    \includegraphics[width=\linewidth]{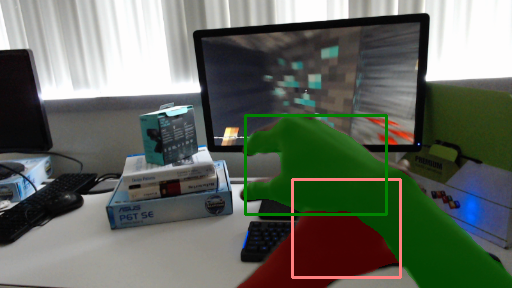}
  \end{subfigure}
  \begin{subfigure}[t]{0.19\linewidth}
    \includegraphics[width=\linewidth]{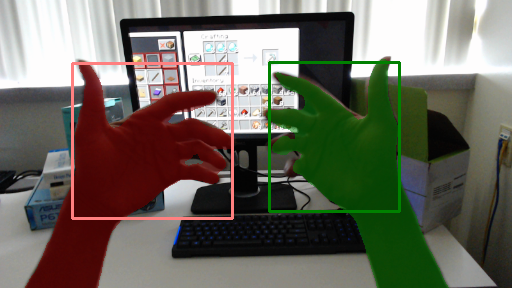}
  \end{subfigure}
  \begin{subfigure}[t]{0.19\linewidth}
    \includegraphics[width=\linewidth]{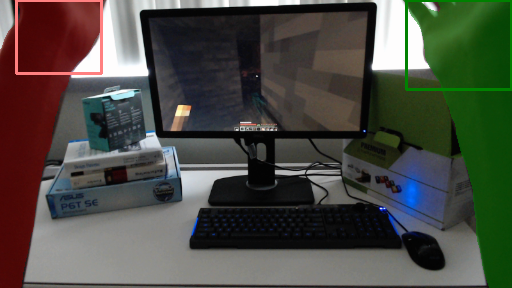}
  \end{subfigure}
  \begin{subfigure}[t]{0.19\linewidth}
    \includegraphics[width=\linewidth]{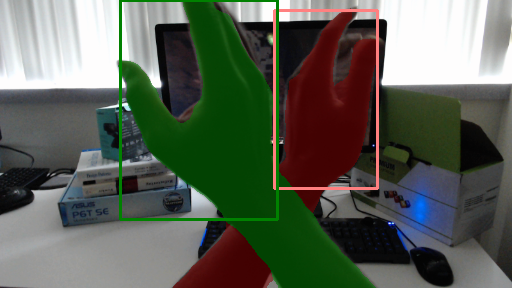}
  \end{subfigure}
  \begin{subfigure}[t]{0.19\linewidth}
    \includegraphics[width=\linewidth]{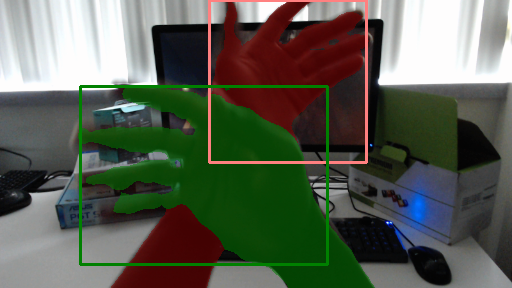}
  \end{subfigure}
  \caption{Qualitative results obtained using scene-adapted ICNet. Odd rows show sample images in evaluation sequences with various skin tones and illumination. Even rows show the output visualization. \textbf{Best viewed in magnification.}}
  \label{fig:qualitative_results}
  \vspace{-4mm}
\end{figure*}
\indent Our experiments provide valuable insights on the trade-off between architectures. In general, ICNet produces excellent pretrained and adapted accuracy for both segmentation and detection with medium model size and inference speed. In cases where the model size is not a limiting factor, RefineNet achieves high generalization accuracy, which we leverage for cross-dataset evaluation. In memory-constrained settings, $\text{UNet}_{1/8}$ has extremely compact model size and achieves promising accuracy, which RecUNet further improves by sacrificing inference speed. In time-sensitive applications, shallow models such as $\text{UNet}_{1/8}$, UNet and SegNet have faster inference speed with lower accuracy. Interestingly, by simply reducing the network width, $\text{UNet}_{1/8}$ outperforms $\text{UNet}$ in nearly every aspect for the task of two-hand segmentation and detection.\\
\indent We reemphasize that the evaluation sequences cover various ranges of illumination and include hands (various skin tones) and scenes not present in the training set of Ego2Hands. Our quantitative results show that the proposed dataset and data composition technique enable models to generalize to the real-world image domain. To provide a proper perspective for our significantly increased level of generalization, as \cite{Cai} recently tried to address the problem of domain adaptation in a specific unseen environment for binary-label hand segmentation, we enable models to achieve high accuracy on two-hand segmentation and detection in a domain-invariant setting with the option to further improve using scene-specific adaptation. We show qualitative results in Fig. \ref{fig:qualitative_results} and demonstrate real-time gesture control using fast and accurate two-hand segmentation/detection estimation in the supplementary video.
\section{Future Work}
\indent Hand segmentation with objects can be useful in applications that involve object handling. However, the introduction of objects in green screen data collection for hand segmentation requires an additional step to segment the objects, making automatic large-scale annotation of segmentation data challenging. Although a naive composition of objects in hand images is achievable, realistic handling of the objects requires additional hand pose information. Since many applications such as hand tracking in VR/AR, gesture/sign language recognition focus more on bare hand scenarios, we leave object interaction as future work.\\
\indent Although our dataset is the first real-world hand segmentation dataset to address inter-hand occlusion, it does not contain two-hand segmentation data for close hand interactions involving interlaced fingers. This is extremely challenging due to the difficulty in obtaining the corresponding segmentation ground truth. In general, accurate two-hand segmentation will be very beneficial for two-hand pose estimation and gesture recognition applications.
\section{Conclusion}
\indent In this work, we introduce a large-scale RGB hand segmentation dataset with a compositing-based data generation approach for two-hand segmentation and detection in unconstrained settings. Cross-dataset comparisons show that our dataset exceeds existing datasets significantly in quantity, diversity, annotation quality and contains two-hand labeling that addresses inter-hand occlusion. Validation and analysis of state-of-the-art models on our benchmark dataset show that training on Ego2Hands enables high-quality generalization to unseen scenes and scene-specific adaptation can achieve further improvements. 

{\small
\bibliographystyle{ieee_fullname}
\bibliography{egbib}
}

\clearpage
\onecolumn

\begin{center}
       {\Large\textbf{Supplementary Document:\\
Ego2Hands: A Dataset for Egocentric Two-hand Segmentation and Detection}}

       \vspace{0.5cm}
\end{center}
\setlength{\fboxsep}{0pt}
\pagenumbering{gobble}
\setcounter{section}{0}
\setcounter{figure}{0}
\setcounter{table}{0}
\section{Ego2Hands Qualitative Examples}
\subsection{Training set} 
\indent We show sample collected images with the corresponding hand heatmap energy for the detection task. The energy map annotates the hand (without the arm) as the foreground and all other regions as the background. Our training data covers a wide range of hand locations, hand poses, hand sizes, skin tones and illuminations.\\
\begin{figure*}[h]
  \centering
  \fbox{\begin{subfigure}[t]{0.155\linewidth}
    \includegraphics[width=\linewidth]{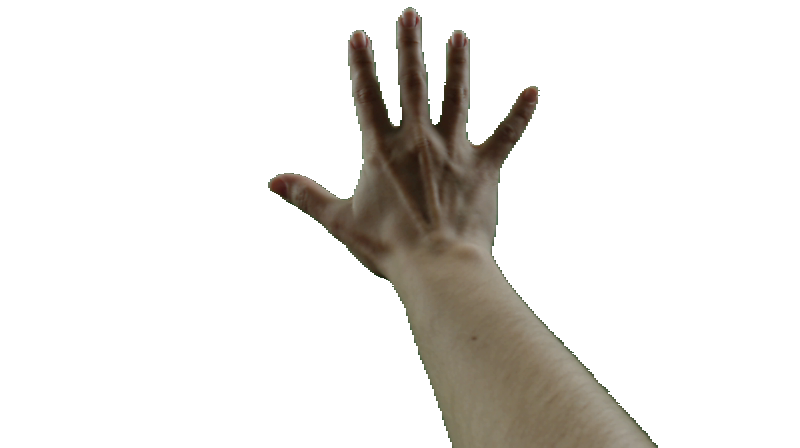}
  \end{subfigure}}
  \fbox{\begin{subfigure}[t]{0.155\linewidth}
    \includegraphics[width=\linewidth]{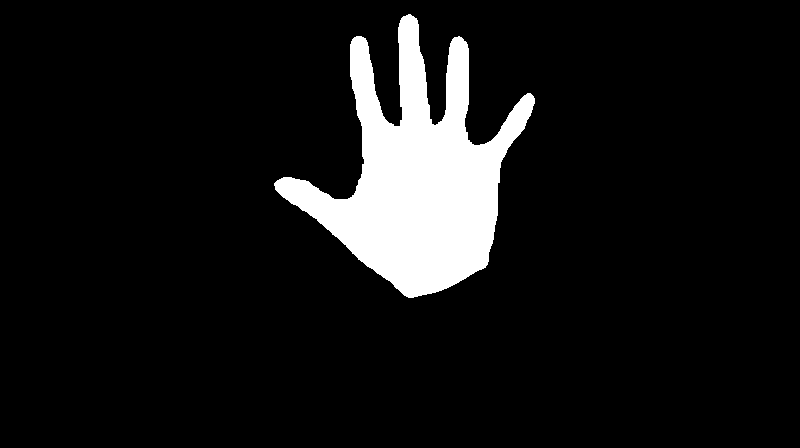}
  \end{subfigure}}
  \fbox{\begin{subfigure}[t]{0.155\linewidth}
    \includegraphics[width=\linewidth]{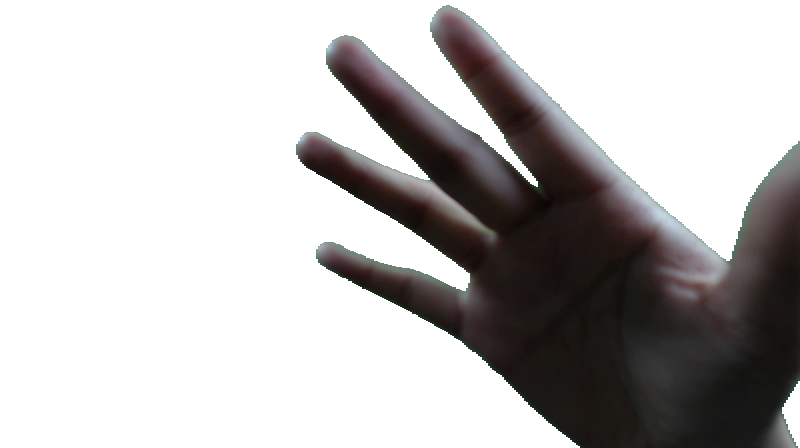}
  \end{subfigure}}
  \fbox{\begin{subfigure}[t]{0.155\linewidth}
    \includegraphics[width=\linewidth]{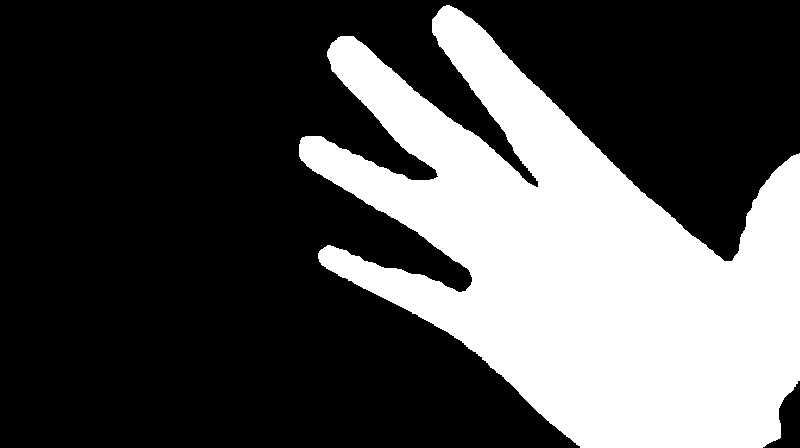}
  \end{subfigure}}
  \fbox{\begin{subfigure}[t]{0.155\linewidth}
    \includegraphics[width=\linewidth]{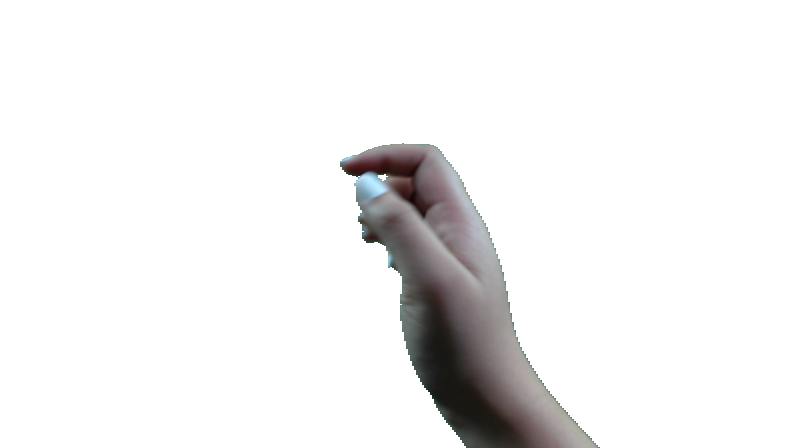}
  \end{subfigure}}
  \vspace{1mm}
  \fbox{\begin{subfigure}[t]{0.155\linewidth}
    \includegraphics[width=\linewidth]{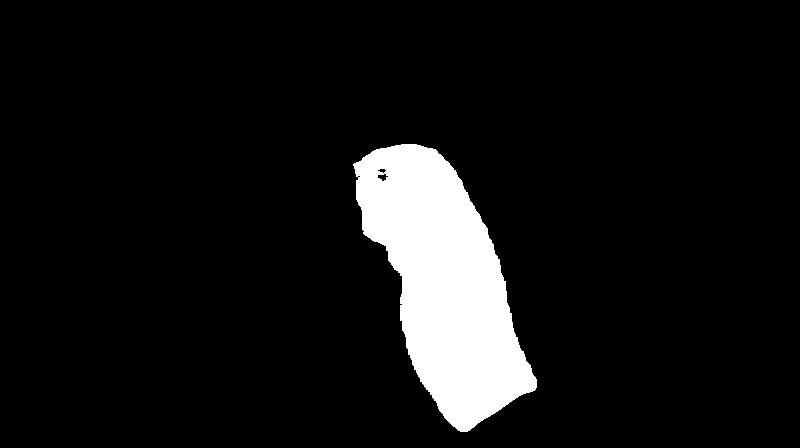}
  \end{subfigure}}
  \fbox{\begin{subfigure}[t]{0.155\linewidth}
    \includegraphics[width=\linewidth]{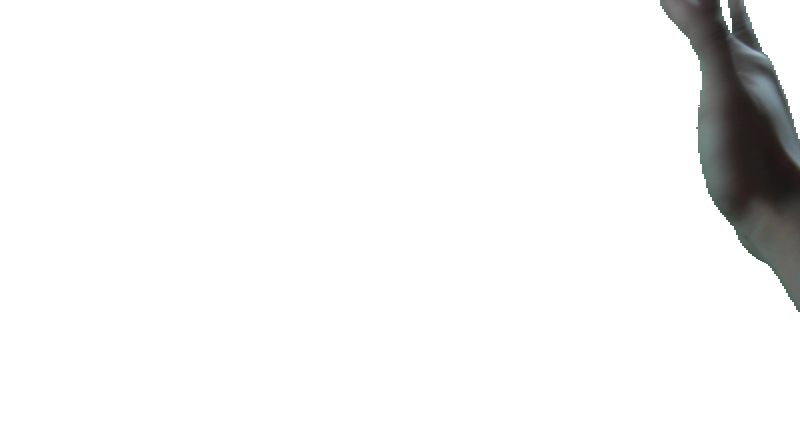}
  \end{subfigure}}
  \fbox{\begin{subfigure}[t]{0.155\linewidth}
    \includegraphics[width=\linewidth]{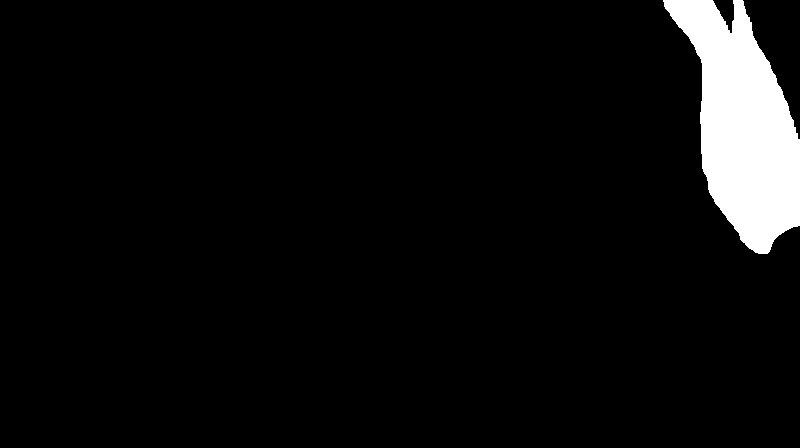}
  \end{subfigure}}
  \fbox{\begin{subfigure}[t]{0.155\linewidth}
    \includegraphics[width=\linewidth]{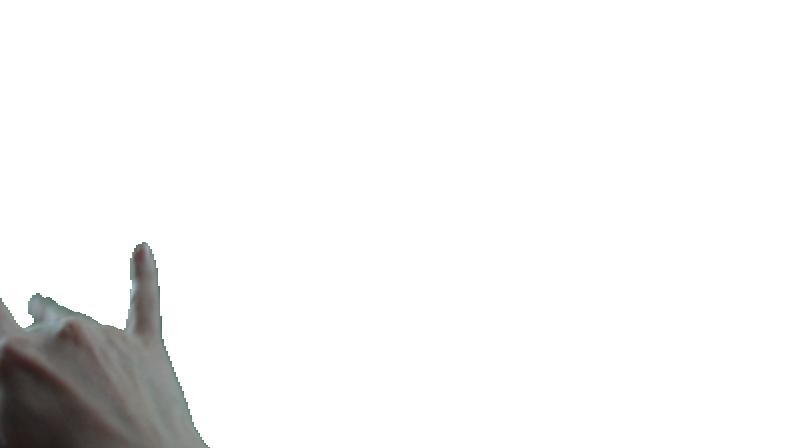}
  \end{subfigure}}
  \fbox{\begin{subfigure}[t]{0.155\linewidth}
    \includegraphics[width=\linewidth]{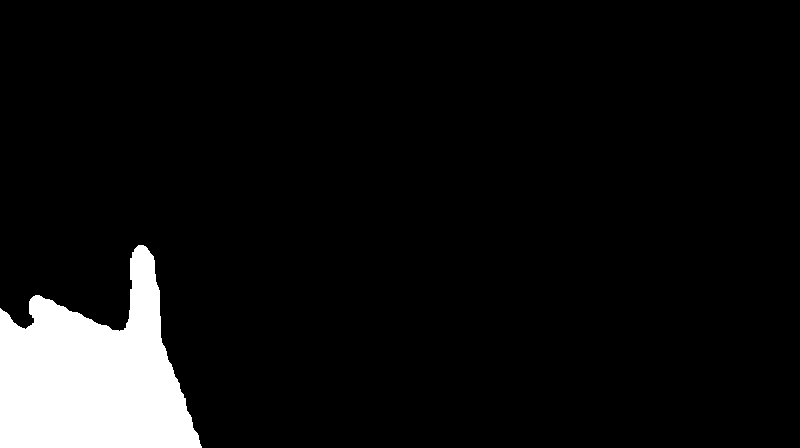}
  \end{subfigure}}
  \fbox{\begin{subfigure}[t]{0.155\linewidth}
    \includegraphics[width=\linewidth]{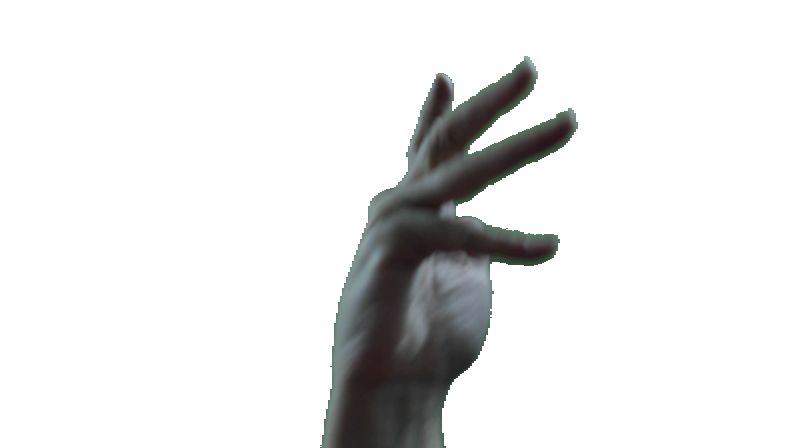}
  \end{subfigure}}
  \vspace{1mm}
  \fbox{\begin{subfigure}[t]{0.155\linewidth}
    \includegraphics[width=\linewidth]{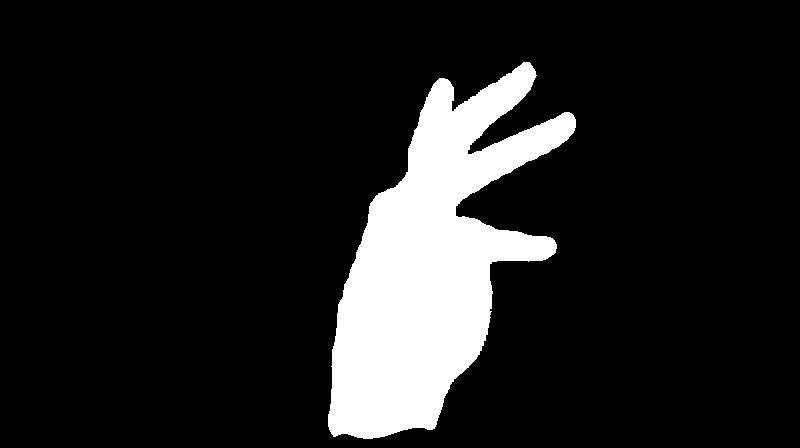}
  \end{subfigure}}
  \fbox{\begin{subfigure}[t]{0.155\linewidth}
    \includegraphics[width=\linewidth]{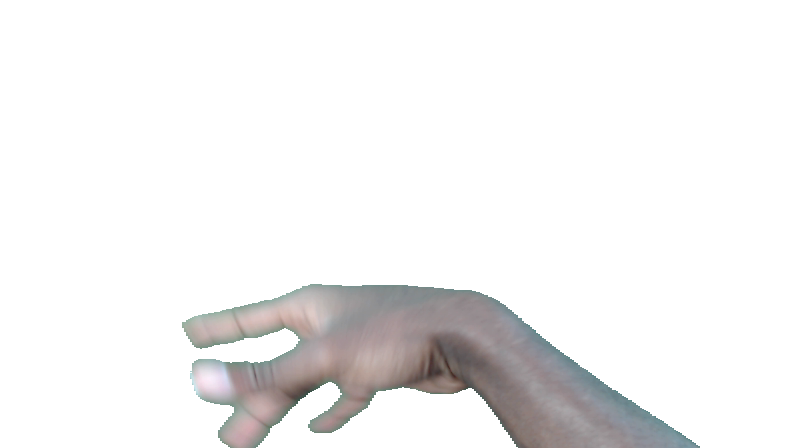}
  \end{subfigure}}
  \fbox{\begin{subfigure}[t]{0.155\linewidth}
    \includegraphics[width=\linewidth]{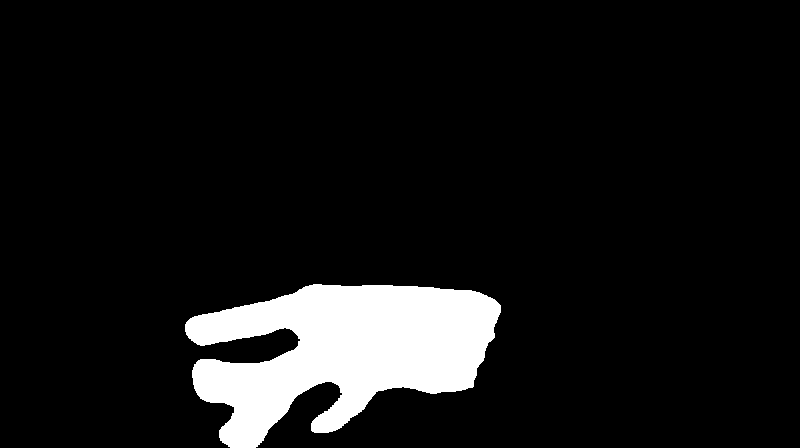}
  \end{subfigure}}
  \fbox{\begin{subfigure}[t]{0.155\linewidth}
    \includegraphics[width=\linewidth]{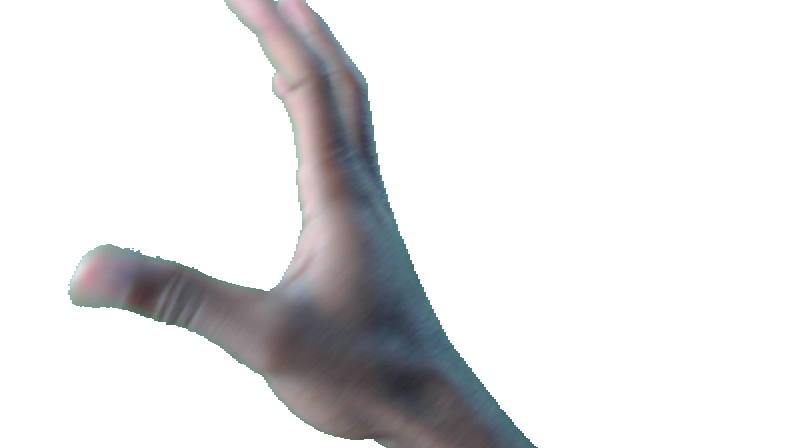}
  \end{subfigure}}
  \fbox{\begin{subfigure}[t]{0.155\linewidth}
    \includegraphics[width=\linewidth]{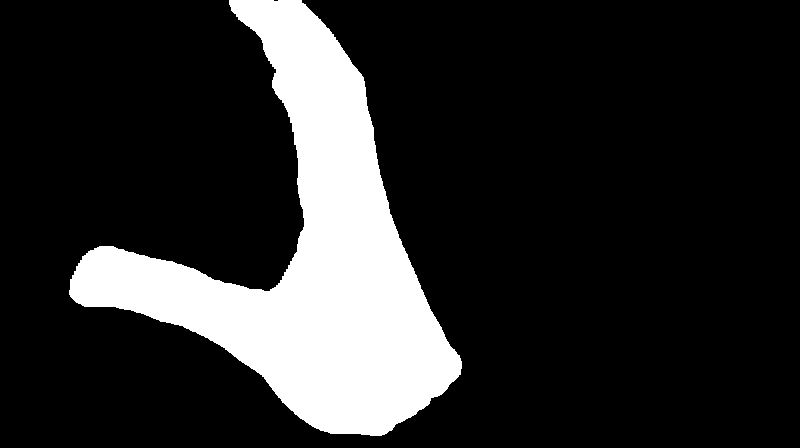}
  \end{subfigure}}
  \fbox{\begin{subfigure}[t]{0.155\linewidth}
    \includegraphics[width=\linewidth]{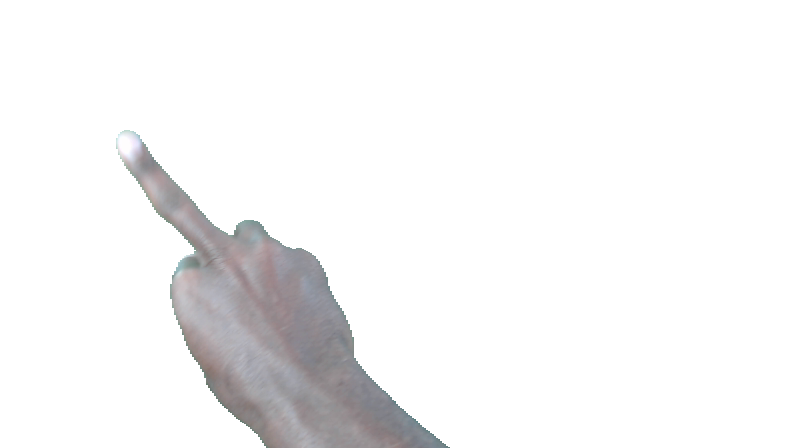}
  \end{subfigure}}
  \vspace{1mm}
  \fbox{\begin{subfigure}[t]{0.155\linewidth}
    \includegraphics[width=\linewidth]{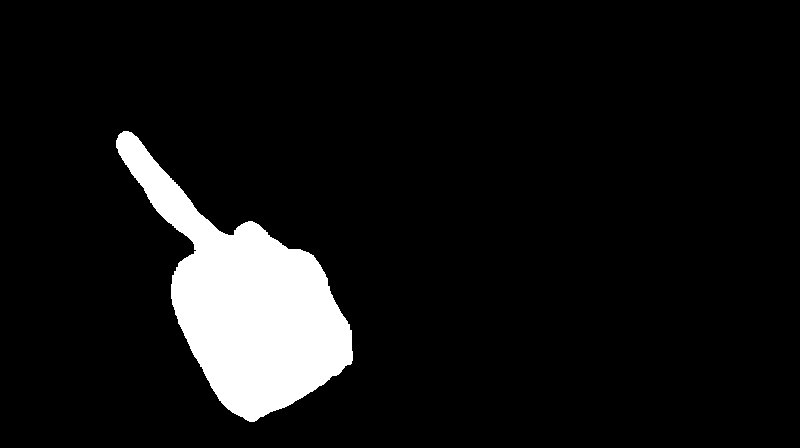}
  \end{subfigure}}
  \fbox{\begin{subfigure}[t]{0.155\linewidth}
    \includegraphics[width=\linewidth]{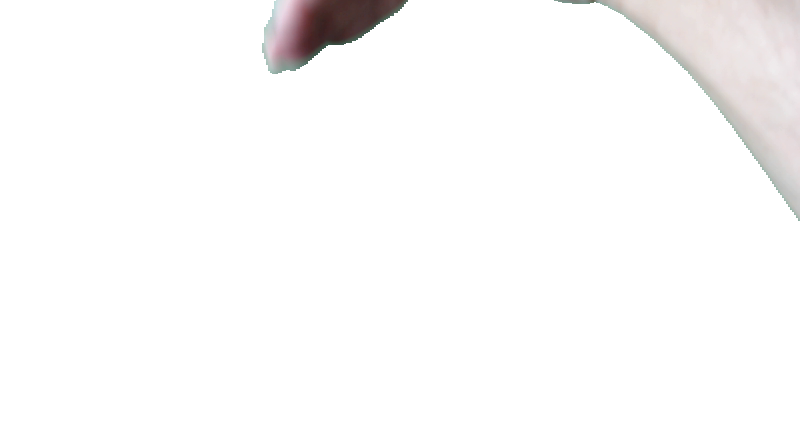}
  \end{subfigure}}
  \fbox{\begin{subfigure}[t]{0.155\linewidth}
    \includegraphics[width=\linewidth]{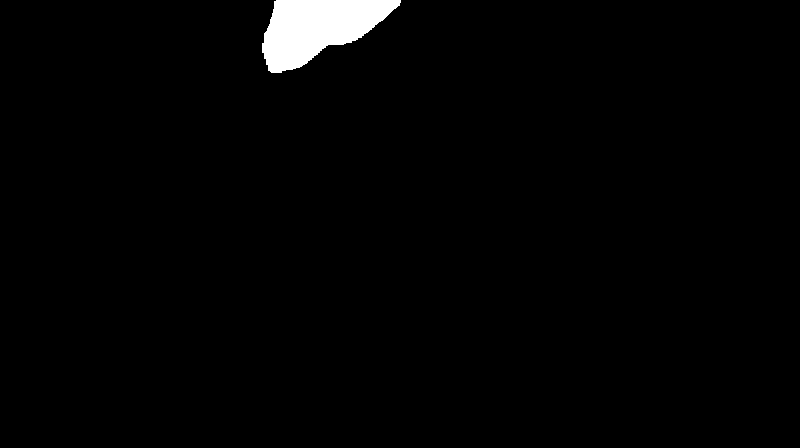}
  \end{subfigure}}
  \fbox{\begin{subfigure}[t]{0.155\linewidth}
    \includegraphics[width=\linewidth]{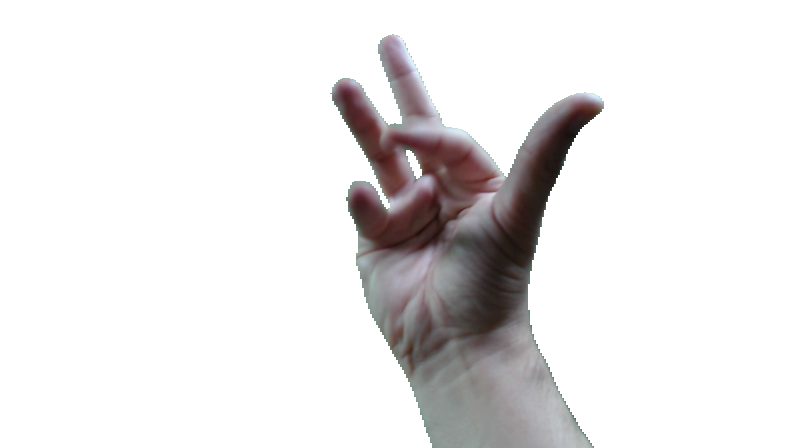}
  \end{subfigure}}
  \fbox{\begin{subfigure}[t]{0.155\linewidth}
    \includegraphics[width=\linewidth]{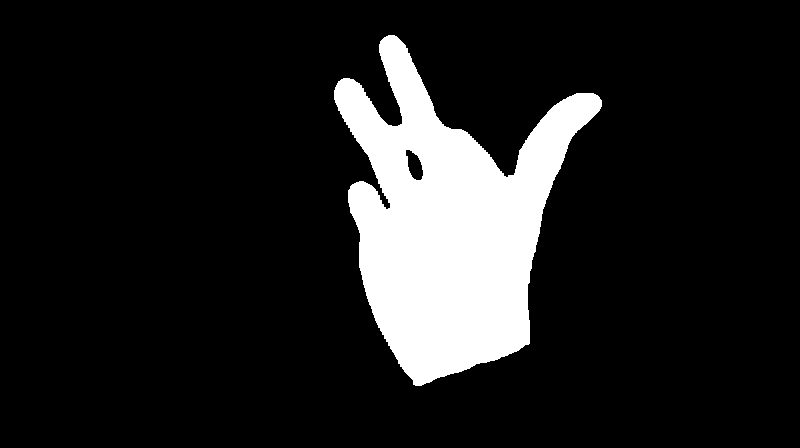}
  \end{subfigure}}
  \fbox{\begin{subfigure}[t]{0.155\linewidth}
    \includegraphics[width=\linewidth]{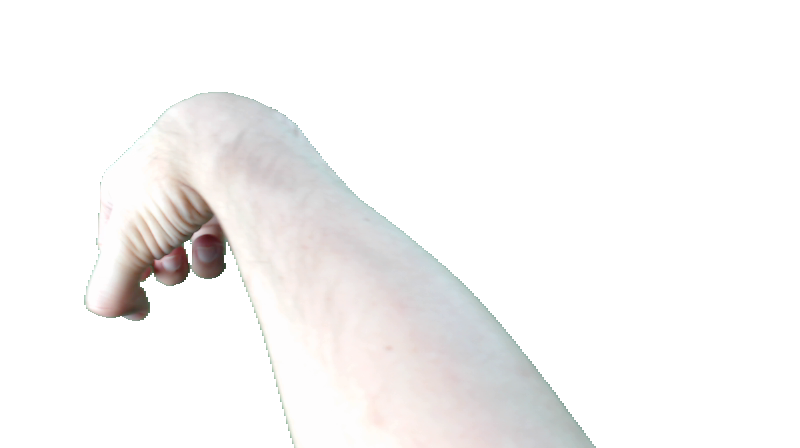}
  \end{subfigure}}
  \vspace{1mm}
  \fbox{\begin{subfigure}[t]{0.155\linewidth}
    \includegraphics[width=\linewidth]{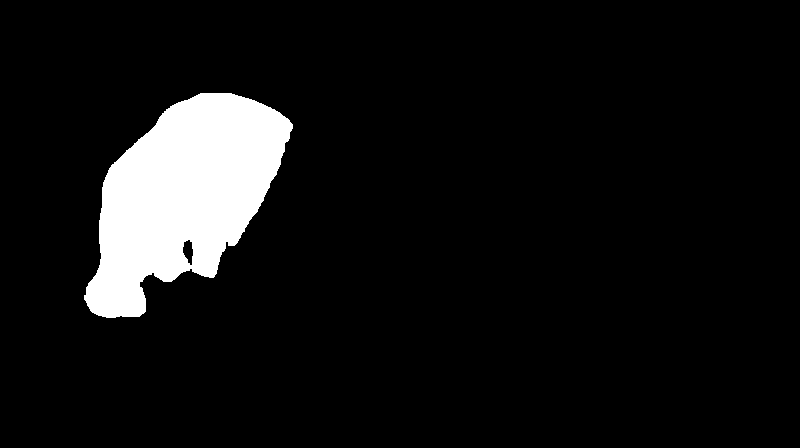}
  \end{subfigure}}
  \fbox{\begin{subfigure}[t]{0.155\linewidth}
    \includegraphics[width=\linewidth]{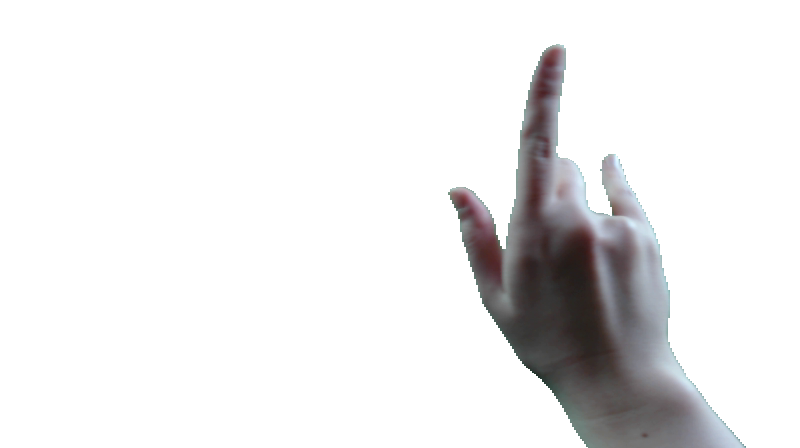}
  \end{subfigure}}
  \fbox{\begin{subfigure}[t]{0.155\linewidth}
    \includegraphics[width=\linewidth]{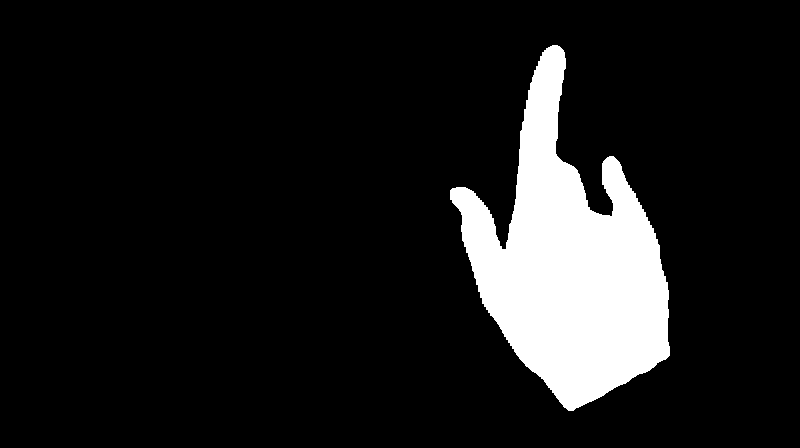}
  \end{subfigure}}
  \fbox{\begin{subfigure}[t]{0.155\linewidth}
    \includegraphics[width=\linewidth]{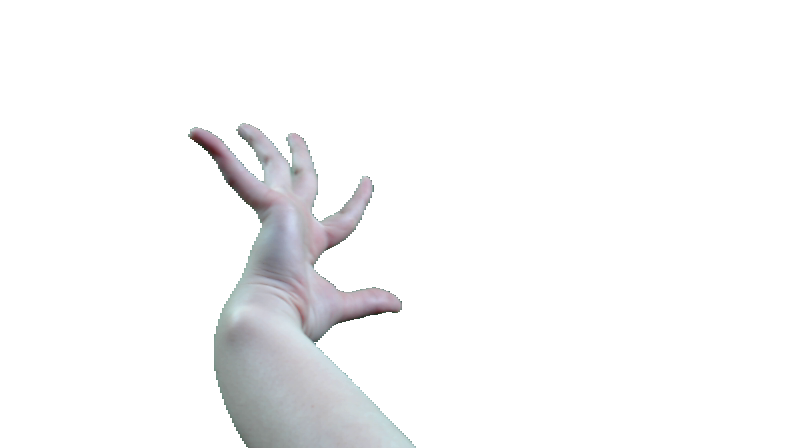}
  \end{subfigure}}
  \fbox{\begin{subfigure}[t]{0.155\linewidth}
    \includegraphics[width=\linewidth]{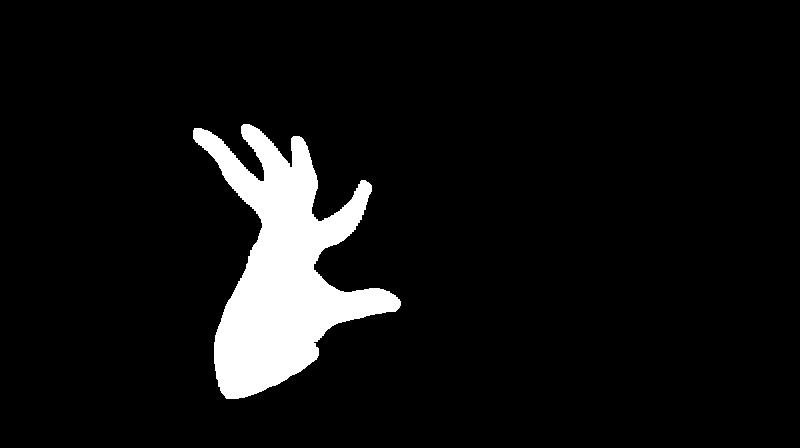}
  \end{subfigure}}
  \fbox{\begin{subfigure}[t]{0.155\linewidth}
    \includegraphics[width=\linewidth]{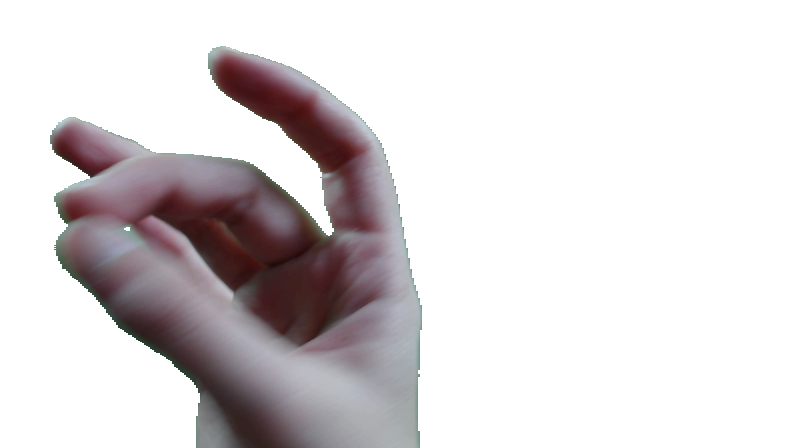}
  \end{subfigure}}
  \vspace{1mm}
  \fbox{\begin{subfigure}[t]{0.155\linewidth}
    \includegraphics[width=\linewidth]{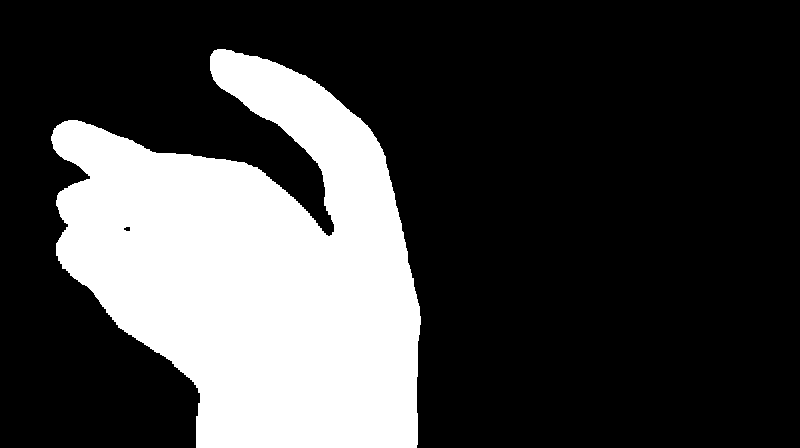}
  \end{subfigure}}
  \fbox{\begin{subfigure}[t]{0.155\linewidth}
    \includegraphics[width=\linewidth]{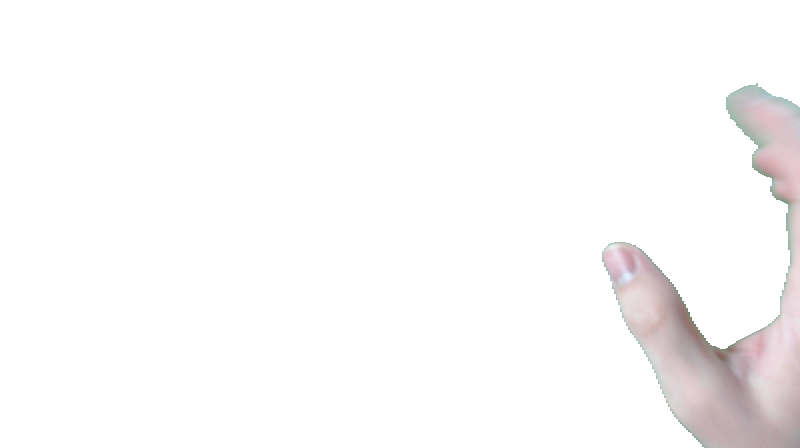}
  \end{subfigure}}
  \fbox{\begin{subfigure}[t]{0.155\linewidth}
    \includegraphics[width=\linewidth]{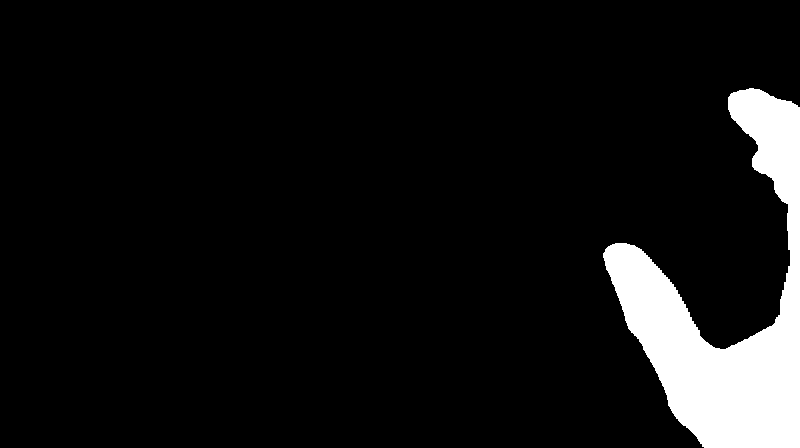}
  \end{subfigure}}
  \fbox{\begin{subfigure}[t]{0.155\linewidth}
    \includegraphics[width=\linewidth]{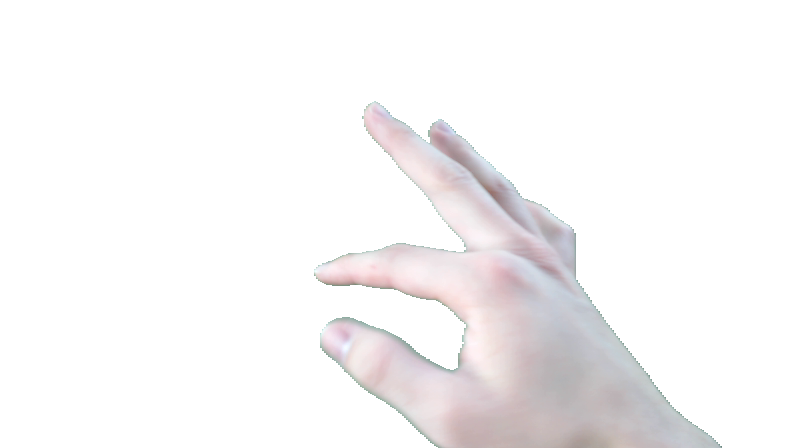}
  \end{subfigure}}
  \fbox{\begin{subfigure}[t]{0.155\linewidth}
    \includegraphics[width=\linewidth]{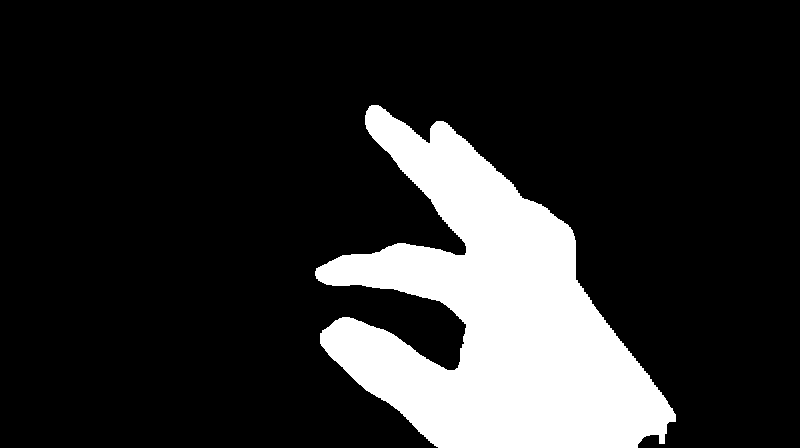}
  \end{subfigure}}
  \fbox{\begin{subfigure}[t]{0.155\linewidth}
    \includegraphics[width=\linewidth]{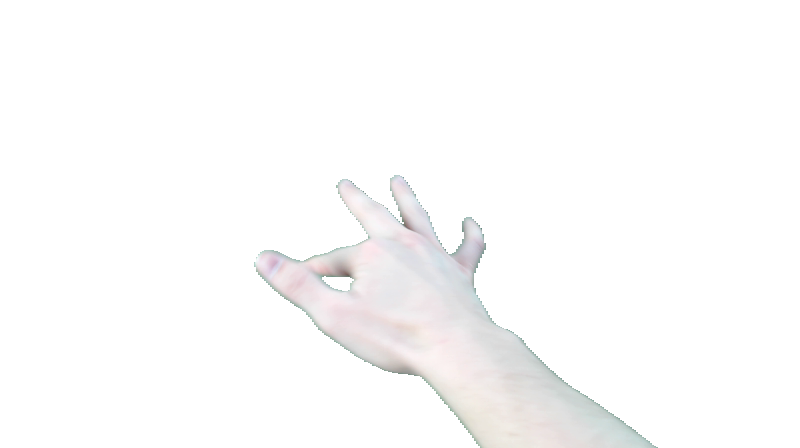}
  \end{subfigure}}
  \vspace{1mm}
  \fbox{\begin{subfigure}[t]{0.155\linewidth}
    \includegraphics[width=\linewidth]{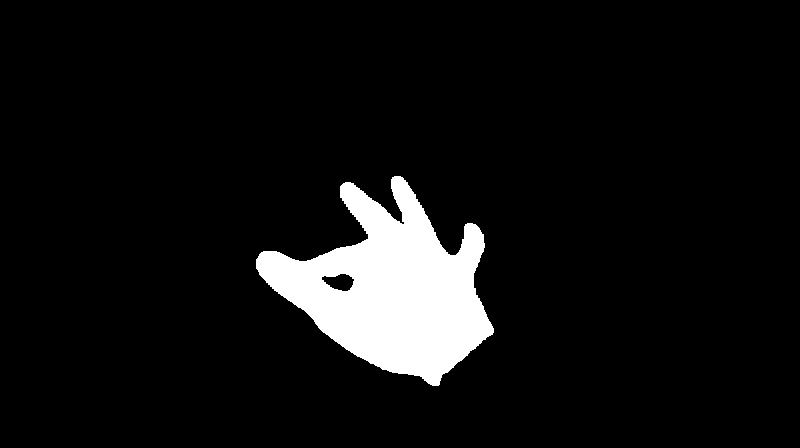}
  \end{subfigure}}
  \fbox{\begin{subfigure}[t]{0.155\linewidth}
    \includegraphics[width=\linewidth]{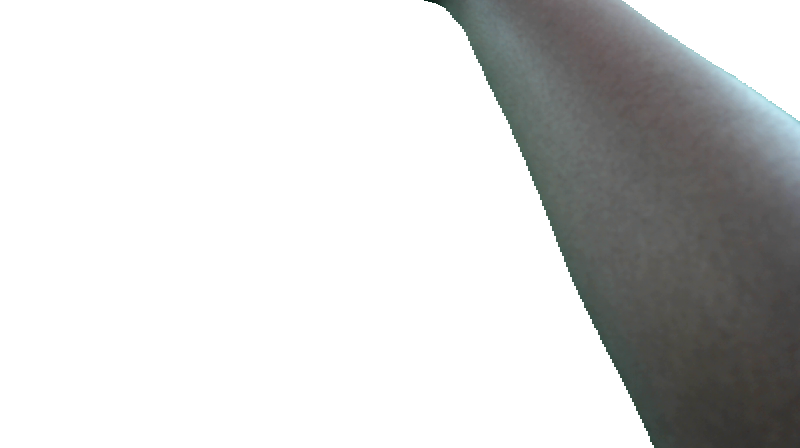}
  \end{subfigure}}
  \fbox{\begin{subfigure}[t]{0.155\linewidth}
    \includegraphics[width=\linewidth]{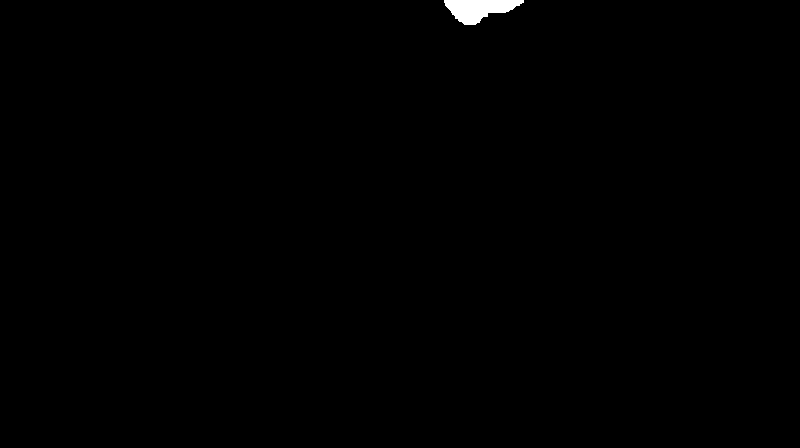}
  \end{subfigure}}
  \fbox{\begin{subfigure}[t]{0.155\linewidth}
    \includegraphics[width=\linewidth]{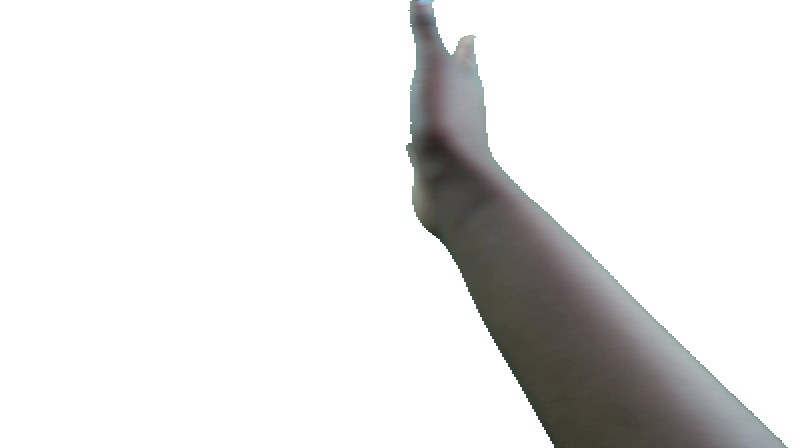}
  \end{subfigure}}
  \fbox{\begin{subfigure}[t]{0.155\linewidth}
    \includegraphics[width=\linewidth]{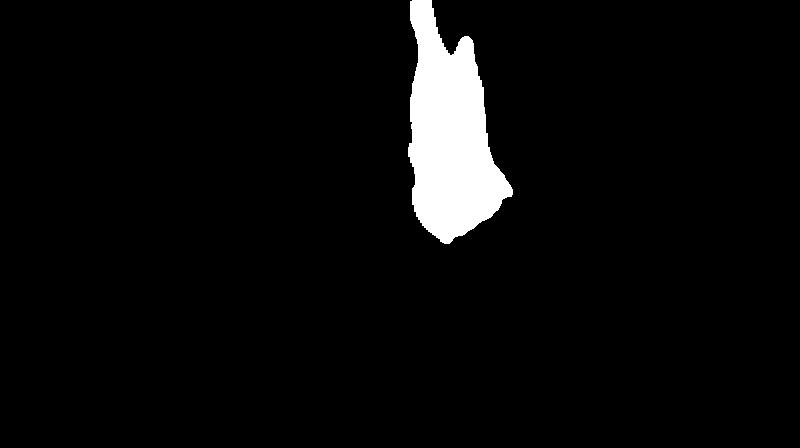}
  \end{subfigure}}
  \fbox{\begin{subfigure}[t]{0.155\linewidth}
    \includegraphics[width=\linewidth]{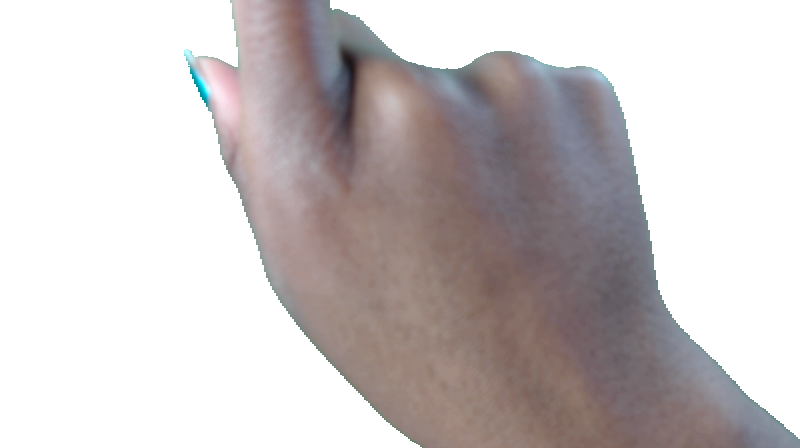}
  \end{subfigure}}
  \vspace{1mm}
  \fbox{\begin{subfigure}[t]{0.155\linewidth}
    \includegraphics[width=\linewidth]{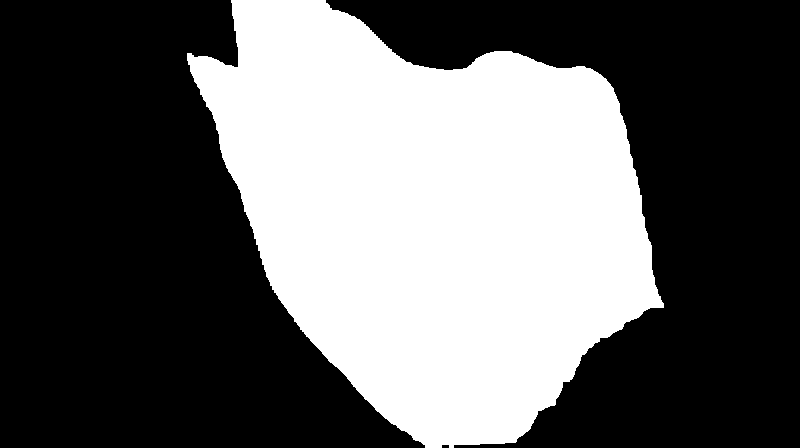}
  \end{subfigure}}
  \fbox{\begin{subfigure}[t]{0.155\linewidth}
    \includegraphics[width=\linewidth]{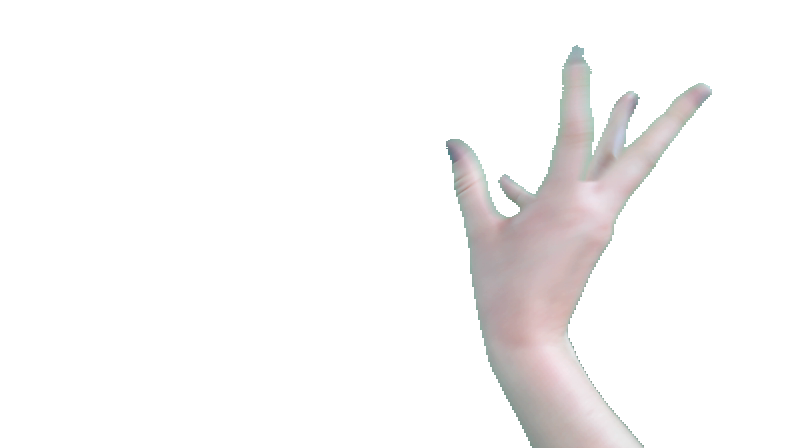}
  \end{subfigure}}
  \fbox{\begin{subfigure}[t]{0.155\linewidth}
    \includegraphics[width=\linewidth]{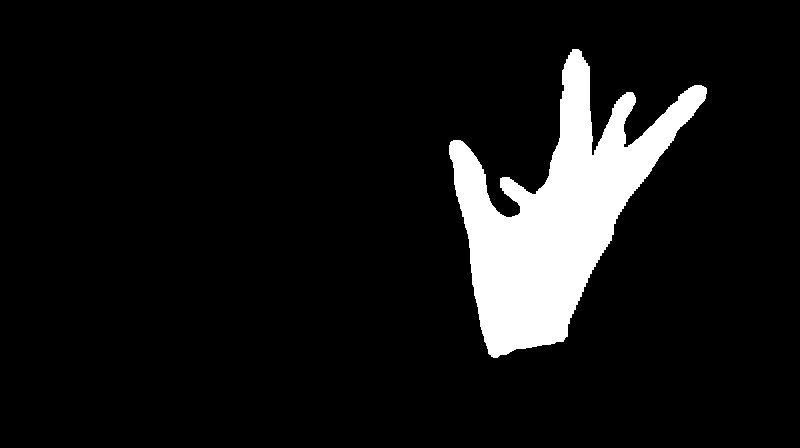}
  \end{subfigure}}
  \fbox{\begin{subfigure}[t]{0.155\linewidth}
    \includegraphics[width=\linewidth]{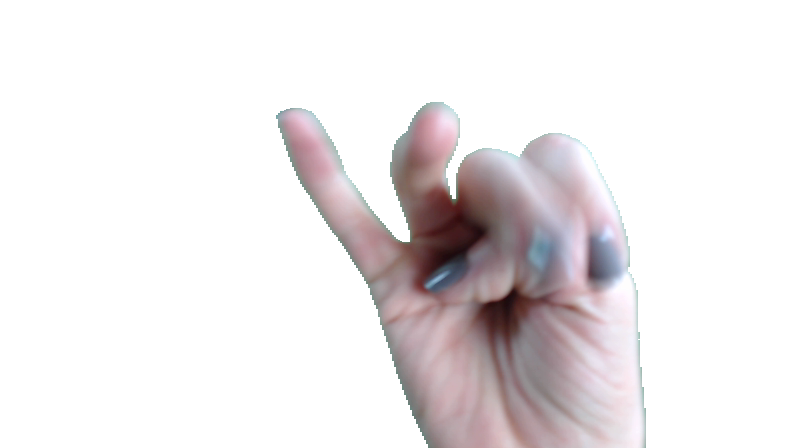}
  \end{subfigure}}
  \fbox{\begin{subfigure}[t]{0.155\linewidth}
    \includegraphics[width=\linewidth]{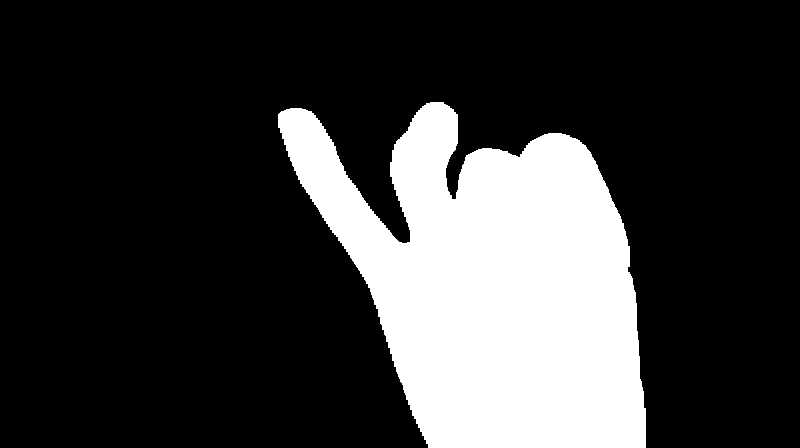}
  \end{subfigure}}
  \fbox{\begin{subfigure}[t]{0.155\linewidth}
    \includegraphics[width=\linewidth]{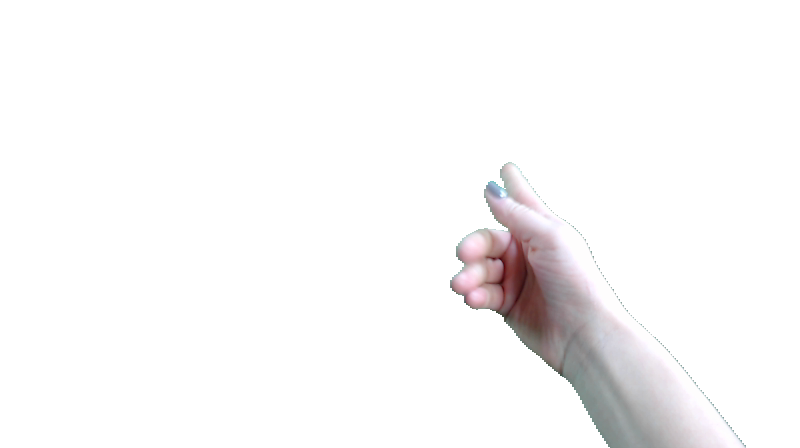}
  \end{subfigure}}
  \vspace{1mm}
  \fbox{\begin{subfigure}[t]{0.155\linewidth}
    \includegraphics[width=\linewidth]{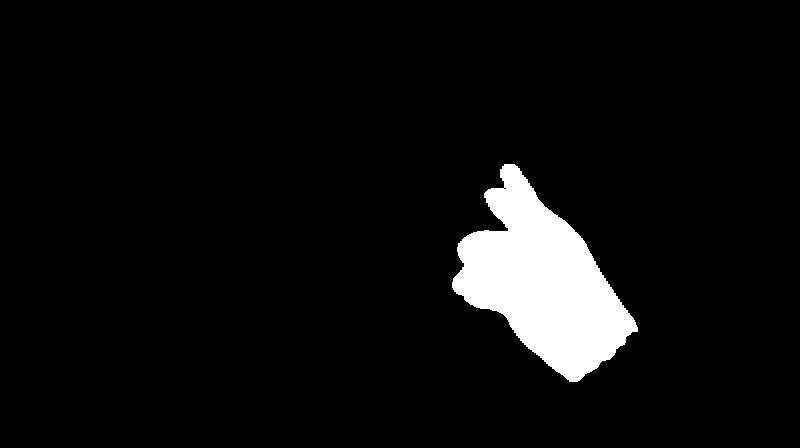}
  \end{subfigure}}
  \fbox{\begin{subfigure}[t]{0.155\linewidth}
    \includegraphics[width=\linewidth]{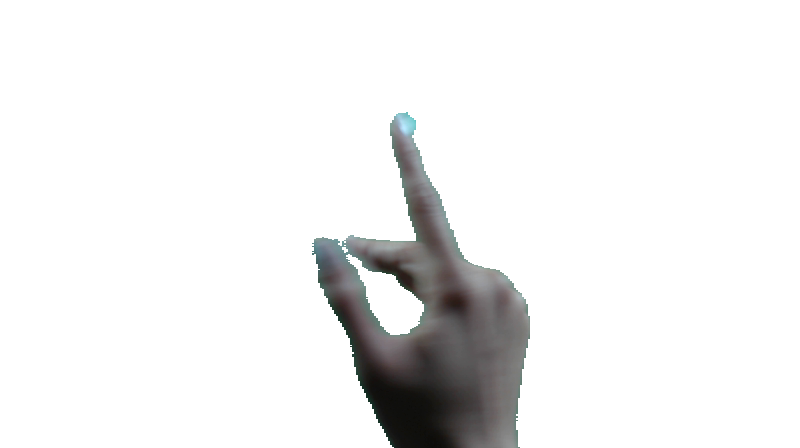}
  \end{subfigure}}
  \fbox{\begin{subfigure}[t]{0.155\linewidth}
    \includegraphics[width=\linewidth]{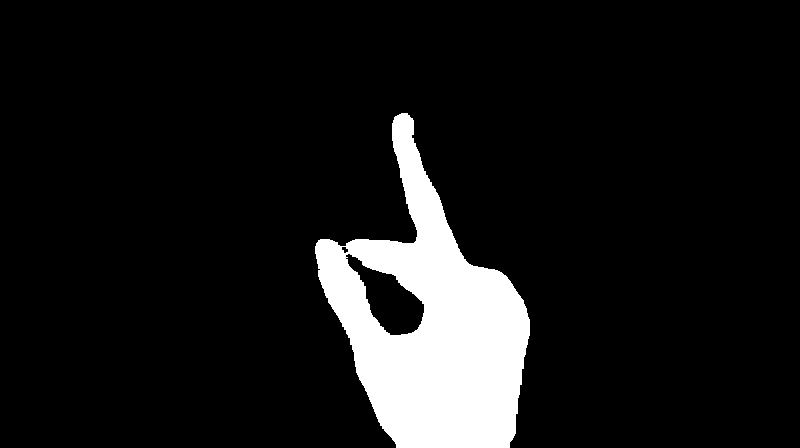}
  \end{subfigure}}
  \fbox{\begin{subfigure}[t]{0.155\linewidth}
    \includegraphics[width=\linewidth]{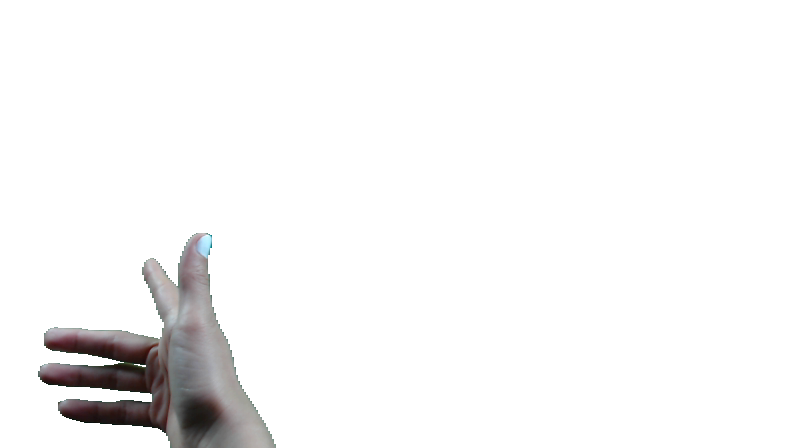}
  \end{subfigure}}
  \fbox{\begin{subfigure}[t]{0.155\linewidth}
    \includegraphics[width=\linewidth]{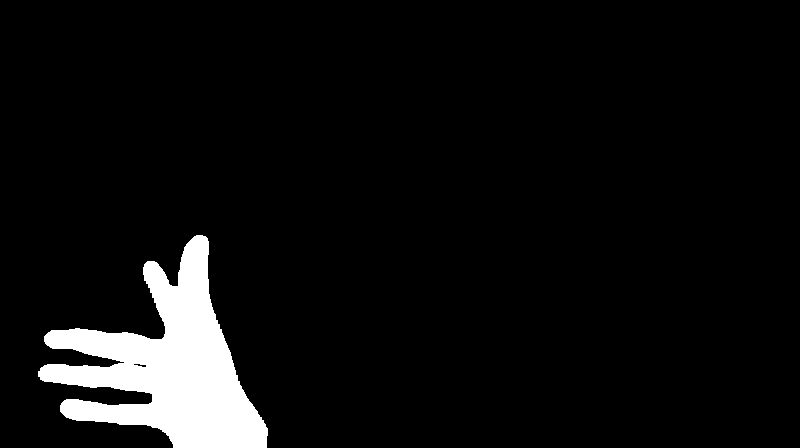}
  \end{subfigure}}
  \fbox{\begin{subfigure}[t]{0.155\linewidth}
    \includegraphics[width=\linewidth]{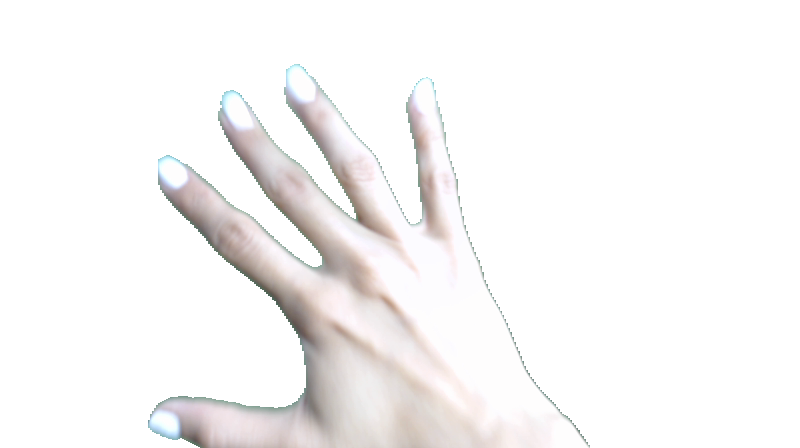}
  \end{subfigure}}
  \vspace{1mm}
  \fbox{\begin{subfigure}[t]{0.155\linewidth}
    \includegraphics[width=\linewidth]{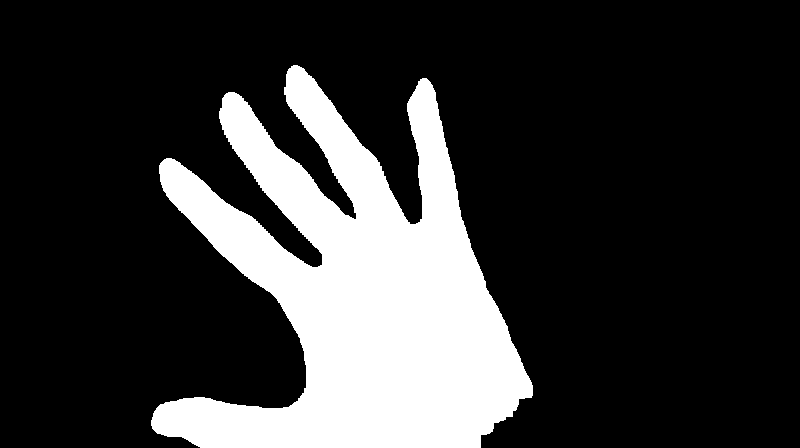}
  \end{subfigure}}
  \caption{Sample images from the training set of Ego2Hands (odd columns). Energy for hand detection is annotated for the hand region (even columns). The rows contain annotated instances from subject 0 to 8 respectively.}
  \label{fig:qualitative_results}
\end{figure*}
\subsection{Evaluation set} 
We show sample images from all 8 sequences below to demonstrate the diversity and annotation accuracy of our evaluation set. The evaluation sequences contain free two-hand motion with various skin tones and illuminations, possible heavy occlusion and motion blur. All annotations are provided with the original image resolution of $800\times448$.
\begin{figure*}[h]
  \centering
  \begin{subfigure}[t]{0.16\linewidth}
    \includegraphics[width=\linewidth]{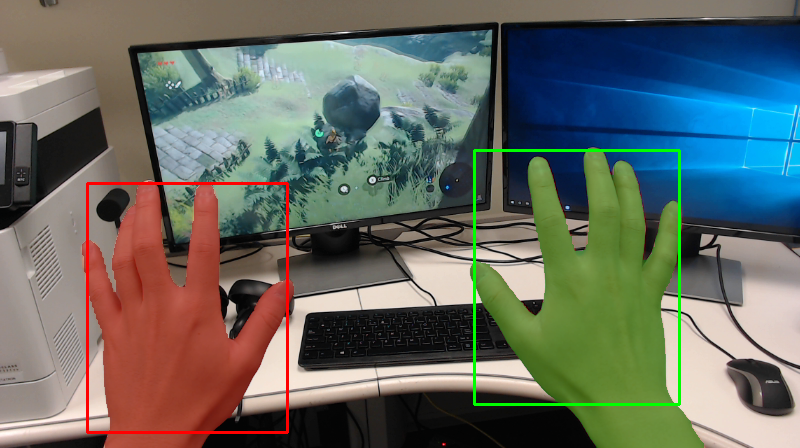}
  \end{subfigure}
  \begin{subfigure}[t]{0.16\linewidth}
    \includegraphics[width=\linewidth]{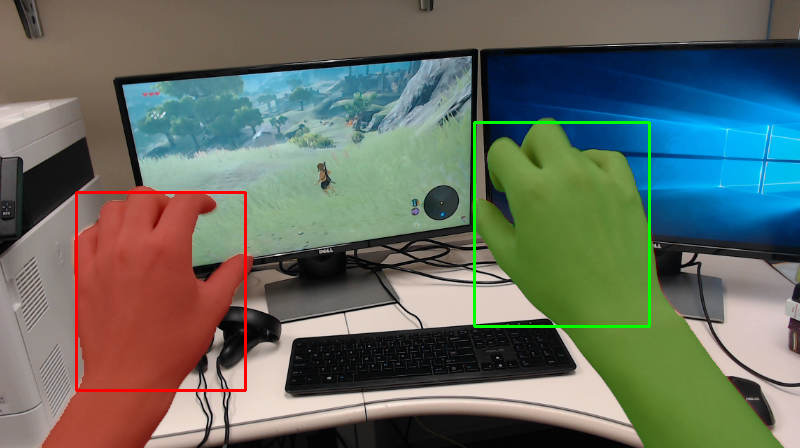}
  \end{subfigure}
  \begin{subfigure}[t]{0.16\linewidth}
    \includegraphics[width=\linewidth]{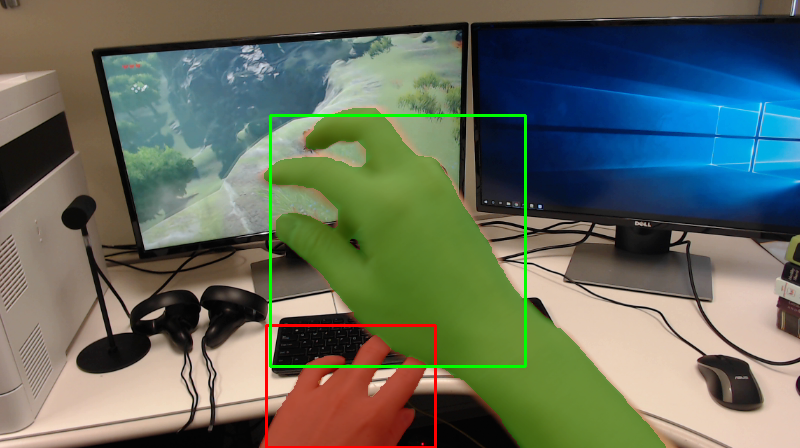}
  \end{subfigure}
  \begin{subfigure}[t]{0.16\linewidth}
    \includegraphics[width=\linewidth]{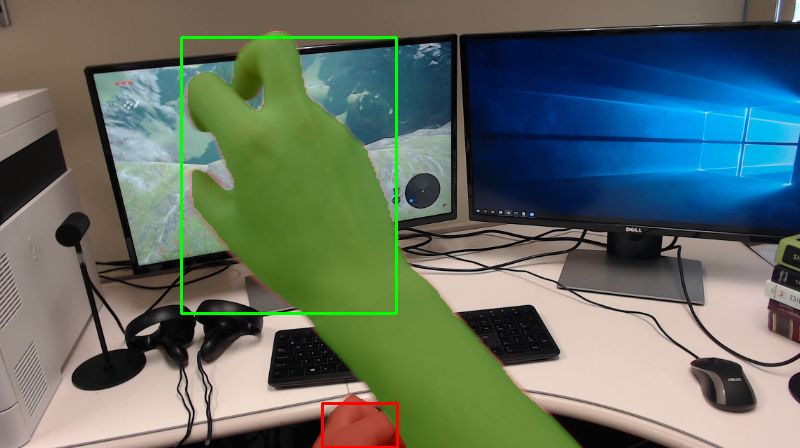}
  \end{subfigure}
  \begin{subfigure}[t]{0.16\linewidth}
    \includegraphics[width=\linewidth]{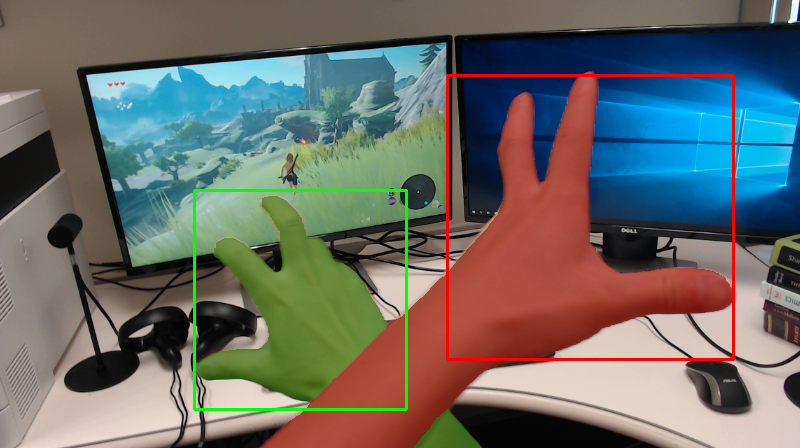}
  \end{subfigure}
  \vspace{1mm}
  \begin{subfigure}[t]{0.16\linewidth}
    \includegraphics[width=\linewidth]{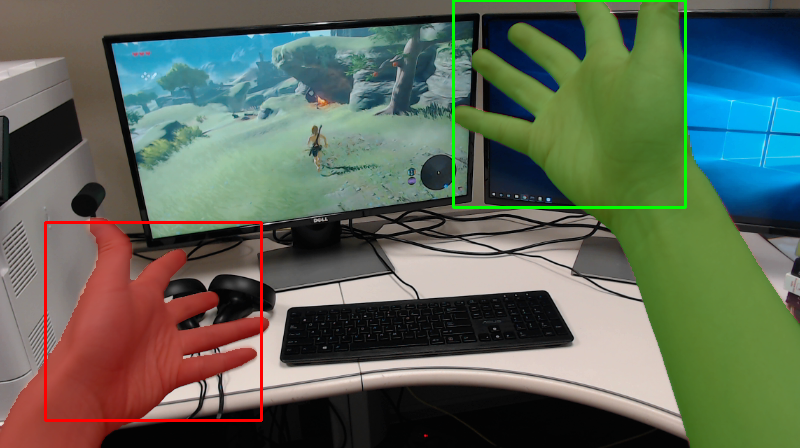}
  \end{subfigure}
   \begin{subfigure}[t]{0.16\linewidth}
    \includegraphics[width=\linewidth]{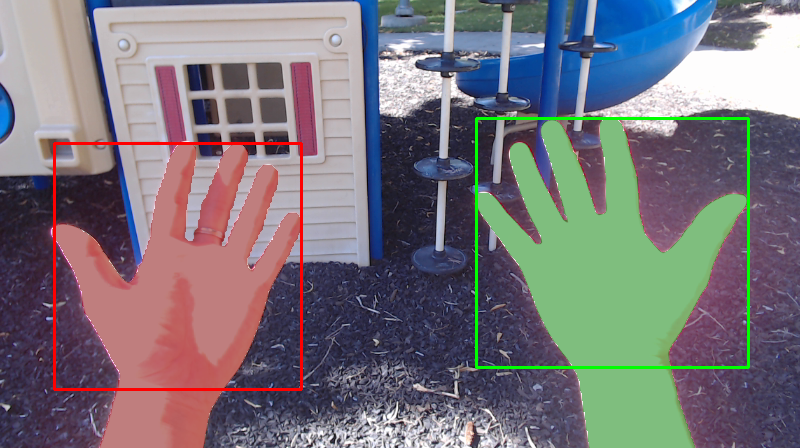}
  \end{subfigure}
  \begin{subfigure}[t]{0.16\linewidth}
    \includegraphics[width=\linewidth]{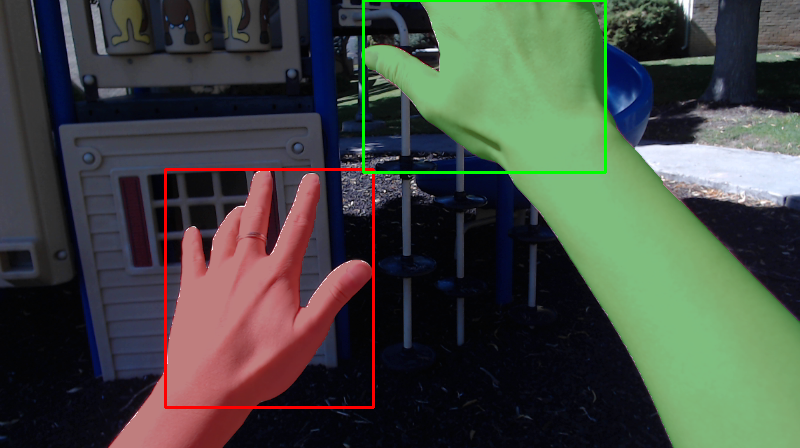}
  \end{subfigure}
  \begin{subfigure}[t]{0.16\linewidth}
    \includegraphics[width=\linewidth]{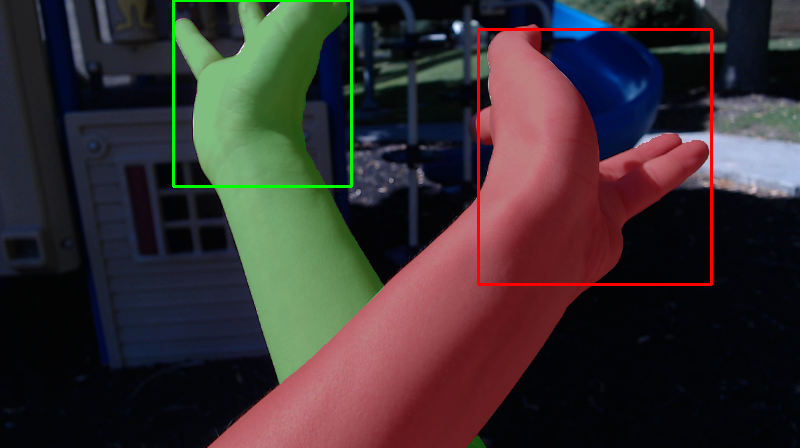}
  \end{subfigure}
  \begin{subfigure}[t]{0.16\linewidth}
    \includegraphics[width=\linewidth]{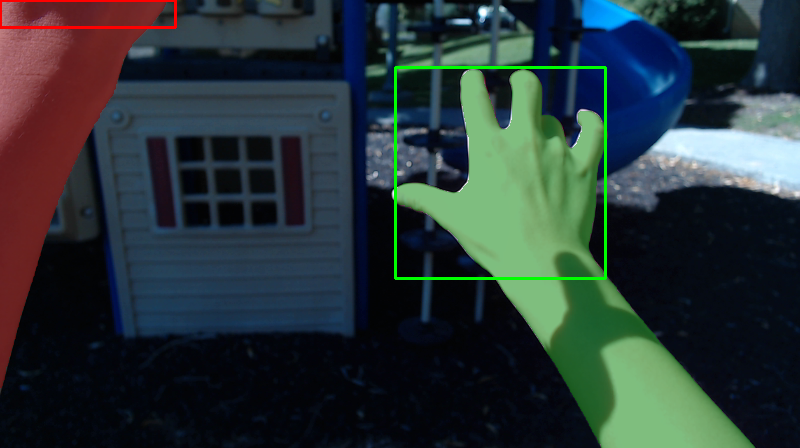}
  \end{subfigure}
  \begin{subfigure}[t]{0.16\linewidth}
    \includegraphics[width=\linewidth]{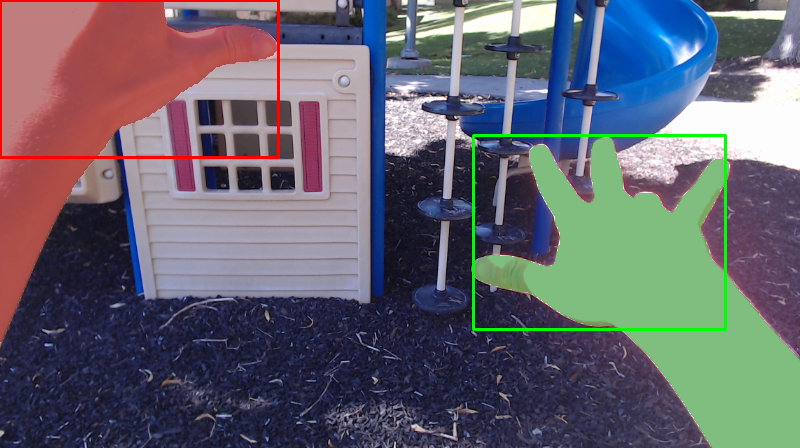}
  \end{subfigure}
  \vspace{1mm}
  \begin{subfigure}[t]{0.16\linewidth}
    \includegraphics[width=\linewidth]{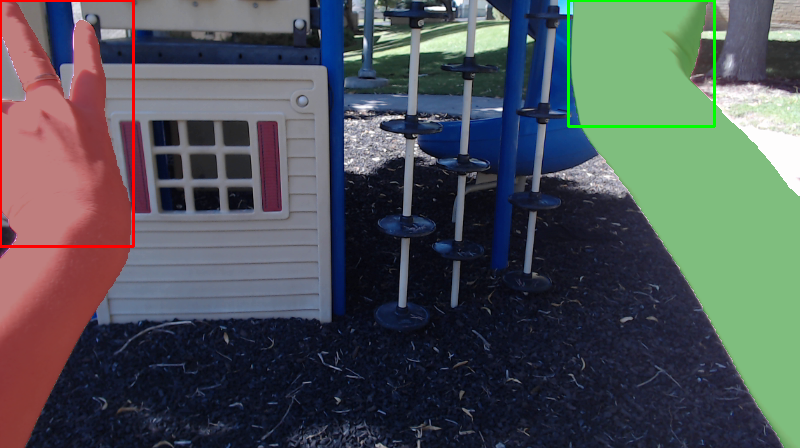}
  \end{subfigure}
   \begin{subfigure}[t]{0.16\linewidth}
    \includegraphics[width=\linewidth]{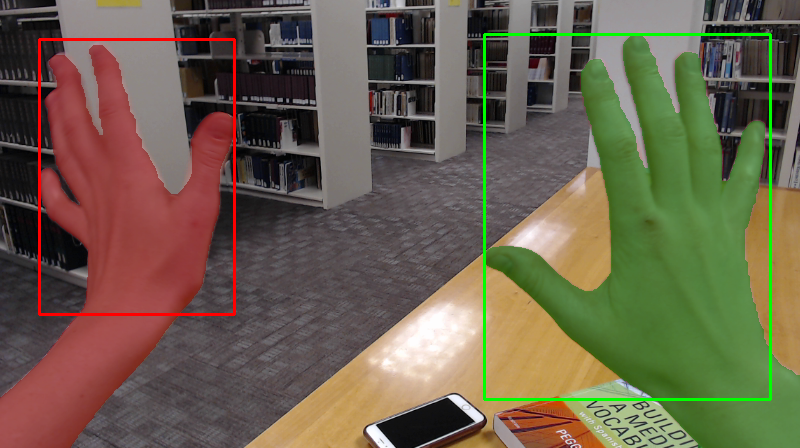}
  \end{subfigure}
  \begin{subfigure}[t]{0.16\linewidth}
    \includegraphics[width=\linewidth]{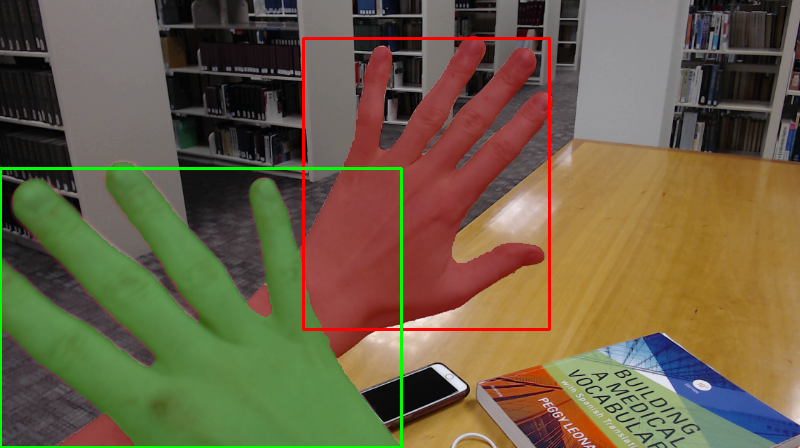}
  \end{subfigure}
  \begin{subfigure}[t]{0.16\linewidth}
    \includegraphics[width=\linewidth]{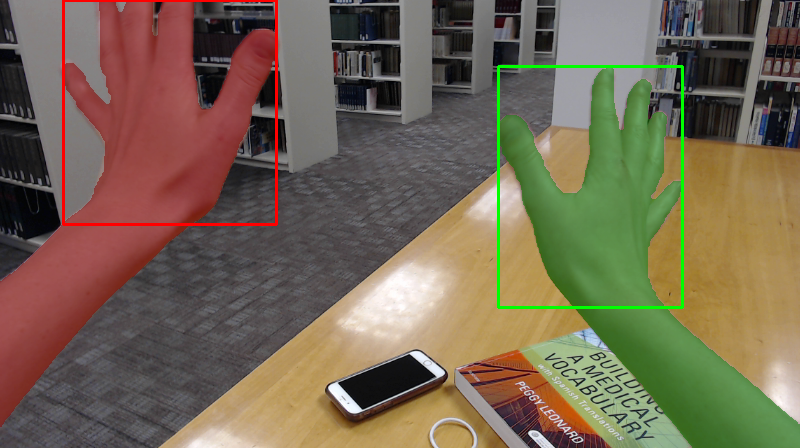}
  \end{subfigure}
  \begin{subfigure}[t]{0.16\linewidth}
    \includegraphics[width=\linewidth]{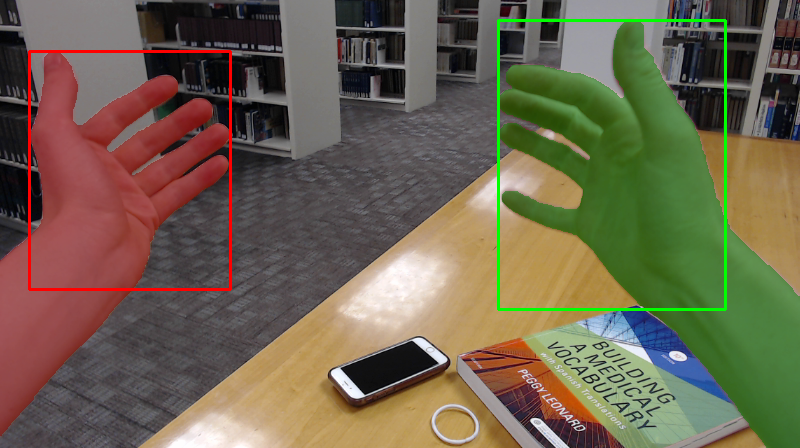}
  \end{subfigure}
  \begin{subfigure}[t]{0.16\linewidth}
    \includegraphics[width=\linewidth]{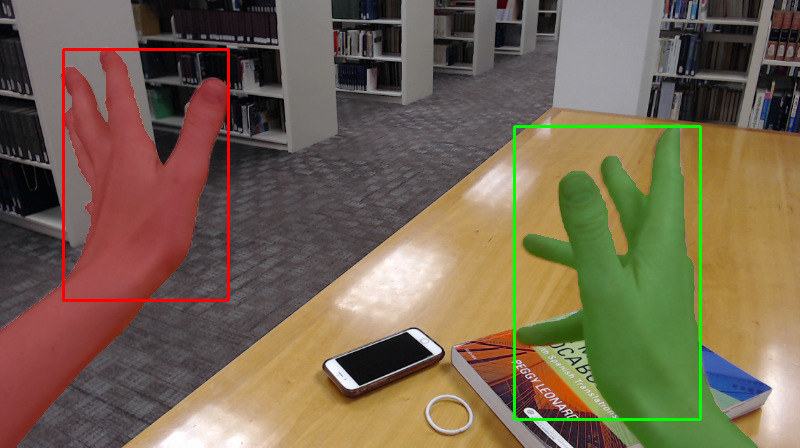}
  \end{subfigure}
  \vspace{1mm}
  \begin{subfigure}[t]{0.16\linewidth}
    \includegraphics[width=\linewidth]{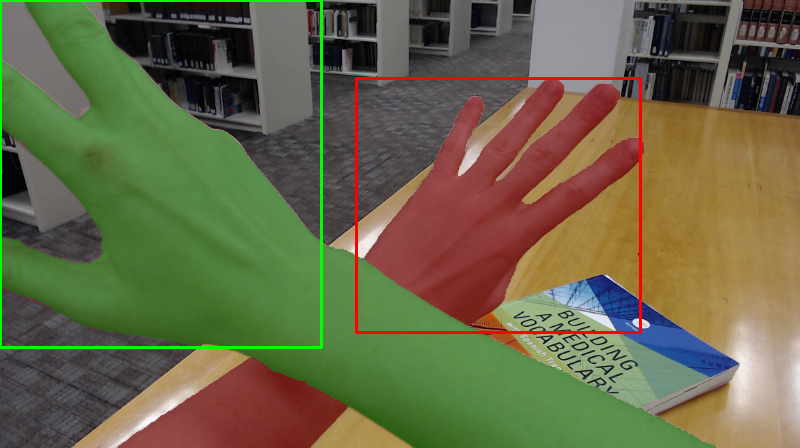}
  \end{subfigure}
   \begin{subfigure}[t]{0.16\linewidth}
    \includegraphics[width=\linewidth]{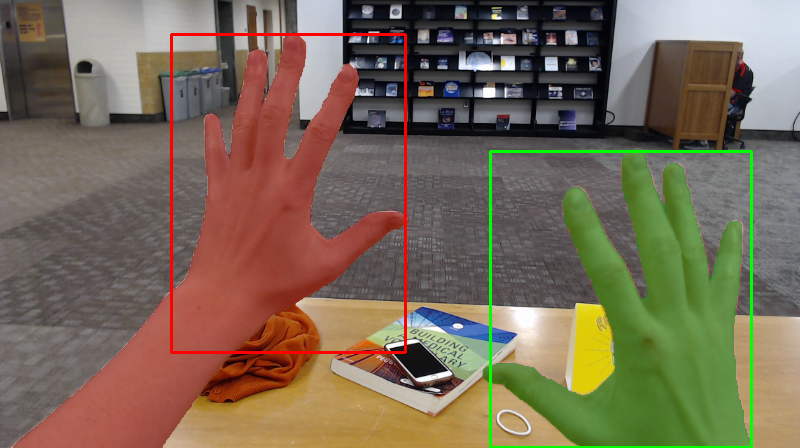}
  \end{subfigure}
  \begin{subfigure}[t]{0.16\linewidth}
    \includegraphics[width=\linewidth]{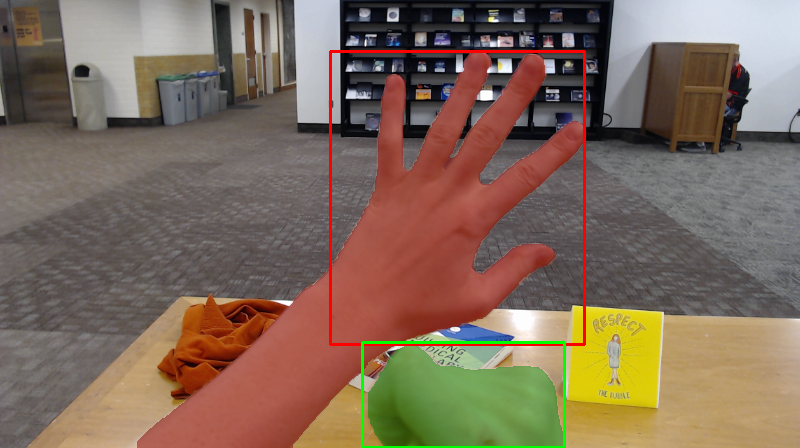}
  \end{subfigure}
  \begin{subfigure}[t]{0.16\linewidth}
    \includegraphics[width=\linewidth]{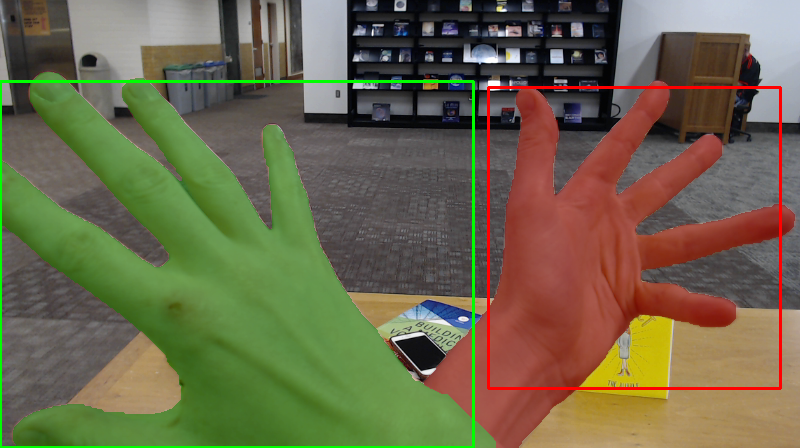}
  \end{subfigure}
  \begin{subfigure}[t]{0.16\linewidth}
    \includegraphics[width=\linewidth]{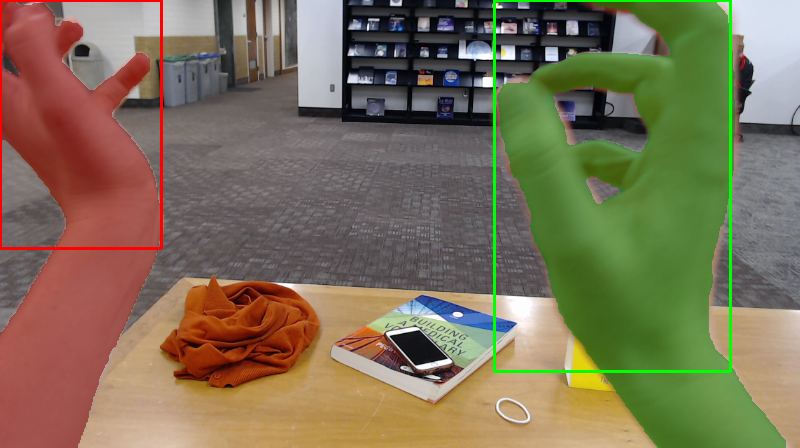}
  \end{subfigure}
  \begin{subfigure}[t]{0.16\linewidth}
    \includegraphics[width=\linewidth]{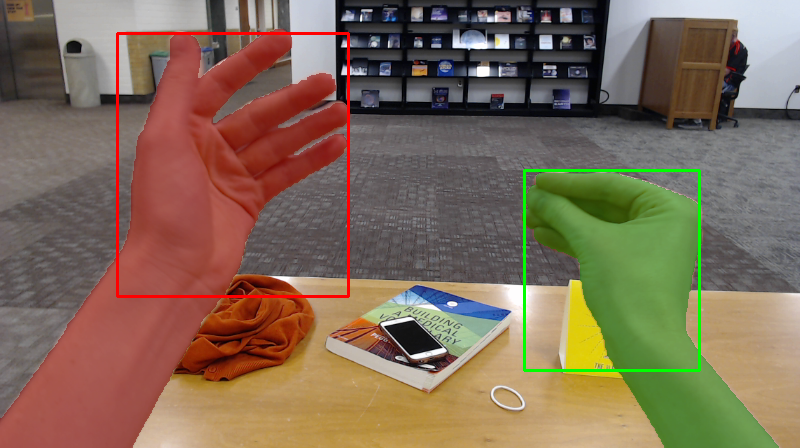}
  \end{subfigure}
  \vspace{1mm}
  \begin{subfigure}[t]{0.16\linewidth}
    \includegraphics[width=\linewidth]{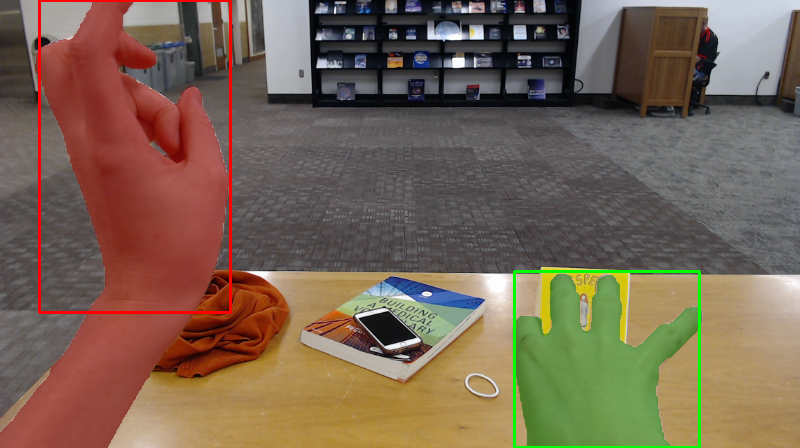}
  \end{subfigure}
   \begin{subfigure}[t]{0.16\linewidth}
    \includegraphics[width=\linewidth]{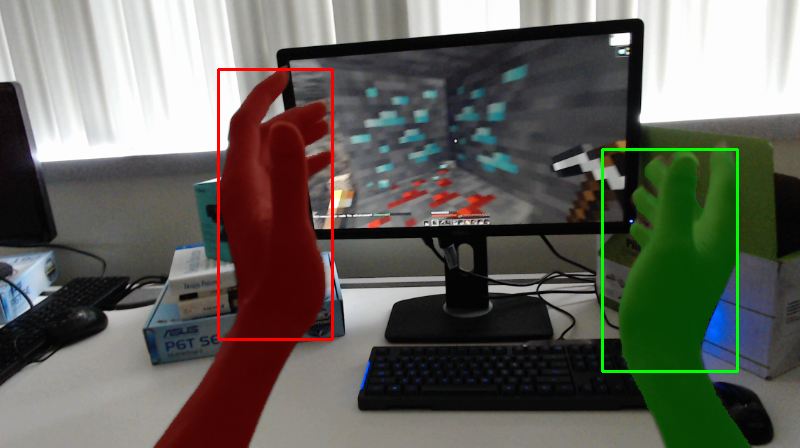}
  \end{subfigure}
  \begin{subfigure}[t]{0.16\linewidth}
    \includegraphics[width=\linewidth]{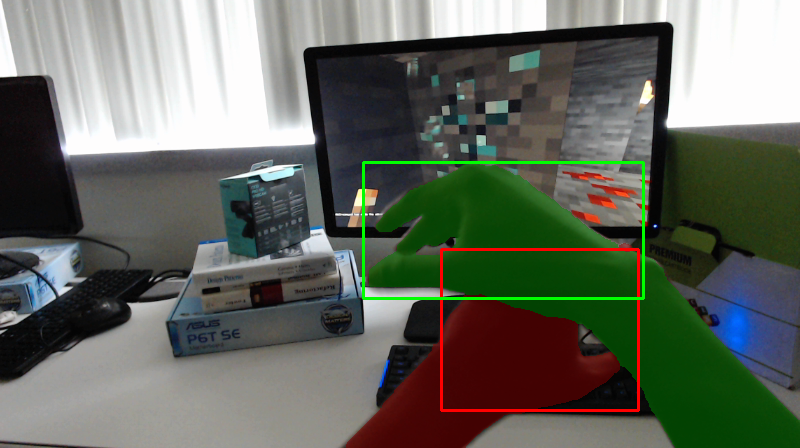}
  \end{subfigure}
  \begin{subfigure}[t]{0.16\linewidth}
    \includegraphics[width=\linewidth]{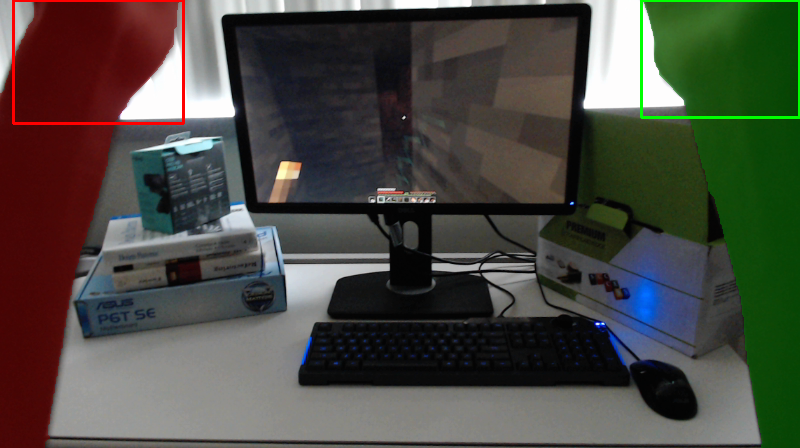}
  \end{subfigure}
  \begin{subfigure}[t]{0.16\linewidth}
    \includegraphics[width=\linewidth]{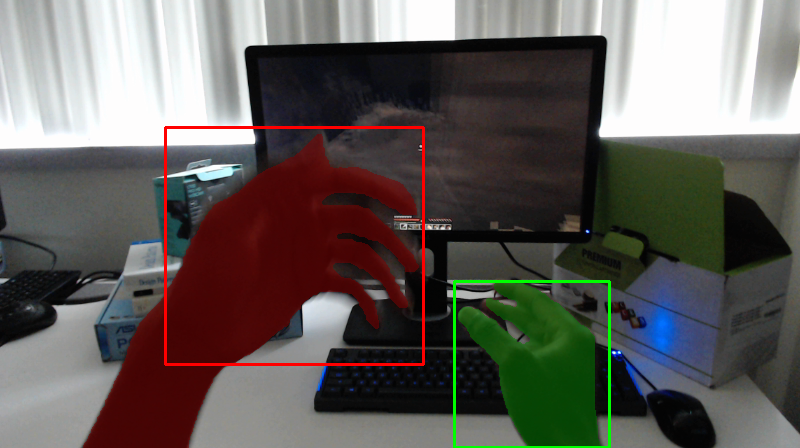}
  \end{subfigure}
  \begin{subfigure}[t]{0.16\linewidth}
    \includegraphics[width=\linewidth]{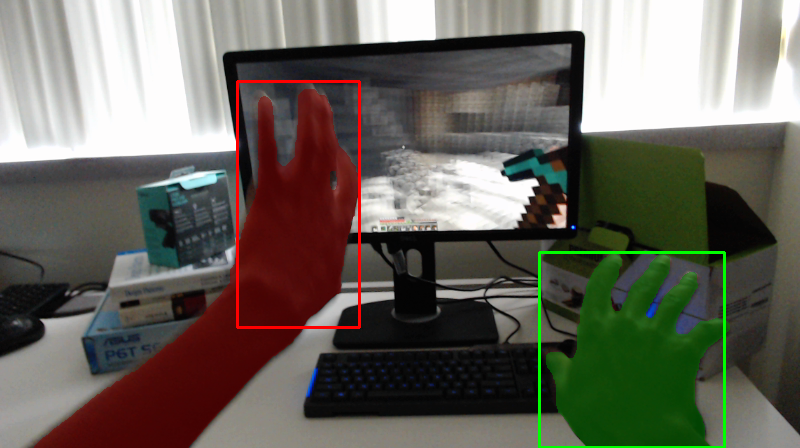}
  \end{subfigure}
  \vspace{1mm}
  \begin{subfigure}[t]{0.16\linewidth}
    \includegraphics[width=\linewidth]{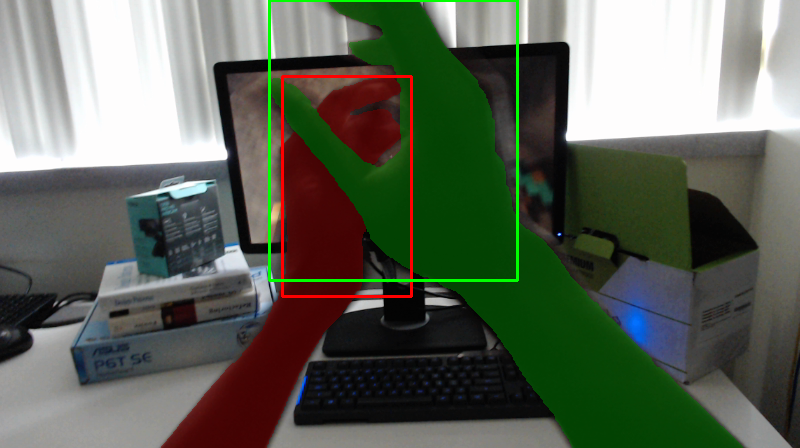}
  \end{subfigure}
   \begin{subfigure}[t]{0.16\linewidth}
    \includegraphics[width=\linewidth]{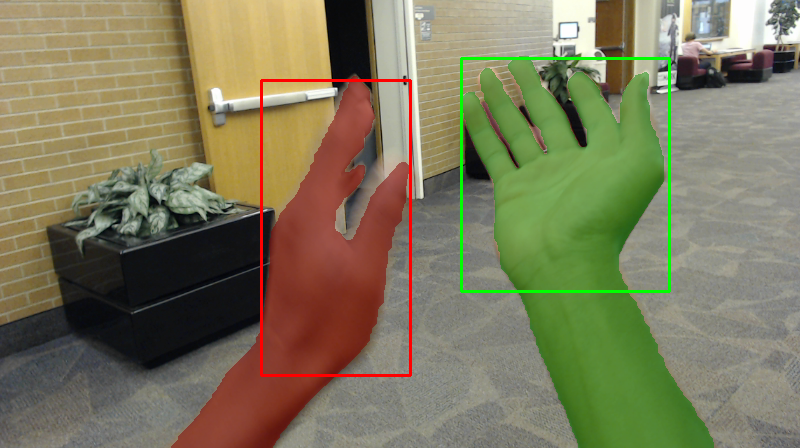}
  \end{subfigure}
  \begin{subfigure}[t]{0.16\linewidth}
    \includegraphics[width=\linewidth]{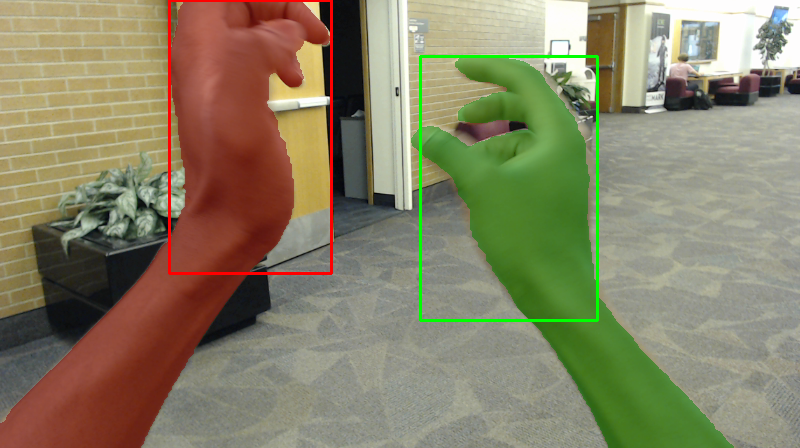}
  \end{subfigure}
  \begin{subfigure}[t]{0.16\linewidth}
    \includegraphics[width=\linewidth]{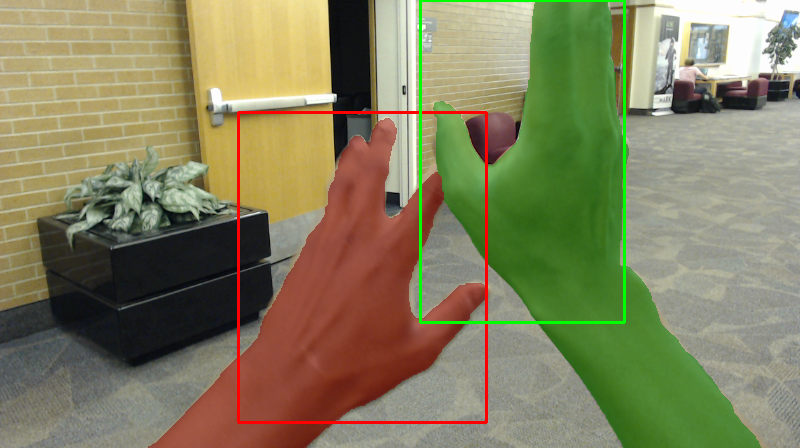}
  \end{subfigure}
  \begin{subfigure}[t]{0.16\linewidth}
    \includegraphics[width=\linewidth]{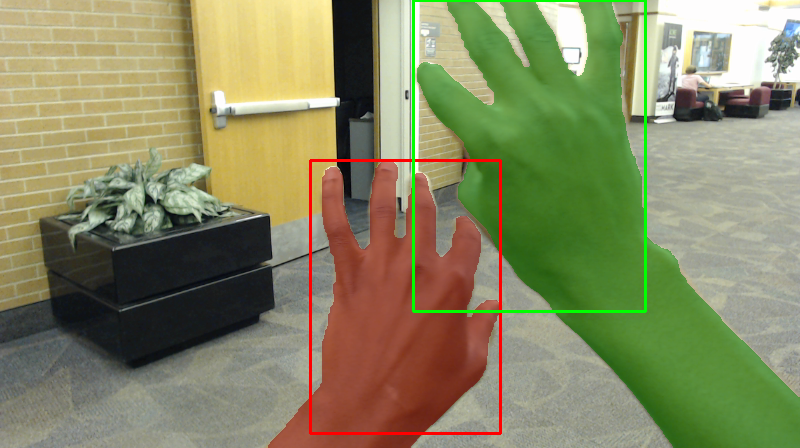}
  \end{subfigure}
  \begin{subfigure}[t]{0.16\linewidth}
    \includegraphics[width=\linewidth]{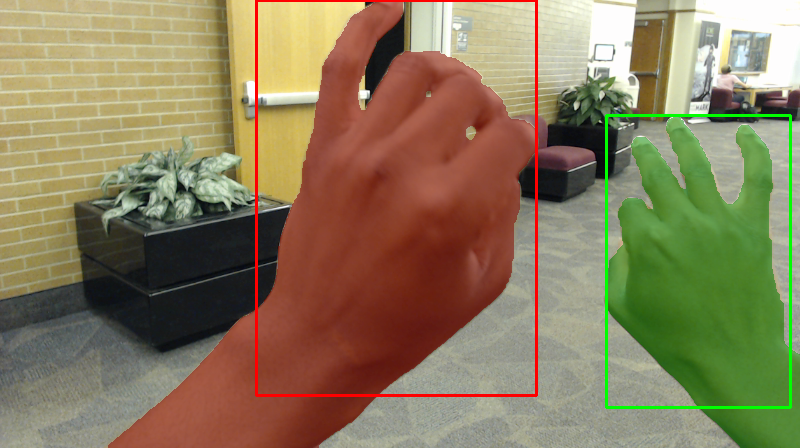}
  \end{subfigure}
  \vspace{1mm}
  \begin{subfigure}[t]{0.16\linewidth}
    \includegraphics[width=\linewidth]{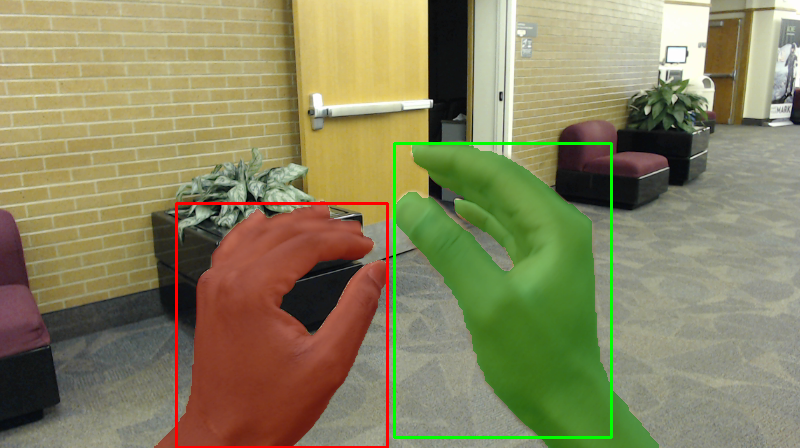}
  \end{subfigure}
   \begin{subfigure}[t]{0.16\linewidth}
    \includegraphics[width=\linewidth]{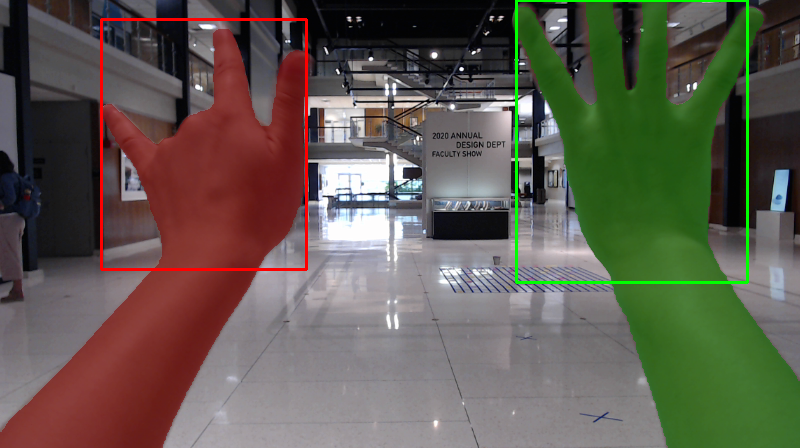}
  \end{subfigure}
  \begin{subfigure}[t]{0.16\linewidth}
    \includegraphics[width=\linewidth]{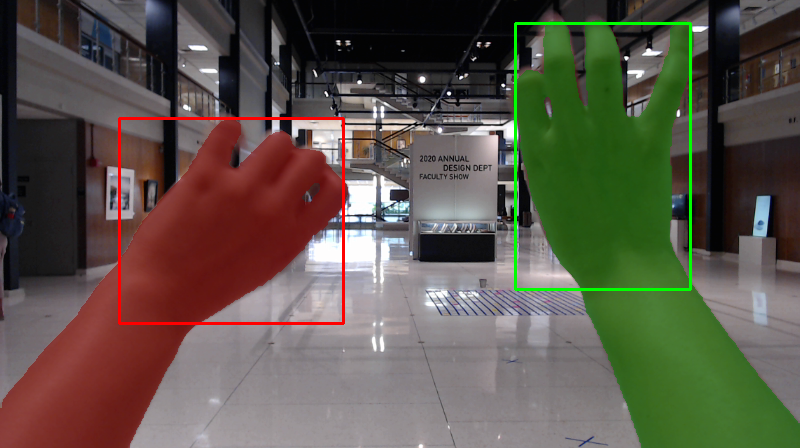}
  \end{subfigure}
  \begin{subfigure}[t]{0.16\linewidth}
    \includegraphics[width=\linewidth]{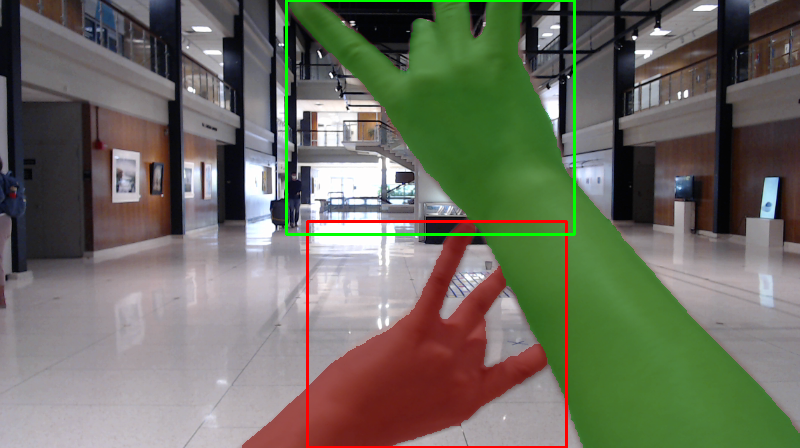}
  \end{subfigure}
  \begin{subfigure}[t]{0.16\linewidth}
    \includegraphics[width=\linewidth]{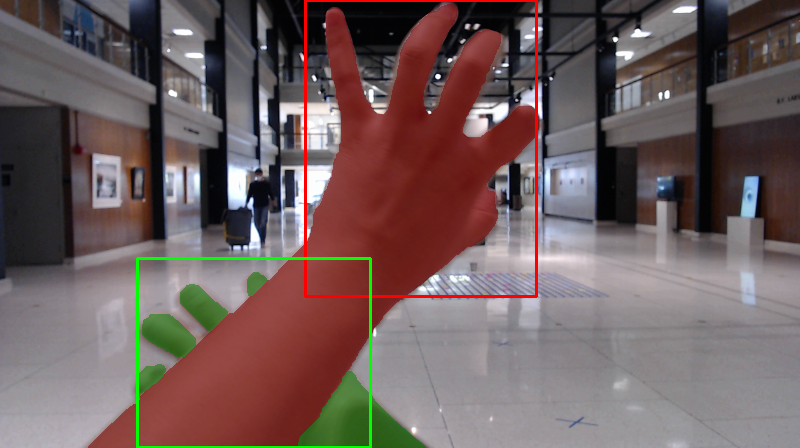}
  \end{subfigure}
  \begin{subfigure}[t]{0.16\linewidth}
    \includegraphics[width=\linewidth]{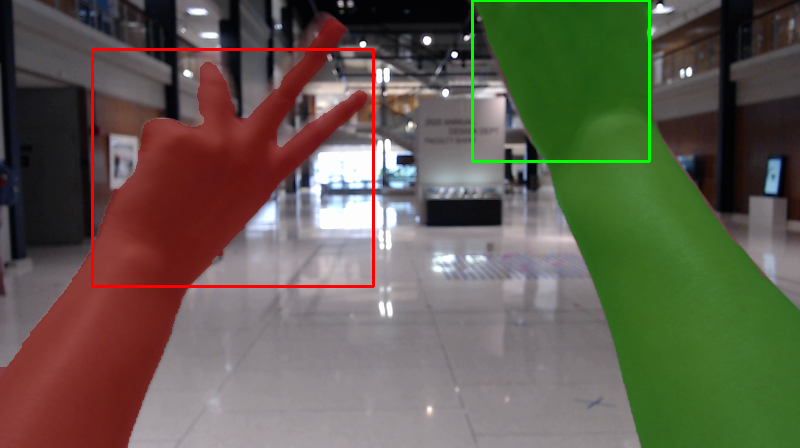}
  \end{subfigure}
  \vspace{1mm}
  \begin{subfigure}[t]{0.16\linewidth}
    \includegraphics[width=\linewidth]{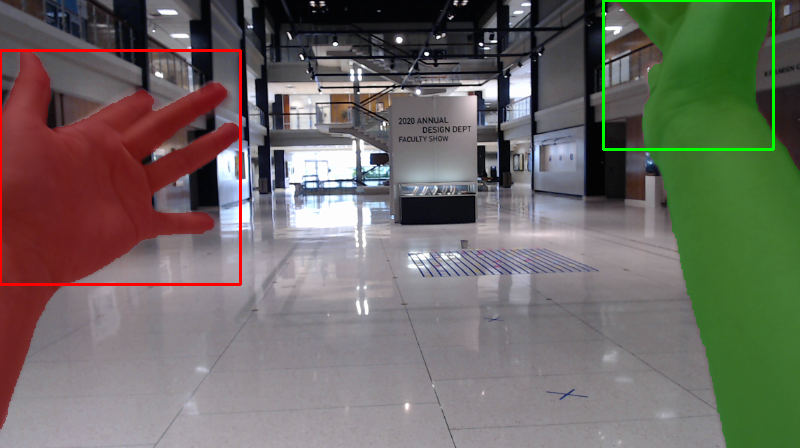}
  \end{subfigure}
   \begin{subfigure}[t]{0.16\linewidth}
    \includegraphics[width=\linewidth]{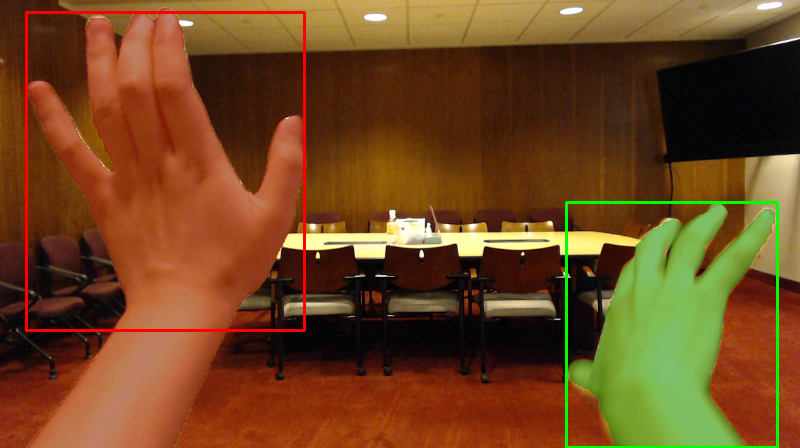}
    \caption*{t = 0}
  \end{subfigure}
  \begin{subfigure}[t]{0.16\linewidth}
    \includegraphics[width=\linewidth]{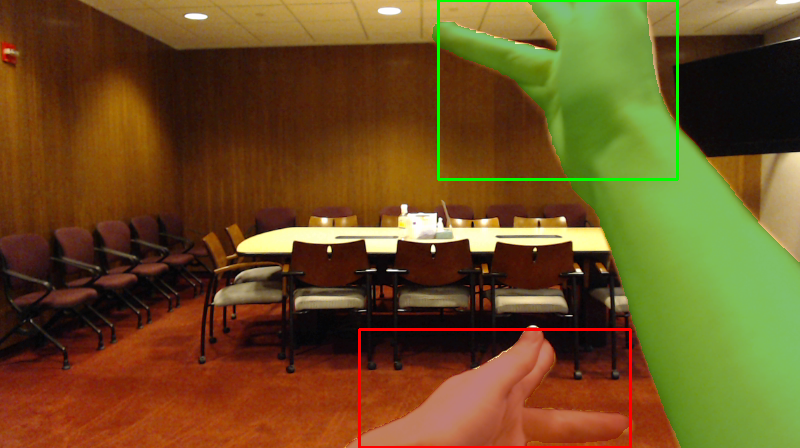}
    \caption*{t = 49}
  \end{subfigure}
  \begin{subfigure}[t]{0.16\linewidth}
    \includegraphics[width=\linewidth]{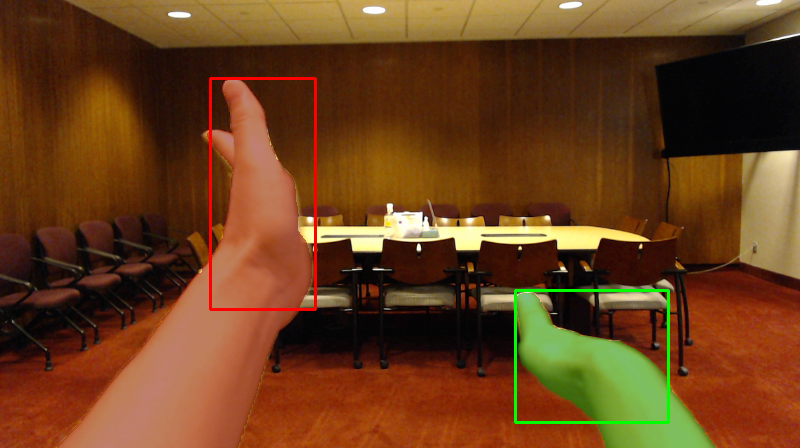}
    \caption*{t = 99}
  \end{subfigure}
  \begin{subfigure}[t]{0.16\linewidth}
    \includegraphics[width=\linewidth]{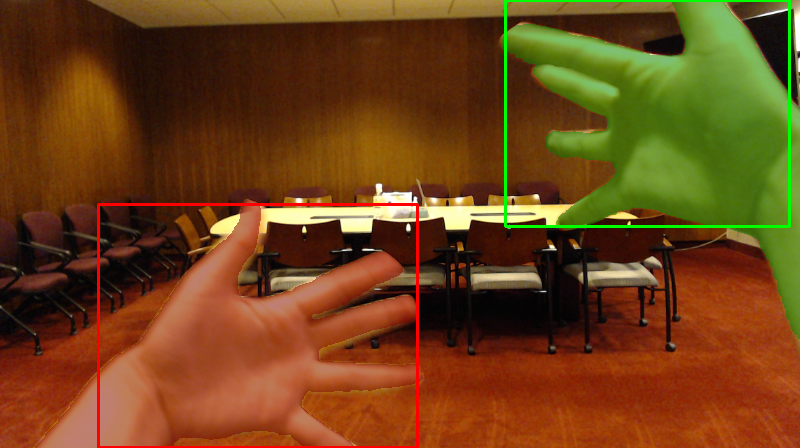}
    \caption*{t = 149}
  \end{subfigure}
  \begin{subfigure}[t]{0.16\linewidth}
    \includegraphics[width=\linewidth]{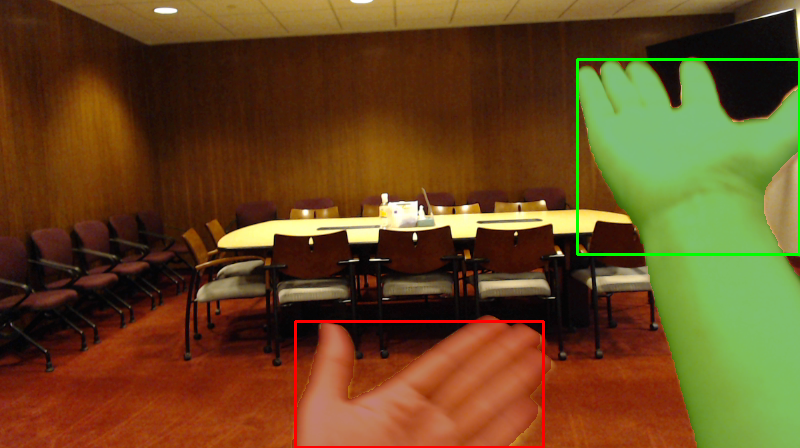}
    \caption*{t = 199}
  \end{subfigure}
  \vspace{4mm}
  \begin{subfigure}[t]{0.16\linewidth}
    \includegraphics[width=\linewidth]{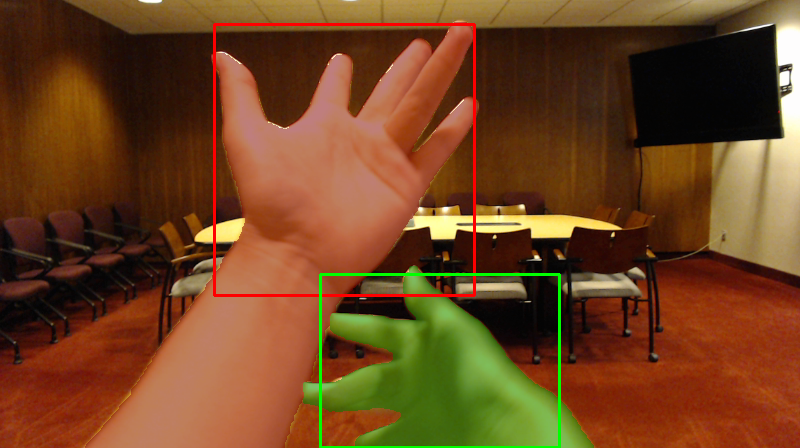}
    \caption*{t = 249}
  \end{subfigure}
  \caption{Sample annotated images from the 8 sequences of the evaluation set of Ego2Hands. Segmentation annotation is overlaid in colors. Detection is annotated as energy and visualized as bounding boxes. Best viewed in magnification.}
  \label{fig:qualitative_results}
\end{figure*}
\section{Cross-dataset Evaluation Comparison}
\begin{figure*}[t]
  \centering
  \begin{subfigure}[t]{0.16\linewidth}
    \includegraphics[width=\linewidth]{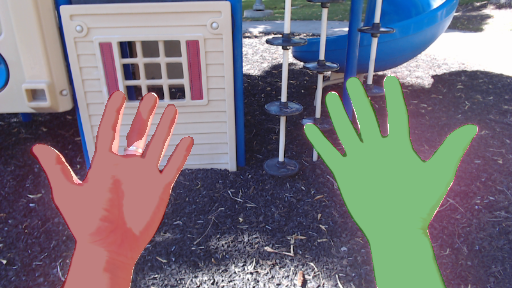}
  \end{subfigure}
  \begin{subfigure}[t]{0.16\linewidth}
    \includegraphics[width=\linewidth]{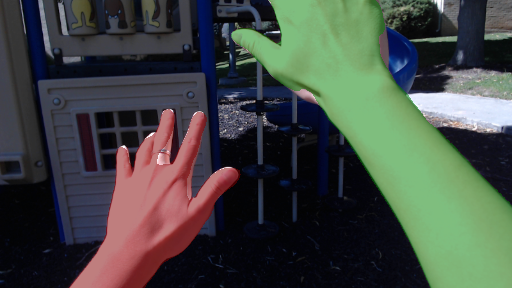}
  \end{subfigure}
  \begin{subfigure}[t]{0.16\linewidth}
    \includegraphics[width=\linewidth]{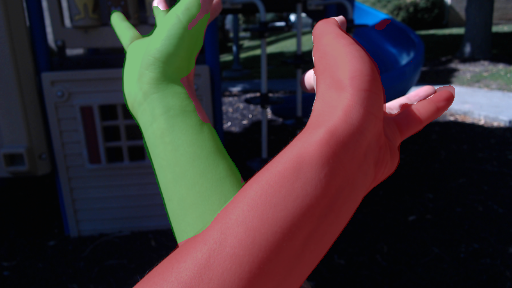}
  \end{subfigure}
  \begin{subfigure}[t]{0.16\linewidth}
    \includegraphics[width=\linewidth]{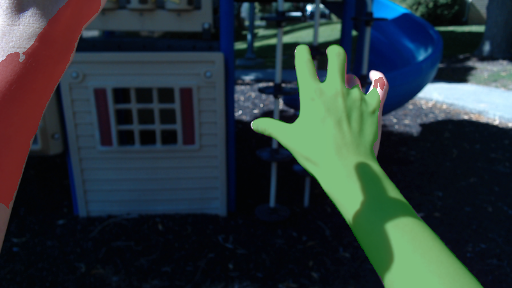}
  \end{subfigure}
  \begin{subfigure}[t]{0.16\linewidth}
    \includegraphics[width=\linewidth]{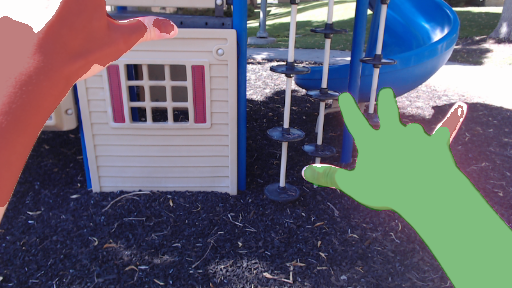}
  \end{subfigure}
  \vspace{1mm}
  \begin{subfigure}[t]{0.16\linewidth}
    \includegraphics[width=\linewidth]{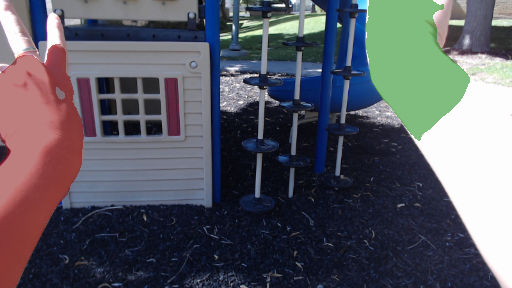}
  \end{subfigure}
   \begin{subfigure}[t]{0.16\linewidth}
    \includegraphics[width=\linewidth]{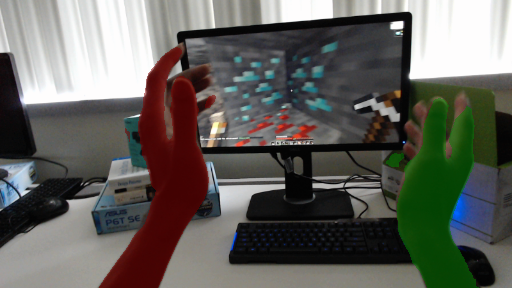}
  \end{subfigure}
  \begin{subfigure}[t]{0.16\linewidth}
    \includegraphics[width=\linewidth]{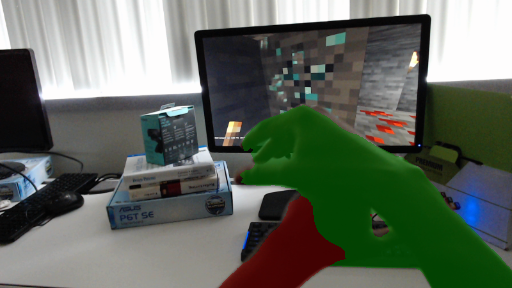}
  \end{subfigure}
  \begin{subfigure}[t]{0.16\linewidth}
    \includegraphics[width=\linewidth]{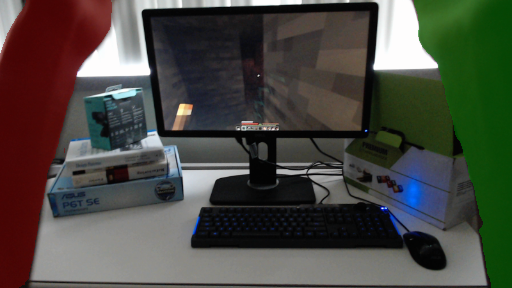}
  \end{subfigure}
  \begin{subfigure}[t]{0.16\linewidth}
    \includegraphics[width=\linewidth]{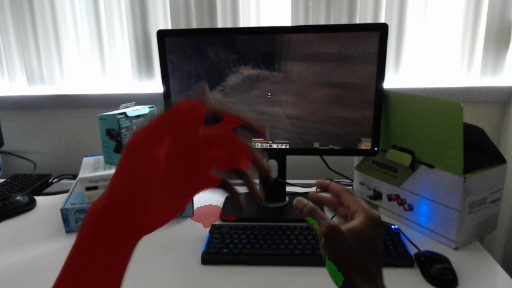}
  \end{subfigure}
  \begin{subfigure}[t]{0.16\linewidth}
    \includegraphics[width=\linewidth]{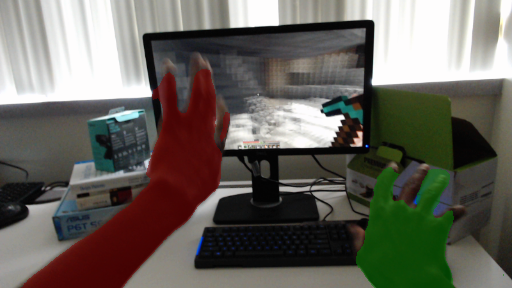}
  \end{subfigure}
  \vspace{1mm}
  \begin{subfigure}[t]{0.16\linewidth}
    \includegraphics[width=\linewidth]{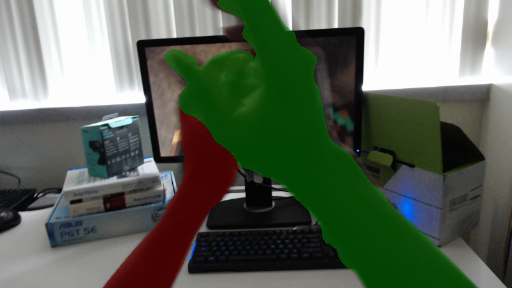}
  \end{subfigure}
   \begin{subfigure}[t]{0.16\linewidth}
    \includegraphics[width=\linewidth]{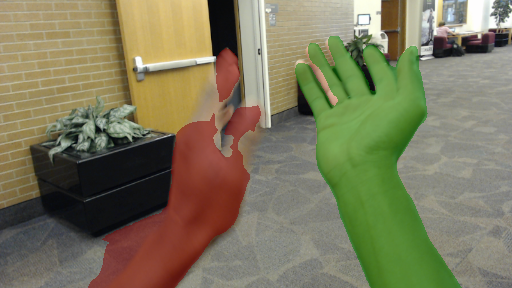}
  \end{subfigure}
  \begin{subfigure}[t]{0.16\linewidth}
    \includegraphics[width=\linewidth]{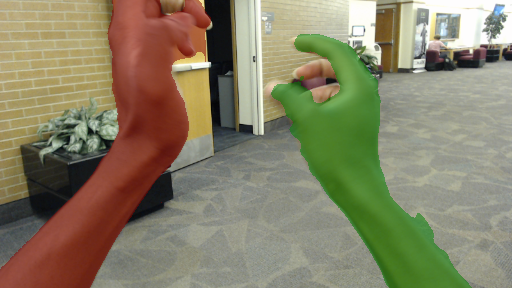}
  \end{subfigure}
  \begin{subfigure}[t]{0.16\linewidth}
    \includegraphics[width=\linewidth]{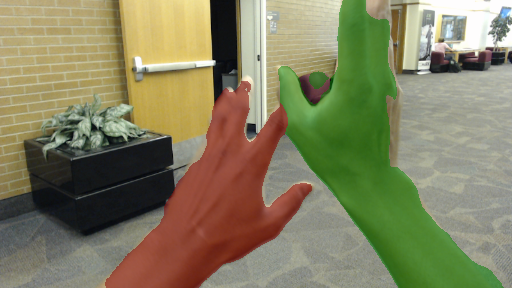}
  \end{subfigure}
  \begin{subfigure}[t]{0.16\linewidth}
    \includegraphics[width=\linewidth]{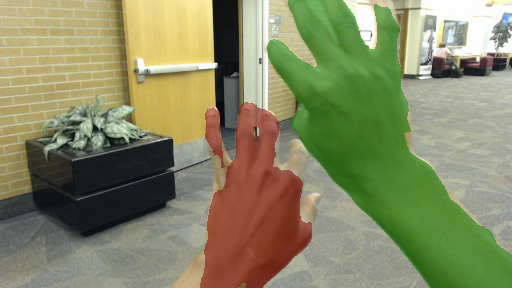}
  \end{subfigure}
  \begin{subfigure}[t]{0.16\linewidth}
    \includegraphics[width=\linewidth]{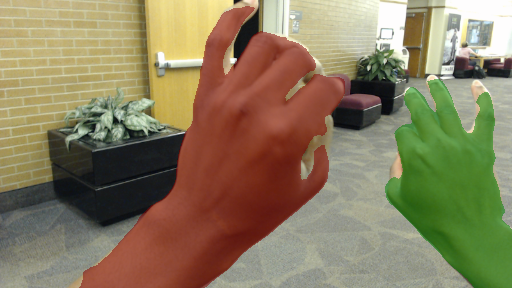}
  \end{subfigure}
  \vspace{1mm}
  \begin{subfigure}[t]{0.16\linewidth}
    \includegraphics[width=\linewidth]{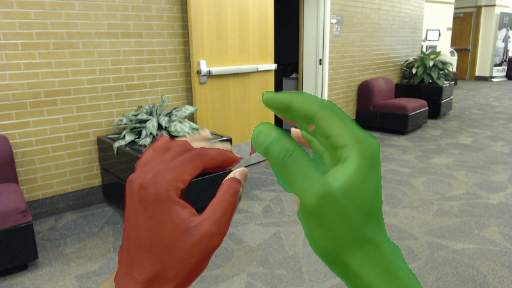}
  \end{subfigure}
   \begin{subfigure}[t]{0.16\linewidth}
    \includegraphics[width=\linewidth]{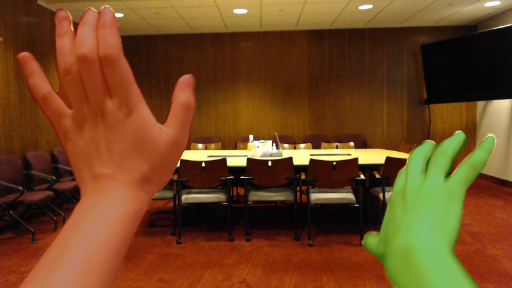}
    \caption*{t = 0}
  \end{subfigure}
  \begin{subfigure}[t]{0.16\linewidth}
    \includegraphics[width=\linewidth]{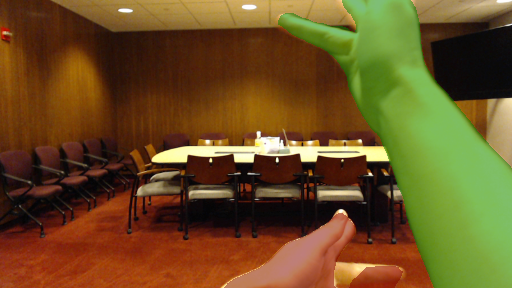}
    \caption*{t = 49}
  \end{subfigure}
  \begin{subfigure}[t]{0.16\linewidth}
    \includegraphics[width=\linewidth]{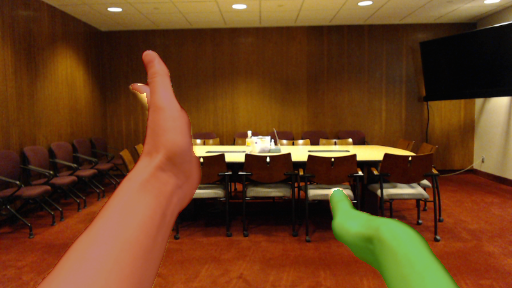}
    \caption*{t = 99}
  \end{subfigure}
  \begin{subfigure}[t]{0.16\linewidth}
    \includegraphics[width=\linewidth]{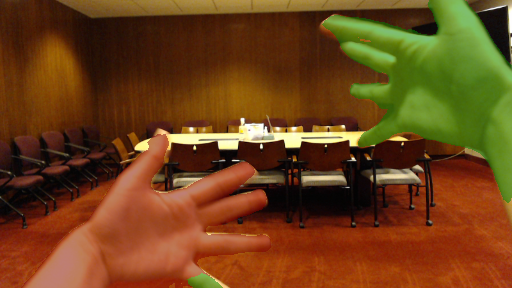}
    \caption*{t = 149}
  \end{subfigure}
  \begin{subfigure}[t]{0.16\linewidth}
    \includegraphics[width=\linewidth]{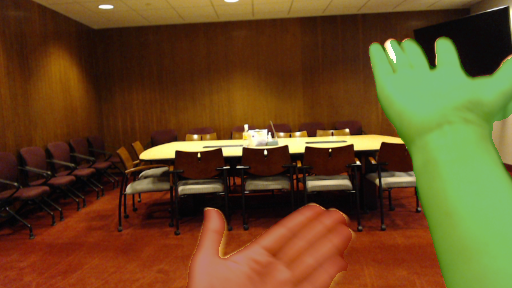}
    \caption*{t = 199}
  \end{subfigure}
  \vspace{4mm}
  \begin{subfigure}[t]{0.16\linewidth}
    \includegraphics[width=\linewidth]{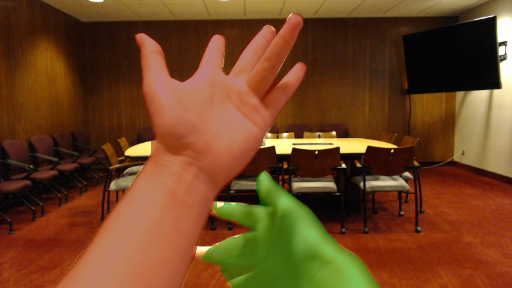}
    \caption*{t = 249}
  \end{subfigure}
  \caption{Qualitative samples of model trained on $\textbf{Ego2Hands}_\textbf{\textit{train}}$ on sequence 2, 5, 6, 8 of $\text{Ego2Hands}_{test}$.}
  \label{fig:cross_eval_ego2hands}
\end{figure*}

\begin{figure*}[t]
  \centering
  \begin{subfigure}[t]{0.16\linewidth}
    \includegraphics[width=\linewidth]{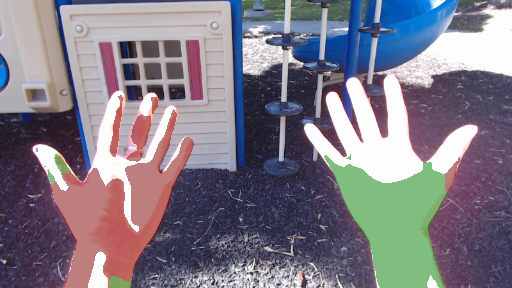}
  \end{subfigure}
  \begin{subfigure}[t]{0.16\linewidth}
    \includegraphics[width=\linewidth]{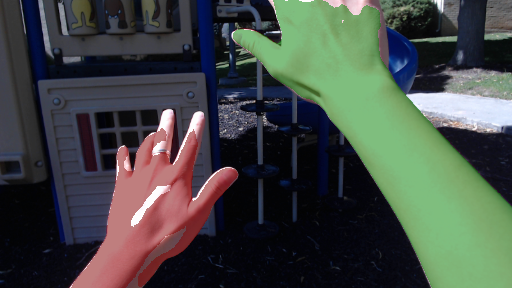}
  \end{subfigure}
  \begin{subfigure}[t]{0.16\linewidth}
    \includegraphics[width=\linewidth]{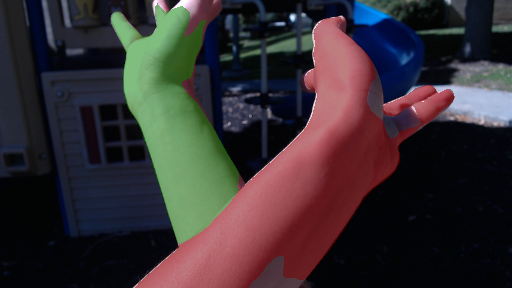}
  \end{subfigure}
  \begin{subfigure}[t]{0.16\linewidth}
    \includegraphics[width=\linewidth]{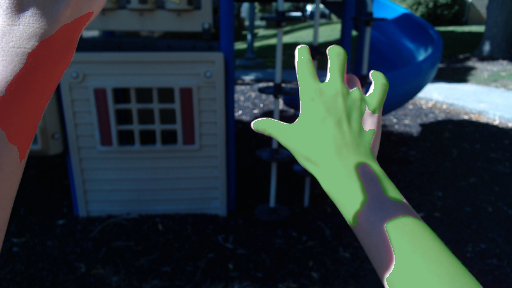}
  \end{subfigure}
  \begin{subfigure}[t]{0.16\linewidth}
    \includegraphics[width=\linewidth]{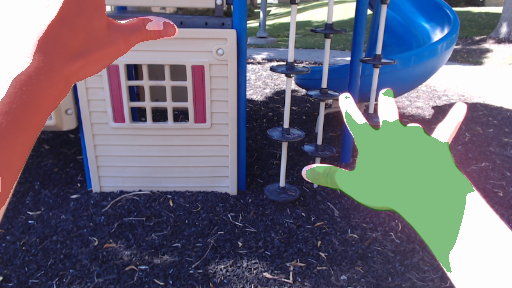}
  \end{subfigure}
  \vspace{1mm}
  \begin{subfigure}[t]{0.16\linewidth}
    \includegraphics[width=\linewidth]{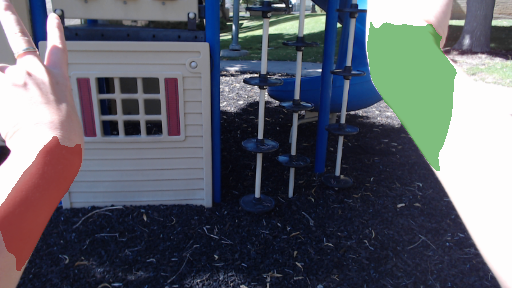}
  \end{subfigure}
   \begin{subfigure}[t]{0.16\linewidth}
    \includegraphics[width=\linewidth]{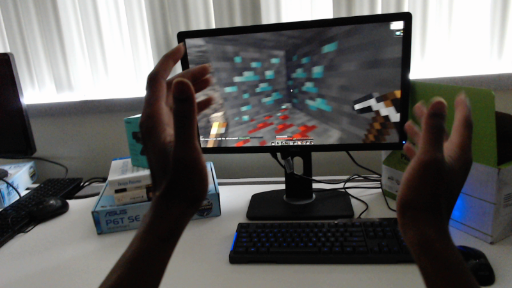}
  \end{subfigure}
  \begin{subfigure}[t]{0.16\linewidth}
    \includegraphics[width=\linewidth]{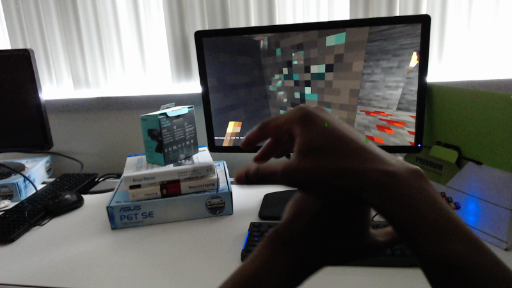}
  \end{subfigure}
  \begin{subfigure}[t]{0.16\linewidth}
    \includegraphics[width=\linewidth]{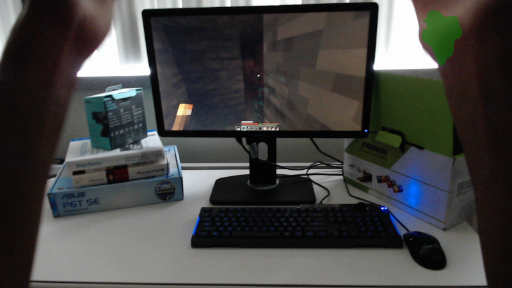}
  \end{subfigure}
  \begin{subfigure}[t]{0.16\linewidth}
    \includegraphics[width=\linewidth]{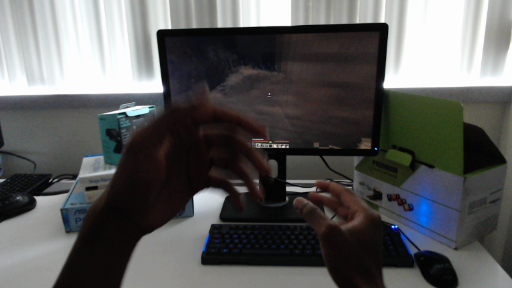}
  \end{subfigure}
  \begin{subfigure}[t]{0.16\linewidth}
    \includegraphics[width=\linewidth]{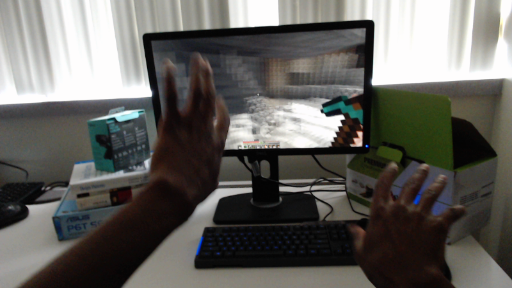}
  \end{subfigure}
  \vspace{1mm}
  \begin{subfigure}[t]{0.16\linewidth}
    \includegraphics[width=\linewidth]{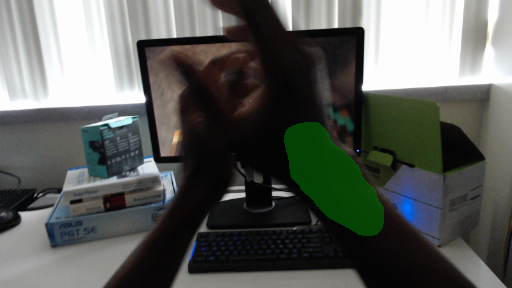}
  \end{subfigure}
   \begin{subfigure}[t]{0.16\linewidth}
    \includegraphics[width=\linewidth]{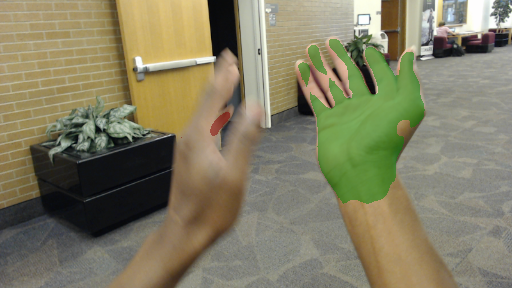}
  \end{subfigure}
  \begin{subfigure}[t]{0.16\linewidth}
    \includegraphics[width=\linewidth]{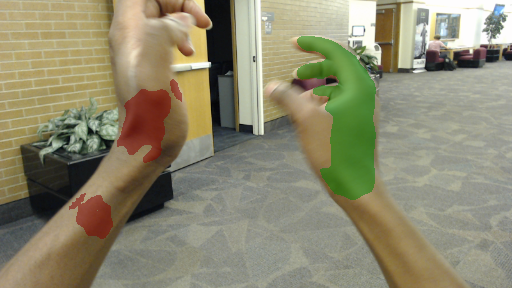}
  \end{subfigure}
  \begin{subfigure}[t]{0.16\linewidth}
    \includegraphics[width=\linewidth]{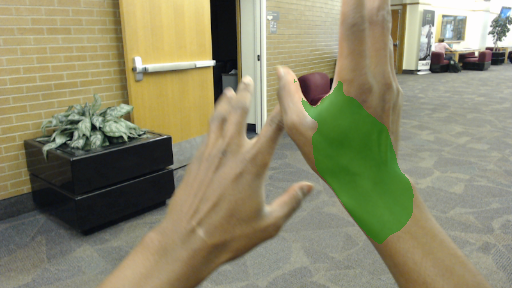}
  \end{subfigure}
  \begin{subfigure}[t]{0.16\linewidth}
    \includegraphics[width=\linewidth]{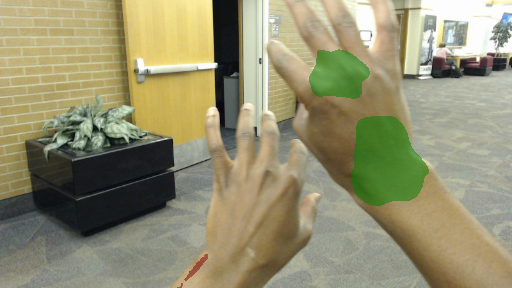}
  \end{subfigure}
  \begin{subfigure}[t]{0.16\linewidth}
    \includegraphics[width=\linewidth]{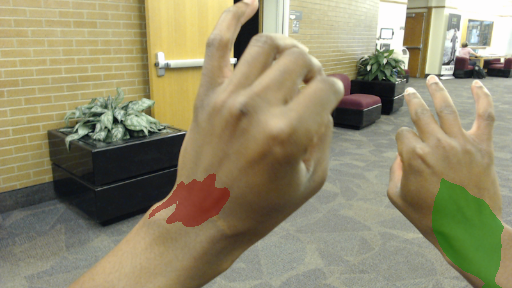}
  \end{subfigure}
  \vspace{1mm}
  \begin{subfigure}[t]{0.16\linewidth}
    \includegraphics[width=\linewidth]{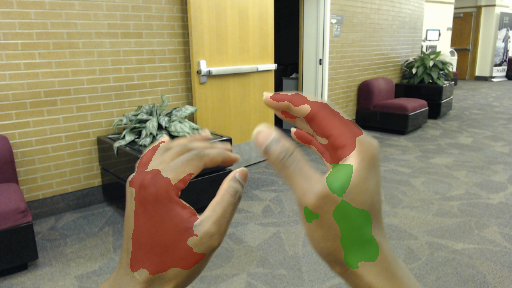}
  \end{subfigure}
   \begin{subfigure}[t]{0.16\linewidth}
    \includegraphics[width=\linewidth]{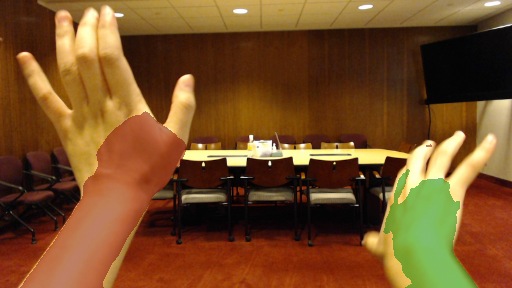}
    \caption*{t = 0}
  \end{subfigure}
  \begin{subfigure}[t]{0.16\linewidth}
    \includegraphics[width=\linewidth]{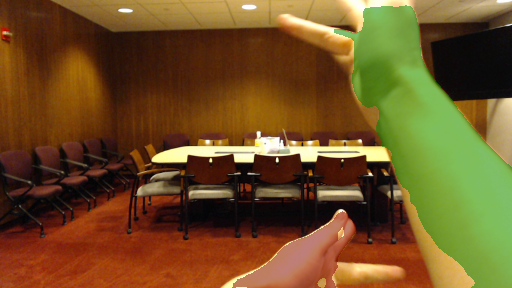}
    \caption*{t = 49}
  \end{subfigure}
  \begin{subfigure}[t]{0.16\linewidth}
    \includegraphics[width=\linewidth]{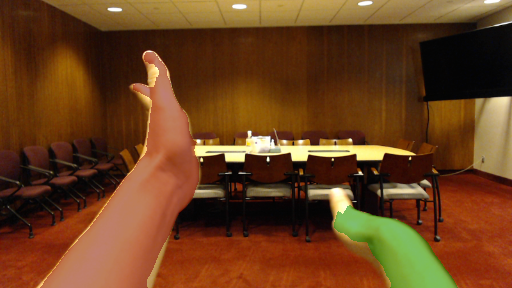}
    \caption*{t = 99}
  \end{subfigure}
  \begin{subfigure}[t]{0.16\linewidth}
    \includegraphics[width=\linewidth]{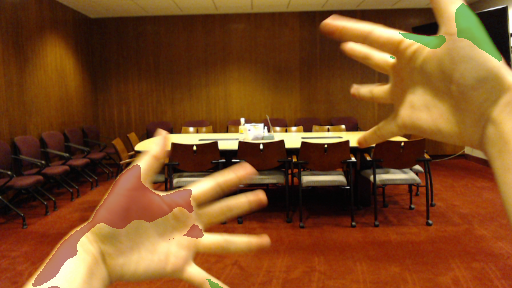}
    \caption*{t = 149}
  \end{subfigure}
  \begin{subfigure}[t]{0.16\linewidth}
    \includegraphics[width=\linewidth]{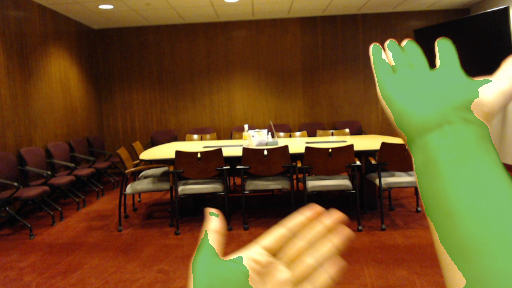}
    \caption*{t = 199}
  \end{subfigure}
  \vspace{4mm}
  \begin{subfigure}[t]{0.16\linewidth}
    \includegraphics[width=\linewidth]{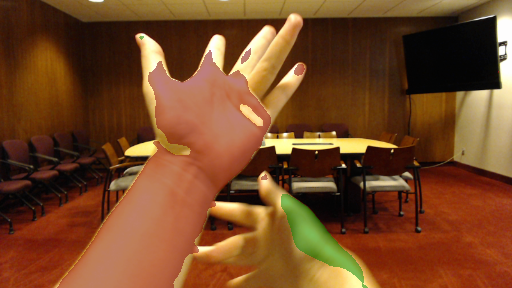}
    \caption*{t = 249}
  \end{subfigure}
  \caption{Qualitative samples of model trained on \textbf{Ego3DHands} on sequence 2, 5, 6, 8 of $\text{Ego2Hands}_{test}$.}
  \label{fig:cross_eval_ego3dhands}
\end{figure*}

\begin{figure*}[t]
  \centering
  \begin{subfigure}[t]{0.16\linewidth}
    \includegraphics[width=\linewidth]{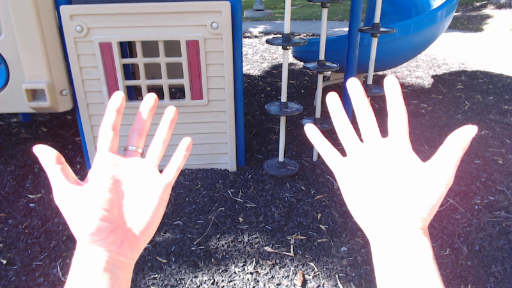}
  \end{subfigure}
  \begin{subfigure}[t]{0.16\linewidth}
    \includegraphics[width=\linewidth]{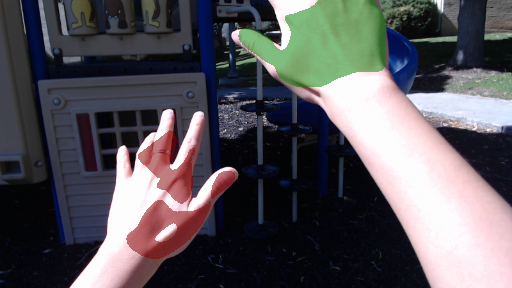}
  \end{subfigure}
  \begin{subfigure}[t]{0.16\linewidth}
    \includegraphics[width=\linewidth]{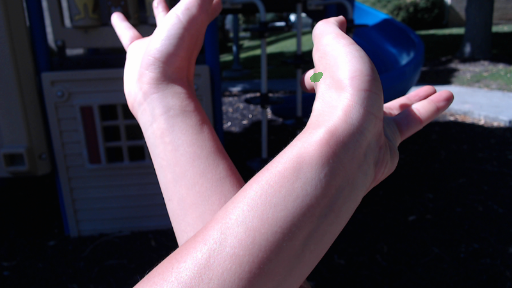}
  \end{subfigure}
  \begin{subfigure}[t]{0.16\linewidth}
    \includegraphics[width=\linewidth]{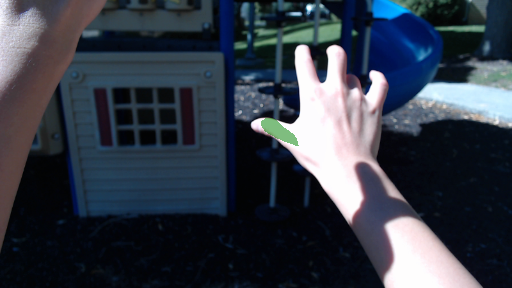}
  \end{subfigure}
  \begin{subfigure}[t]{0.16\linewidth}
    \includegraphics[width=\linewidth]{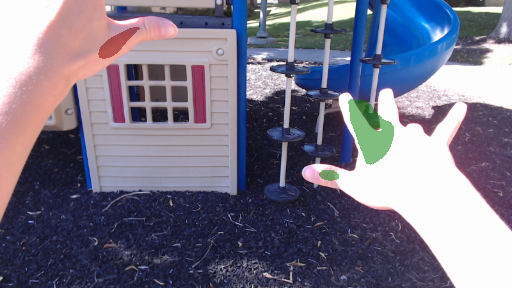}
  \end{subfigure}
  \vspace{1mm}
  \begin{subfigure}[t]{0.16\linewidth}
    \includegraphics[width=\linewidth]{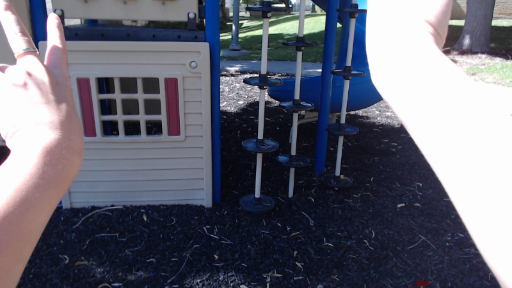}
  \end{subfigure}
   \begin{subfigure}[t]{0.16\linewidth}
    \includegraphics[width=\linewidth]{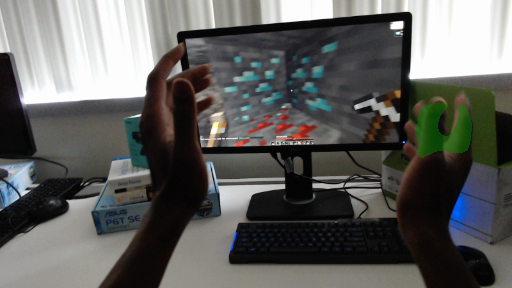}
  \end{subfigure}
  \begin{subfigure}[t]{0.16\linewidth}
    \includegraphics[width=\linewidth]{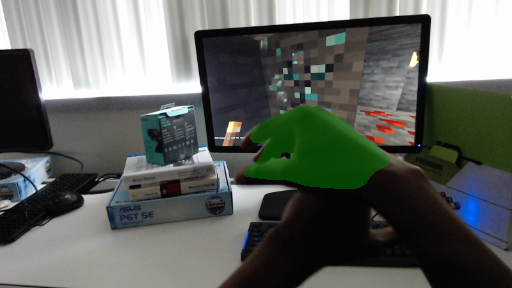}
  \end{subfigure}
  \begin{subfigure}[t]{0.16\linewidth}
    \includegraphics[width=\linewidth]{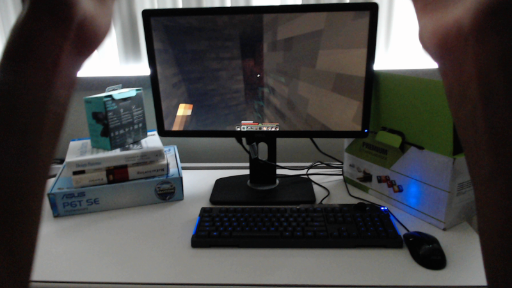}
  \end{subfigure}
  \begin{subfigure}[t]{0.16\linewidth}
    \includegraphics[width=\linewidth]{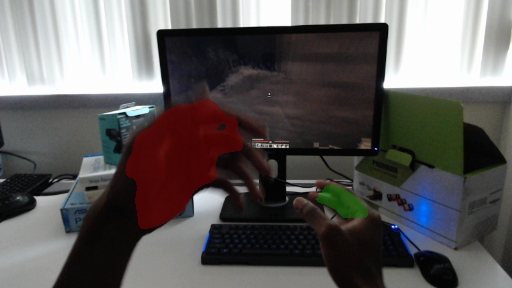}
  \end{subfigure}
  \begin{subfigure}[t]{0.16\linewidth}
    \includegraphics[width=\linewidth]{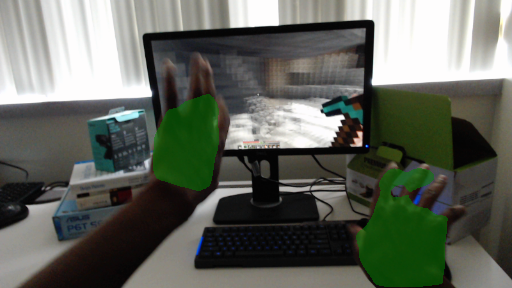}
  \end{subfigure}
  \vspace{1mm}
  \begin{subfigure}[t]{0.16\linewidth}
    \includegraphics[width=\linewidth]{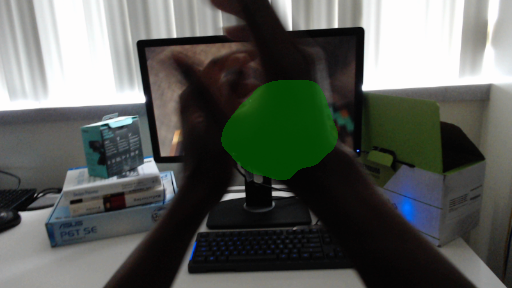}
  \end{subfigure}
   \begin{subfigure}[t]{0.16\linewidth}
    \includegraphics[width=\linewidth]{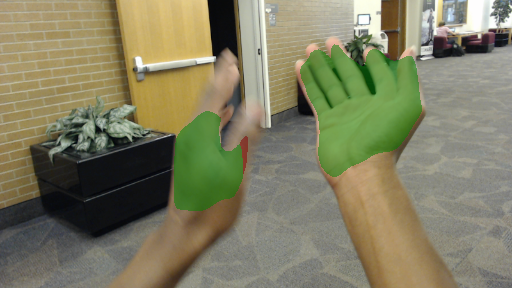}
  \end{subfigure}
  \begin{subfigure}[t]{0.16\linewidth}
    \includegraphics[width=\linewidth]{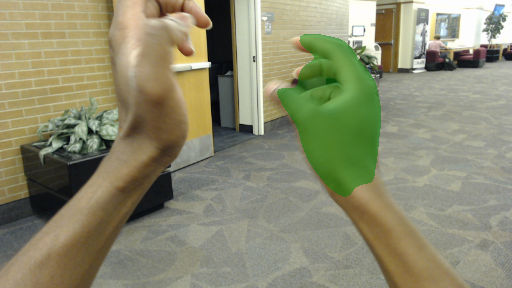}
  \end{subfigure}
  \begin{subfigure}[t]{0.16\linewidth}
    \includegraphics[width=\linewidth]{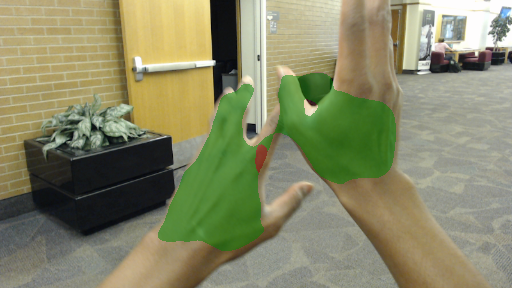}
  \end{subfigure}
  \begin{subfigure}[t]{0.16\linewidth}
    \includegraphics[width=\linewidth]{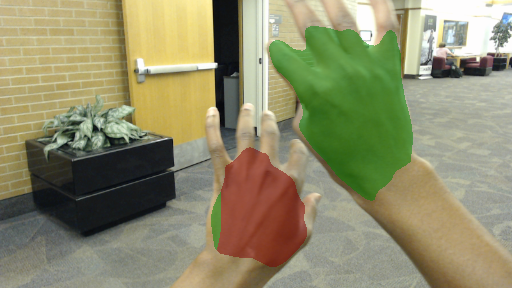}
  \end{subfigure}
  \begin{subfigure}[t]{0.16\linewidth}
    \includegraphics[width=\linewidth]{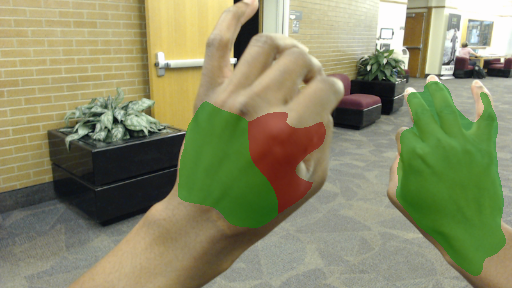}
  \end{subfigure}
  \vspace{1mm}
  \begin{subfigure}[t]{0.16\linewidth}
    \includegraphics[width=\linewidth]{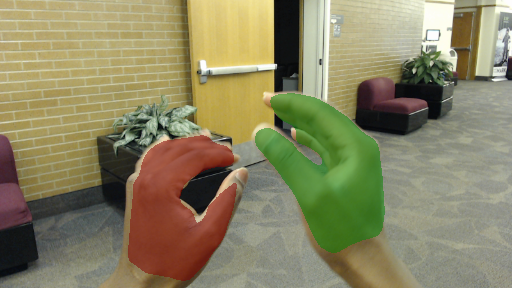}
  \end{subfigure}
   \begin{subfigure}[t]{0.16\linewidth}
    \includegraphics[width=\linewidth]{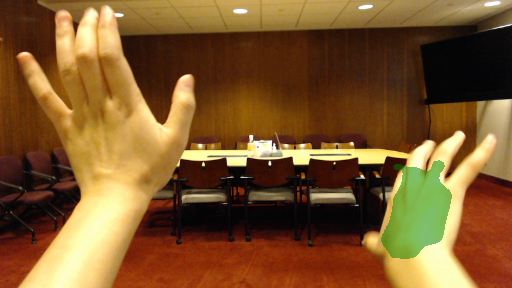}
    \caption*{t = 0}
  \end{subfigure}
  \begin{subfigure}[t]{0.16\linewidth}
    \includegraphics[width=\linewidth]{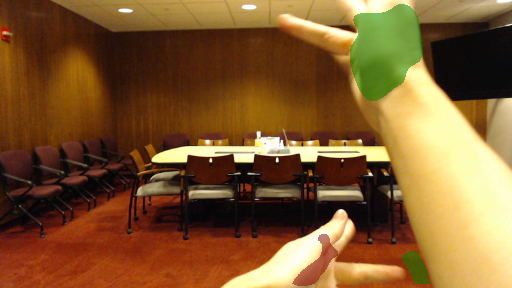}
    \caption*{t = 49}
  \end{subfigure}
  \begin{subfigure}[t]{0.16\linewidth}
    \includegraphics[width=\linewidth]{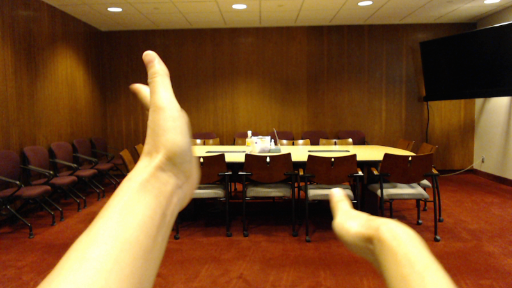}
    \caption*{t = 99}
  \end{subfigure}
  \begin{subfigure}[t]{0.16\linewidth}
    \includegraphics[width=\linewidth]{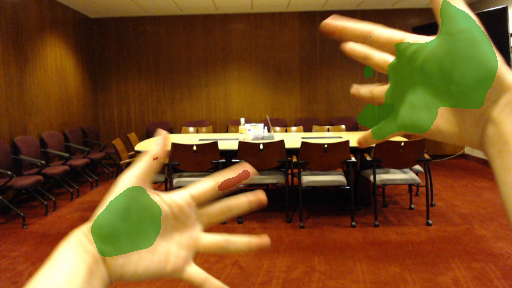}
    \caption*{t = 149}
  \end{subfigure}
  \begin{subfigure}[t]{0.16\linewidth}
    \includegraphics[width=\linewidth]{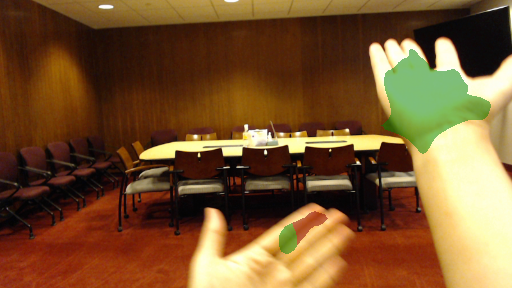}
    \caption*{t = 199}
  \end{subfigure}
  \vspace{4mm}
  \begin{subfigure}[t]{0.16\linewidth}
    \includegraphics[width=\linewidth]{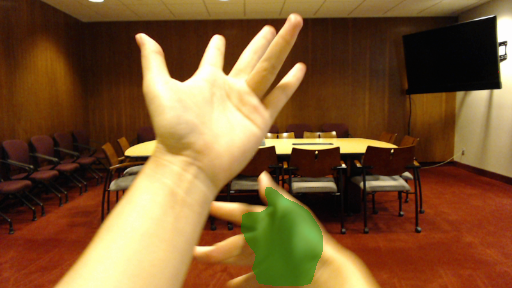}
    \caption*{t = 249}
  \end{subfigure}
  \caption{Qualitative samples of model trained on \textbf{EgoHands} on sequence 2, 5, 6, 8 of $\text{Ego2Hands}_{test}$.}
  \label{fig:cross_eval_egohands}
\end{figure*}

\begin{figure*}[t]
  \centering
  \begin{subfigure}[t]{0.16\linewidth}
    \includegraphics[width=\linewidth]{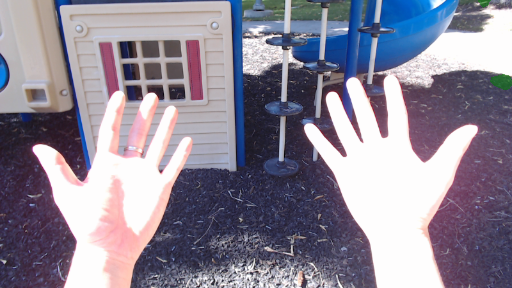}
  \end{subfigure}
  \begin{subfigure}[t]{0.16\linewidth}
    \includegraphics[width=\linewidth]{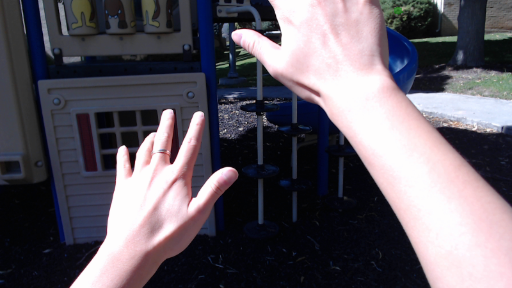}
  \end{subfigure}
  \begin{subfigure}[t]{0.16\linewidth}
    \includegraphics[width=\linewidth]{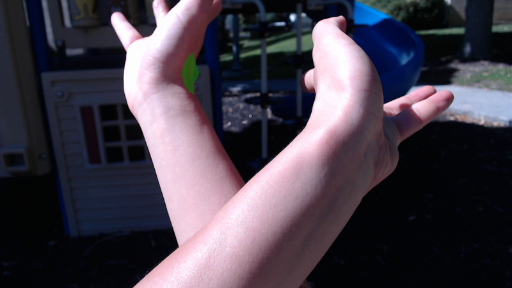}
  \end{subfigure}
  \begin{subfigure}[t]{0.16\linewidth}
    \includegraphics[width=\linewidth]{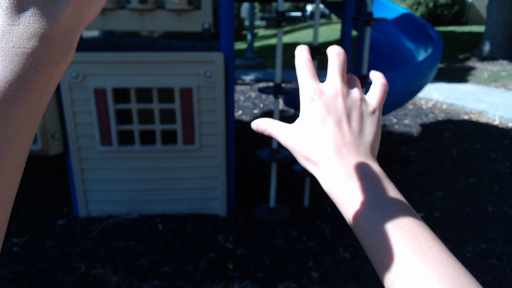}
  \end{subfigure}
  \begin{subfigure}[t]{0.16\linewidth}
    \includegraphics[width=\linewidth]{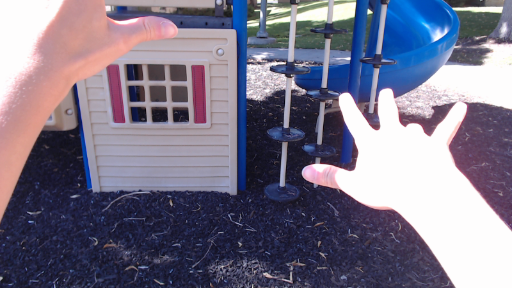}
  \end{subfigure}
  \vspace{1mm}
  \begin{subfigure}[t]{0.16\linewidth}
    \includegraphics[width=\linewidth]{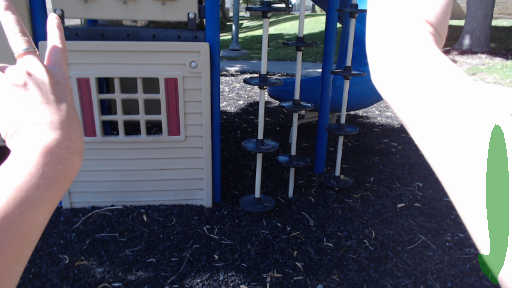}
  \end{subfigure}
   \begin{subfigure}[t]{0.16\linewidth}
    \includegraphics[width=\linewidth]{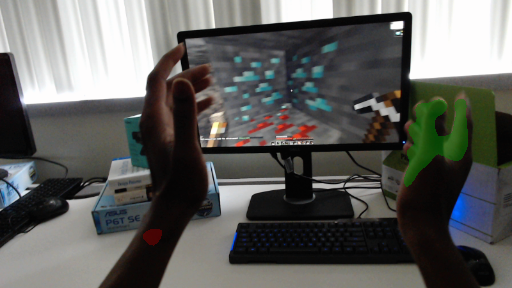}
  \end{subfigure}
  \begin{subfigure}[t]{0.16\linewidth}
    \includegraphics[width=\linewidth]{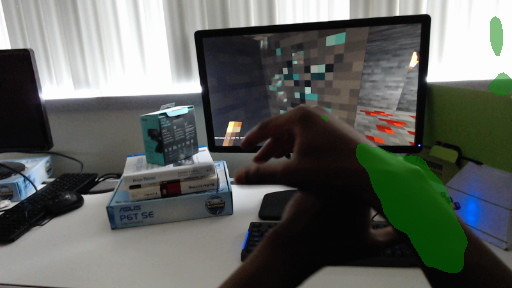}
  \end{subfigure}
  \begin{subfigure}[t]{0.16\linewidth}
    \includegraphics[width=\linewidth]{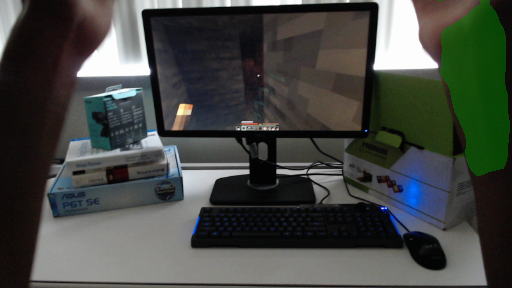}
  \end{subfigure}
  \begin{subfigure}[t]{0.16\linewidth}
    \includegraphics[width=\linewidth]{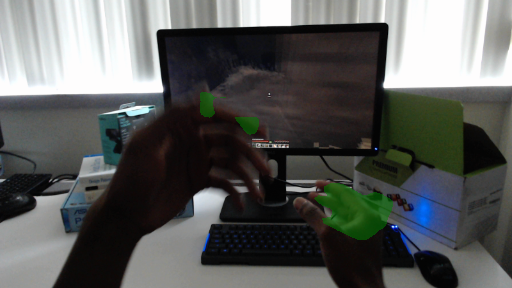}
  \end{subfigure}
  \begin{subfigure}[t]{0.16\linewidth}
    \includegraphics[width=\linewidth]{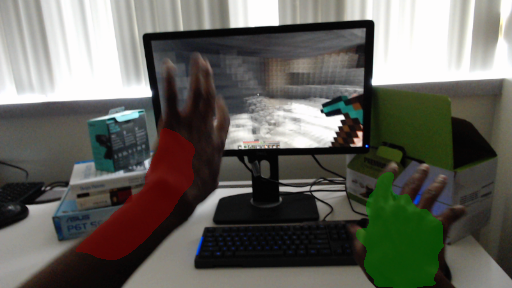}
  \end{subfigure}
  \vspace{1mm}
  \begin{subfigure}[t]{0.16\linewidth}
    \includegraphics[width=\linewidth]{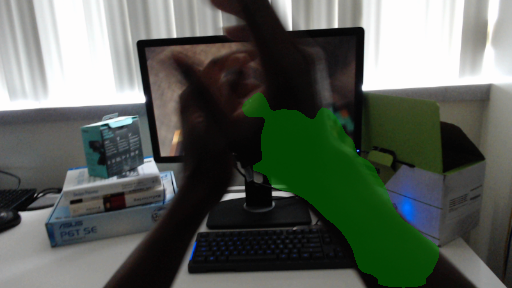}
  \end{subfigure}
   \begin{subfigure}[t]{0.16\linewidth}
    \includegraphics[width=\linewidth]{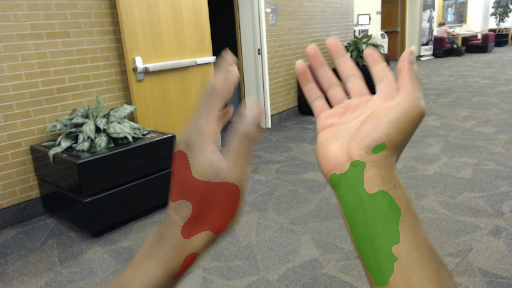}
  \end{subfigure}
  \begin{subfigure}[t]{0.16\linewidth}
    \includegraphics[width=\linewidth]{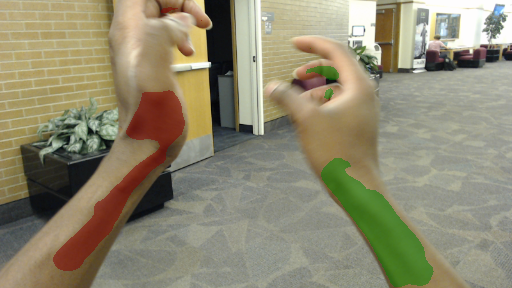}
  \end{subfigure}
  \begin{subfigure}[t]{0.16\linewidth}
    \includegraphics[width=\linewidth]{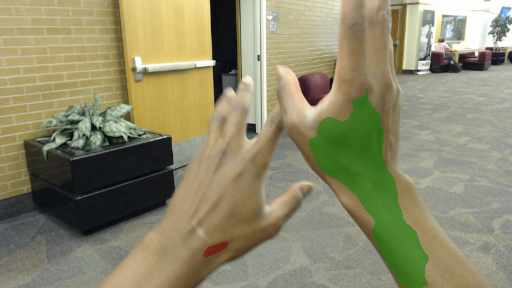}
  \end{subfigure}
  \begin{subfigure}[t]{0.16\linewidth}
    \includegraphics[width=\linewidth]{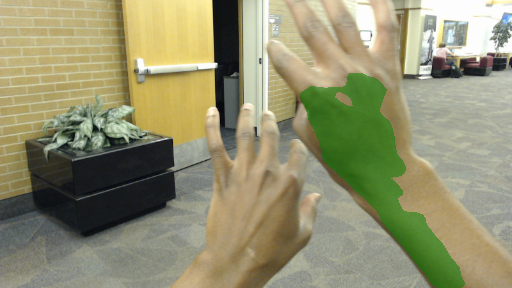}
  \end{subfigure}
  \begin{subfigure}[t]{0.16\linewidth}
    \includegraphics[width=\linewidth]{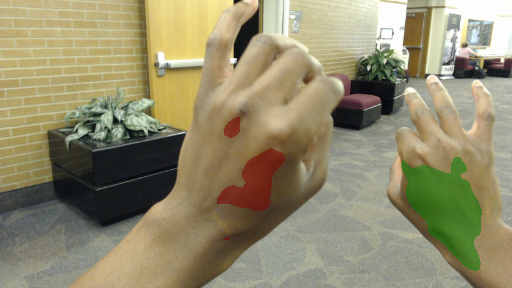}
  \end{subfigure}
  \vspace{1mm}
  \begin{subfigure}[t]{0.16\linewidth}
    \includegraphics[width=\linewidth]{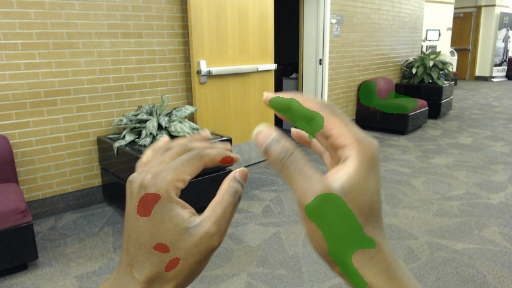}
  \end{subfigure}
   \begin{subfigure}[t]{0.16\linewidth}
    \includegraphics[width=\linewidth]{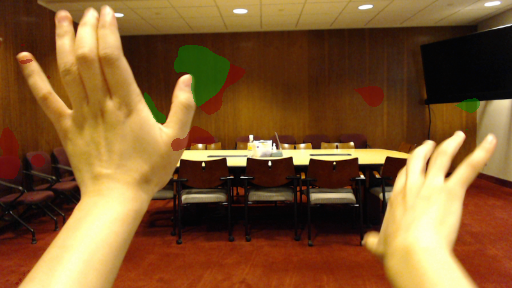}
    \caption*{t = 0}
  \end{subfigure}
  \begin{subfigure}[t]{0.16\linewidth}
    \includegraphics[width=\linewidth]{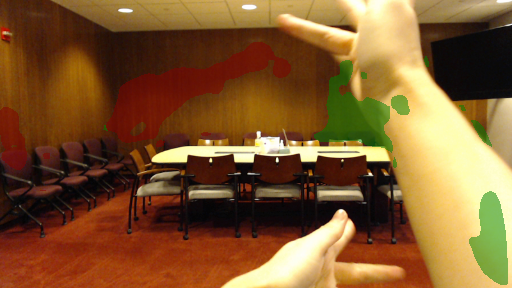}
    \caption*{t = 49}
  \end{subfigure}
  \begin{subfigure}[t]{0.16\linewidth}
    \includegraphics[width=\linewidth]{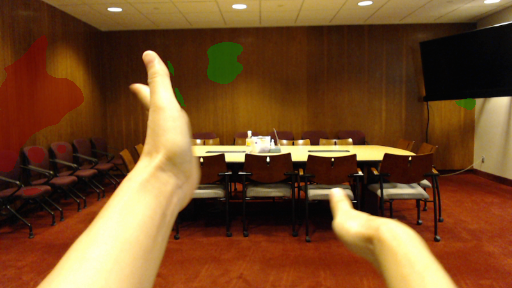}
    \caption*{t = 99}
  \end{subfigure}
  \begin{subfigure}[t]{0.16\linewidth}
    \includegraphics[width=\linewidth]{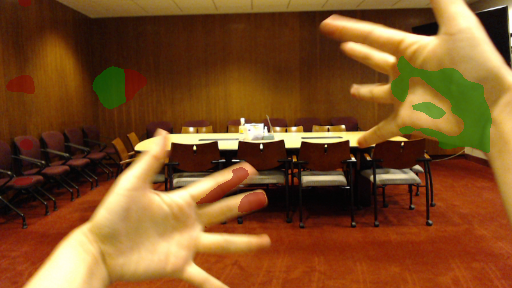}
    \caption*{t = 149}
  \end{subfigure}
  \begin{subfigure}[t]{0.16\linewidth}
    \includegraphics[width=\linewidth]{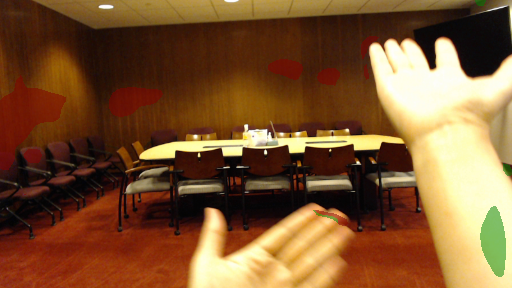}
    \caption*{t = 199}
  \end{subfigure}
  \vspace{4mm}
  \begin{subfigure}[t]{0.16\linewidth}
    \includegraphics[width=\linewidth]{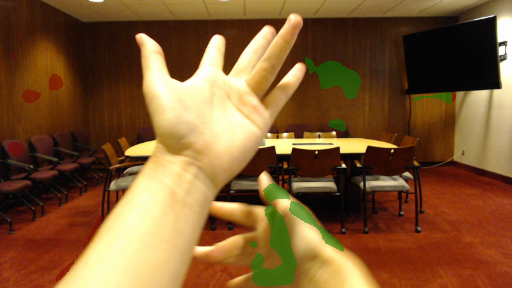}
    \caption*{t = 249}
  \end{subfigure}
  \caption{Qualitative samples of model trained on \textbf{GTEA} on sequence 2, 5, 6, 8 of $\text{Ego2Hands}_{test}$.}
  \label{fig:cross_eval_gtea}
\end{figure*}

\begin{figure*}[t]
  \centering
  \begin{subfigure}[t]{0.16\linewidth}
    \includegraphics[width=\linewidth]{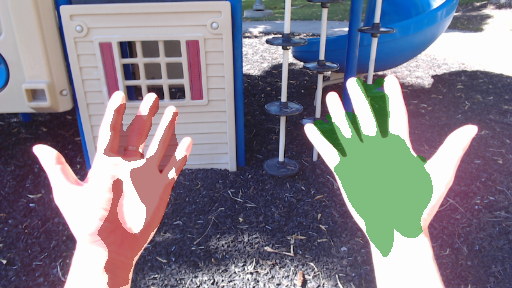}
  \end{subfigure}
  \begin{subfigure}[t]{0.16\linewidth}
    \includegraphics[width=\linewidth]{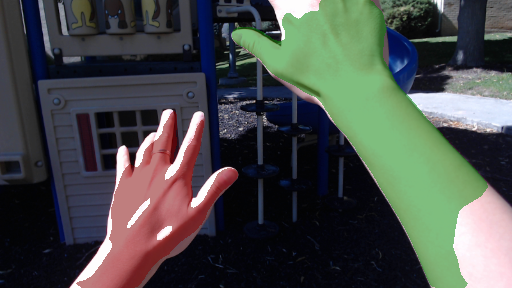}
  \end{subfigure}
  \begin{subfigure}[t]{0.16\linewidth}
    \includegraphics[width=\linewidth]{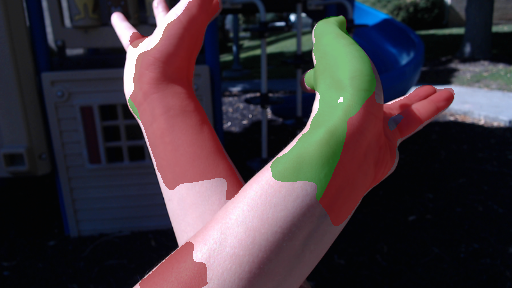}
  \end{subfigure}
  \begin{subfigure}[t]{0.16\linewidth}
    \includegraphics[width=\linewidth]{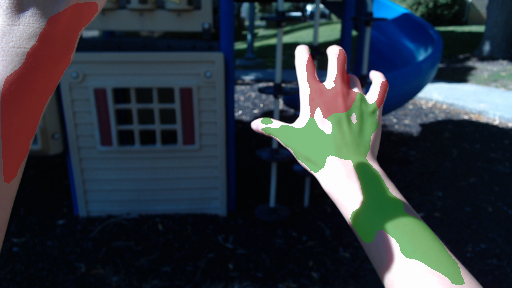}
  \end{subfigure}
  \begin{subfigure}[t]{0.16\linewidth}
    \includegraphics[width=\linewidth]{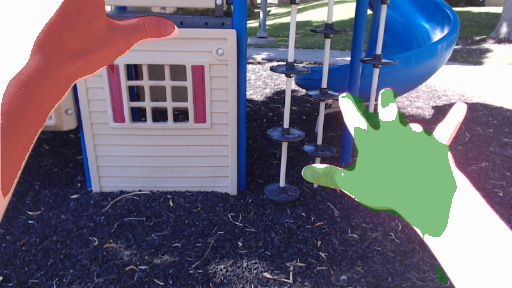}
  \end{subfigure}
  \vspace{1mm}
  \begin{subfigure}[t]{0.16\linewidth}
    \includegraphics[width=\linewidth]{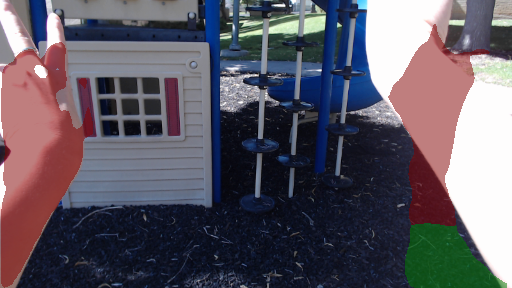}
  \end{subfigure}
   \begin{subfigure}[t]{0.16\linewidth}
    \includegraphics[width=\linewidth]{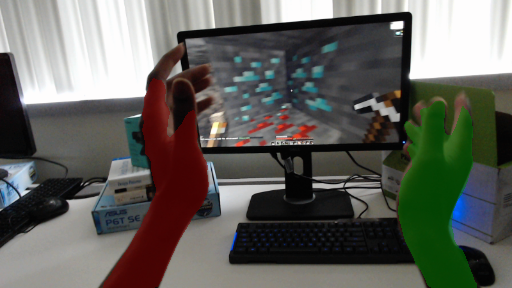}
  \end{subfigure}
  \begin{subfigure}[t]{0.16\linewidth}
    \includegraphics[width=\linewidth]{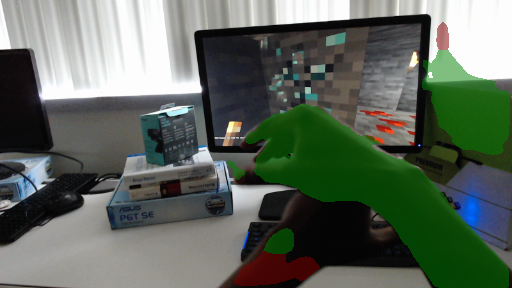}
  \end{subfigure}
  \begin{subfigure}[t]{0.16\linewidth}
    \includegraphics[width=\linewidth]{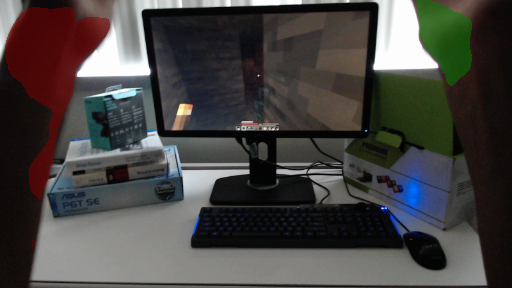}
  \end{subfigure}
  \begin{subfigure}[t]{0.16\linewidth}
    \includegraphics[width=\linewidth]{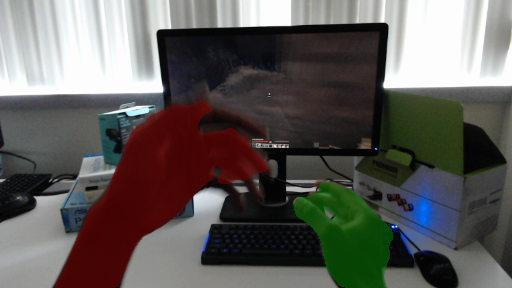}
  \end{subfigure}
  \begin{subfigure}[t]{0.16\linewidth}
    \includegraphics[width=\linewidth]{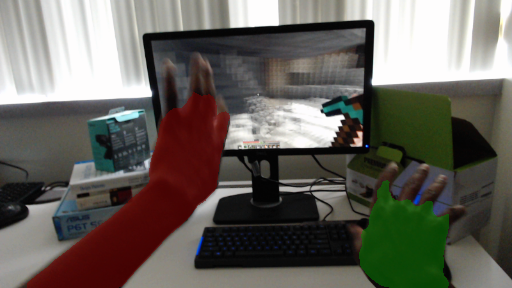}
  \end{subfigure}
  \vspace{1mm}
  \begin{subfigure}[t]{0.16\linewidth}
    \includegraphics[width=\linewidth]{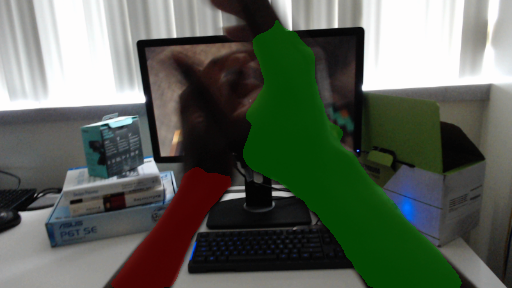}
  \end{subfigure}
   \begin{subfigure}[t]{0.16\linewidth}
    \includegraphics[width=\linewidth]{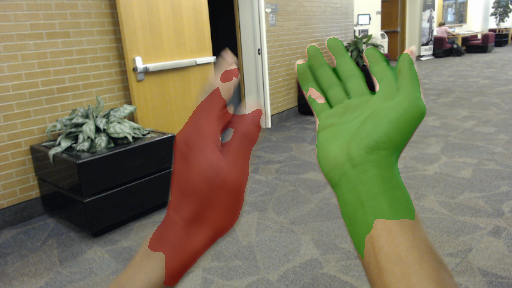}
  \end{subfigure}
  \begin{subfigure}[t]{0.16\linewidth}
    \includegraphics[width=\linewidth]{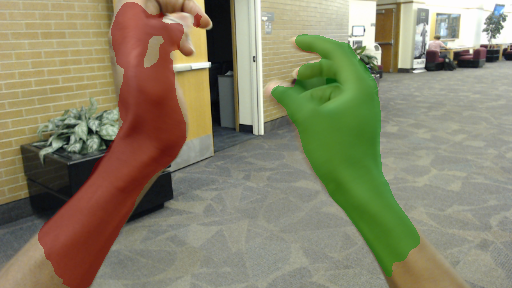}
  \end{subfigure}
  \begin{subfigure}[t]{0.16\linewidth}
    \includegraphics[width=\linewidth]{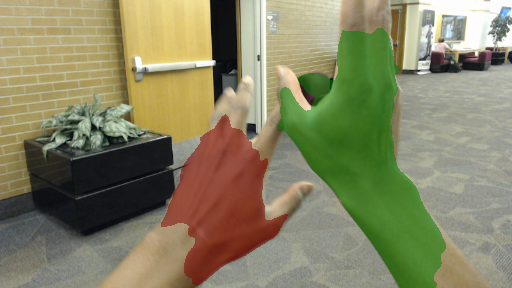}
  \end{subfigure}
  \begin{subfigure}[t]{0.16\linewidth}
    \includegraphics[width=\linewidth]{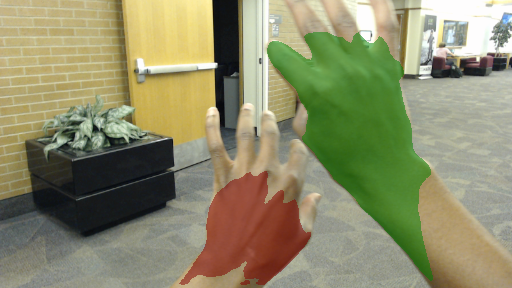}
  \end{subfigure}
  \begin{subfigure}[t]{0.16\linewidth}
    \includegraphics[width=\linewidth]{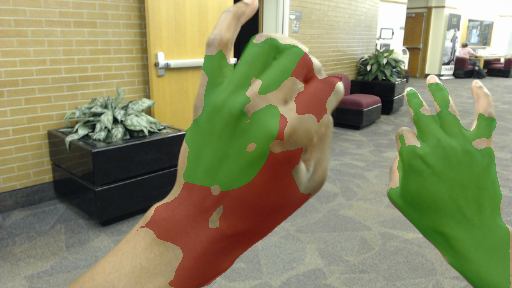}
  \end{subfigure}
  \vspace{1mm}
  \begin{subfigure}[t]{0.16\linewidth}
    \includegraphics[width=\linewidth]{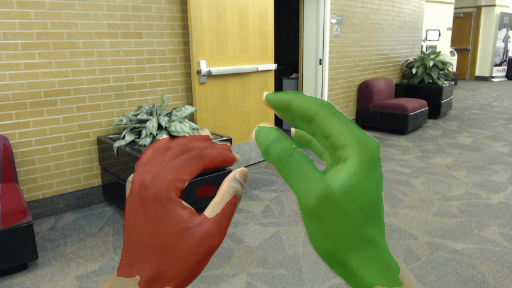}
  \end{subfigure}
   \begin{subfigure}[t]{0.16\linewidth}
    \includegraphics[width=\linewidth]{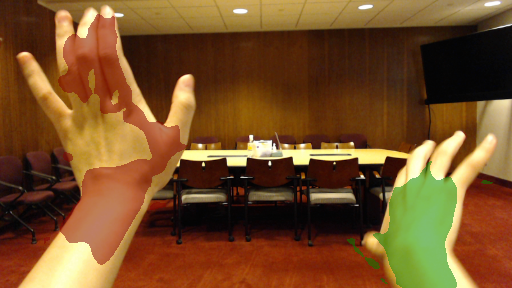}
    \caption*{t = 0}
  \end{subfigure}
  \begin{subfigure}[t]{0.16\linewidth}
    \includegraphics[width=\linewidth]{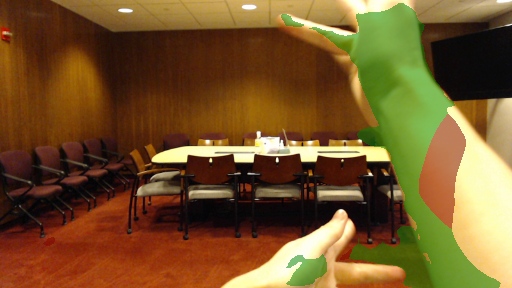}
    \caption*{t = 49}
  \end{subfigure}
  \begin{subfigure}[t]{0.16\linewidth}
    \includegraphics[width=\linewidth]{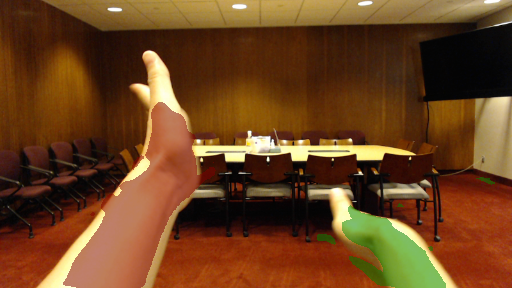}
    \caption*{t = 99}
  \end{subfigure}
  \begin{subfigure}[t]{0.16\linewidth}
    \includegraphics[width=\linewidth]{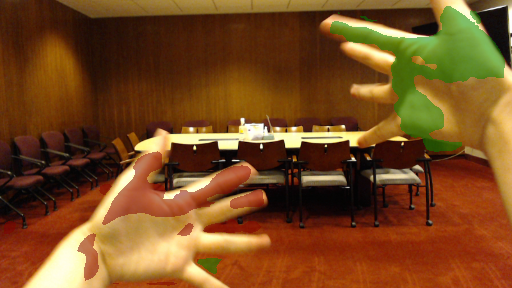}
    \caption*{t = 149}
  \end{subfigure}
  \begin{subfigure}[t]{0.16\linewidth}
    \includegraphics[width=\linewidth]{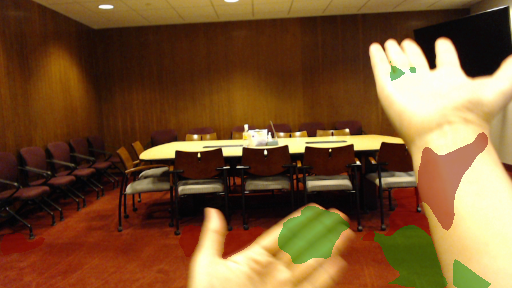}
    \caption*{t = 199}
  \end{subfigure}
  \vspace{4mm}
  \begin{subfigure}[t]{0.16\linewidth}
    \includegraphics[width=\linewidth]{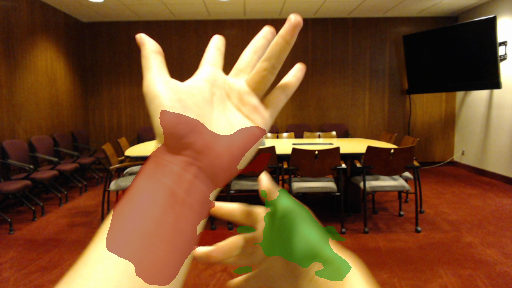}
    \caption*{t = 249}
  \end{subfigure}
  \caption{Qualitative samples of model trained on \textbf{EGTEA} on sequence 2, 5, 6, 8 of $\text{Ego2Hands}_{test}$.}
  \label{fig:cross_eval_egtea}
\end{figure*}

In this section, we show additional qualitative samples of RefineNet evaluating on the $\text{Ego2Hands}_{test}$ for cross-dataset evaluation. In this experiment, models are only trained on the training set of each dataset and are not trained for domain adaptation. To demonstrate that models trained on other datasets with limited quantity and diversity have difficulty generalizing to the unseen domains of Ego2Hands, we show results evaluating on sequence 2, 5, 6, 8 of $\text{Ego2Hands}_{test}$, which contain skin color or illumination different from the standard distribution of the other datasets. We first show results of Refinenet trained on $\text{Ego2Hands}_{train}$ in Fig. \ref{fig:cross_eval_ego2hands}. We emphasize that scenes and subjects in the test set of Ego2Hands are not present in the corresponding training set. As a result, Fig. \ref{fig:cross_eval_ego2hands} indicates that training on the composited data of $\text{Ego2Hands}_{train}$ achieves high generalization accuracy.\\
\indent Fig. \ref{fig:cross_eval_ego3dhands}-\ref{fig:cross_eval_egtea} show results achieved from models trained on Ego3DHands, EgoHands, GTEA and EGTEA respectively. In Fig. \ref{fig:cross_eval_ego3dhands}, we see that training on synthetic data does have some generalization ability on real-world data. Despite EgoHands containing segmentation masks that exclude the arm, fig. \ref{fig:cross_eval_egohands} shows that RefineNet trained on EgoHands has low accuracy for distinguishing the left and right hand. It also has very unstable and inaccurate predictions for both hands in general. Fig. \ref{fig:cross_eval_gtea} shows that training on GTEA has very low generalization accuracy and can completely fail in cases of bright illumination. In Fig. \ref{fig:cross_eval_egtea}, we see that training on EGTEA can enable limited generalization for both hands in diverse scenes. However, its accuracy for two-hand segmentation still significantly falls behind RefineNet trained on our dataset. Note that in the case of inter-hand occlusion and cross-over (e.g. t = 99 in row 1), the model trained on EGTEA mistakenly predicts the right hand as the left hand. This problem is caused by the lack of instances with inter-hand occlusion from datasets such as EgoHands, GTEA and EGTEA. \\
\indent We point out that training on our dataset not only provides high generalization accuracy in two-hand segmentation, but also enables two-hand detection despite inter-hand occlusion, an important feature unavailable from training using other hand segmentation datasets.
\end{document}